\documentclass{cparxiv}

\usepackage{amsthm}
\usepackage{colortbl}
\usepackage{adjustbox}
\usepackage{placeins}
\usepackage{microtype}

\newtheorem{theorem}{Theorem}

\usepackage{tikz}
\usepackage{cuted}
\usepackage{currfile}
\usepackage{multirow}
\usepackage{algorithm}
\usepackage{algpseudocode}
\usepackage{pifont}

\newtheorem{proposition}{Proposition}
\newtheorem{corollary}{Corollary}
\newtheorem{lemma}{Lemma}

\usepackage{tikz}
\usetikzlibrary{positioning,arrows.meta,fit,calc,backgrounds,decorations.pathreplacing,shapes.geometric}

\definecolor{bpOurs}{RGB}{0,114,178}      % our method / our new component
\definecolor{bpPrior}{RGB}{213,94,0}      % prior work, failure, cost
\definecolor{bpData}{RGB}{0,158,115}      % data, queries, inputs
\definecolor{bpAccent}{RGB}{230,159,0}    % a third role if needed
\definecolor{bpGray}{RGB}{110,110,110}    % reused / frozen / given parts, annotations
\definecolor{bpLight}{RGB}{238,238,238}   % neutral fills
\colorlet{bpOursFill}{bpOurs!18}
\colorlet{bpPriorFill}{bpPrior!15}
\colorlet{bpDataFill}{bpData!15}

\tikzset{
  bp/base/.style   ={draw=black!70, line width=0.5pt, rounded corners=2pt, align=center,
                     font=\sffamily\scriptsize, minimum height=0.6cm, inner sep=3pt},
  bp/module/.style ={bp/base, fill=white},
  bp/frozen/.style ={bp/base, fill=bpLight, draw=bpGray, text=black!80},
  bp/ours/.style   ={bp/base, fill=bpOursFill, draw=bpOurs, line width=0.7pt},
  bp/prior/.style  ={bp/base, fill=bpPriorFill, draw=bpPrior},
  bp/data/.style   ={bp/base, fill=bpDataFill, draw=bpData},
  bp/todo/.style   ={bp/base, dashed, draw=bpGray, text=bpGray},
  bp/flow/.style   ={-{Latex[length=1.6mm]}, line width=0.6pt, black!75},
  bp/note/.style   ={font=\sffamily\tiny, text=bpGray, align=center},
  bp/panel/.style  ={font=\sffamily\footnotesize\bfseries, anchor=west},
  bp/frame/.style  ={draw=black!70, fill=white, line width=0.4pt},
}

\definecolor{bpOurs}{HTML}{D1613E}
\definecolor{bpPrior}{HTML}{184198}
\definecolor{bpTeal}{RGB}{83,74,183}
\definecolor{bpCPPanel}{HTML}{C0DBD4}
\colorlet{bpOursFill}{bpCPPanel!45}
\colorlet{bpPriorFill}{bpPrior!12}
\colorlet{bpPanel}{black!5}
\newcommand{\up}[1]{\textsuperscript{\textcolor{bpOurs}{+#1}}}

\newcommand{\cp}{\,+\,CP}
\newcommand{\cpup}{\,+\,CP}
\newcommand{\eacc}{ACC}
\newcommand{\Hv}{\ensuremath{H_{\mathrm{v}}}}
\newcommand{\Hf}{\ensuremath{H_{\mathrm{f}}}}

\AtEndPreamble{\AtBeginDocument{%
  \crefname{appendix}{Supp.}{Supp.}\Crefname{appendix}{Supp.}{Supp.}%
  \crefname{subappendix}{Supp.}{Supp.}\Crefname{subappendix}{Supp.}{Supp.}%
  \crefname{subsubappendix}{Supp.}{Supp.}\Crefname{subsubappendix}{Supp.}{Supp.}}}
\makeatletter
\newcommand{\startsupplement}{%
  \clearpage
  \appendix
  \renewcommand{\thesection}{\Alph{section}}%
  \renewcommand{\thefigure}{S\arabic{figure}}\renewcommand{\theHfigure}{S\arabic{figure}}%
  \renewcommand{\thetable}{S\arabic{table}}\renewcommand{\theHtable}{S\arabic{table}}%
  \renewcommand{\theequation}{S\arabic{equation}}\renewcommand{\theHequation}{S\arabic{equation}}%
  \renewcommand{\theHtheorem}{S\arabic{theorem}}%
  \setcounter{figure}{0}\setcounter{table}{0}\setcounter{equation}{0}\setcounter{theorem}{0}%
  \setlength{\emergencystretch}{1.5em}%
  \renewcommand{\topfraction}{0.9}\renewcommand{\dbltopfraction}{0.9}%
  \renewcommand{\textfraction}{0.1}%
  \renewcommand{\floatpagefraction}{0.8}\renewcommand{\dblfloatpagefraction}{0.8}%
  \setcounter{topnumber}{4}\setcounter{dbltopnumber}{3}\setcounter{totalnumber}{6}%
  \setlength{\@fptop}{0pt}\setlength{\@fpsep}{10pt plus 2pt}\setlength{\@fpbot}{0pt plus 1fil}%
  \setlength{\@dblfptop}{0pt}\setlength{\@dblfpsep}{10pt plus 2pt}\setlength{\@dblfpbot}{0pt plus 1fil}%
  \twocolumn[{%
    {\color{rvblue}\hrule height 2.2pt}\vspace{5mm}%
    {\centering\bfseries\fontsize{16}{20}\selectfont\color{rvink}\@title\par}%
    \vspace{3mm}%
    {\centering\rvsans\bfseries\fontsize{12}{15}\selectfont\color{rvblue}Supplementary Material\par}%
    \vspace{4mm}{\color{rvblue}\hrule height 0.6pt}\vspace{6mm}}]}
\makeatother

\usepackage[breaklinks,colorlinks,allcolors=rvblue,pdfcreator={},pdfproducer={}]{hyperref}
\pdftrailerid{}

\title{From Points to Trajectories:\\Rethinking Multimodal Alignment for Physical Events}
\shorttitle{From Points to Trajectories: Rethinking Multimodal Alignment for Physical Events}

\newsavebox\cplogoA\newsavebox\cplogoB\newsavebox\cplogoC\newlength\cplogogap
\logos{%
  \sbox\cplogoA{\rvlogo[8mm]{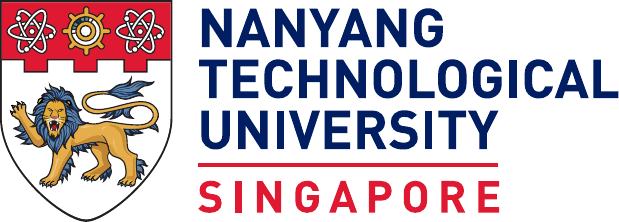}}%
  \sbox\cplogoB{\rvlogo[8mm]{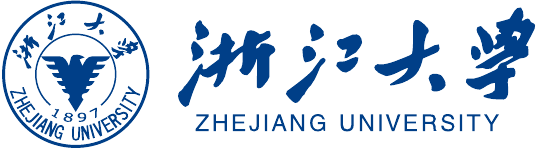}}%
  \sbox\cplogoC{\rvlogo[7mm]{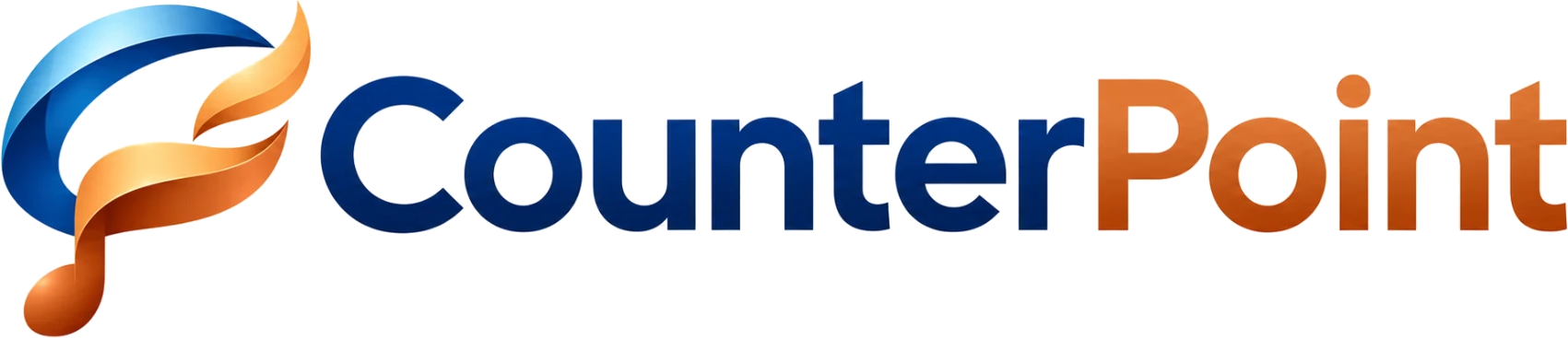}}%
  \setlength\cplogogap{\dimexpr((\textwidth-\wd\cplogoC)/2-\wd\cplogoA-\wd\cplogoB)/2\relax}%
  \usebox\cplogoA\hspace{\cplogogap}\usebox\cplogoB\hspace{\cplogogap}\usebox\cplogoC\hfill}

\author{%
  \textbf{Yiyang Zhu}\textsuperscript{1}, \textbf{Guangyuan Liu}\textsuperscript{1},
  \textbf{Jingdi Lei}\textsuperscript{1}, \textbf{Li Wei}\textsuperscript{2},
  \textbf{Chau Yuen}\textsuperscript{1}}
\affiliation{\textsuperscript{1}Nanyang Technological University,
  \textsuperscript{2}Zhejiang University}

\metadata{%
  \metaitem{Date}{November 18, 2026}%
}

\abstract{%
Advancing physical artificial intelligence (Physical AI) requires moving beyond static object recognition to ground how real-world mechanics (e.g., impact, friction, and force transfer) unfold over time. While multimodal aligners seek to capture these interactions by matching multi-sensor streams, prevailing architectures compress each temporal sequence into a single pooled embedding. We show that this paradigm fails on twin events: distinct physical executions recorded under identical task conditions. Collapsing the temporal dimension averages out transient cross-modal dynamics, effectively reducing physical event alignment to static scene recognition. Formally, we prove that as sequence length increases, any temporally pooled representation converges to the stationary recording condition. Consequently, as windows grow, no choice of loss function or embedding geometry can disambiguate twin events from pooled features alone. Instead, we show that physical events are uniquely identified by the event gap, namely the instantaneous, synchronous co-fluctuations across modalities that are erased by pooling and accumulate only along unpooled trajectories. To recover this signal, we introduce CounterPoint, a lightweight plug-in module that strips out each modality's temporal mean to score frame-level cross-modal agreement. Integrated across four state-of-the-art (SOTA) multimodal aligners, CounterPoint shifts representations from coarse condition matching to fine-grained physical event alignment, consistently outperforming existing temporal alignment methods.
}

\teaser{%
  \centering
  \includegraphics[width=\textwidth]{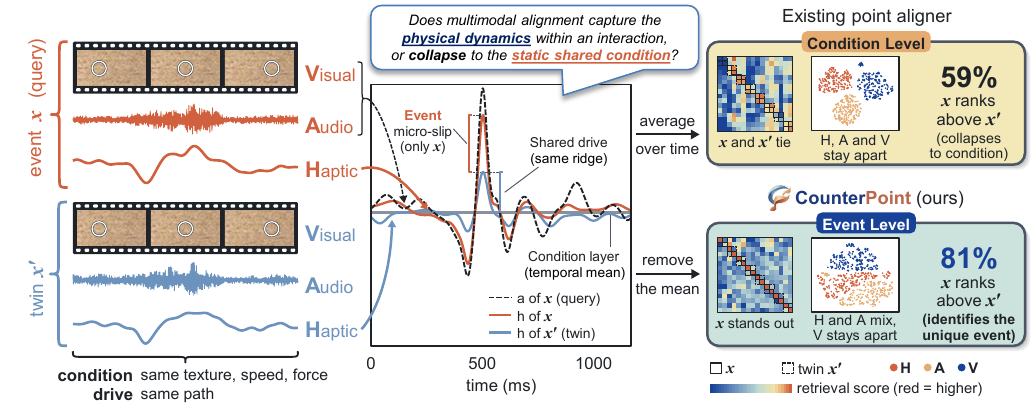}
  \caption{\textbf{Points keep the condition, trajectories distinguish the event.} Left: twin events share texture, speed, force and path. Middle: a micro-slip in $x$ produces synchronous audio--haptic fluctuations. Right: the point nearly ties $x$ with $x'$. CounterPoint separates them using demeaned, time-locked trajectories.}
  \label{fig:teaser}}

\begin{document}
% The table of contents of the supplement lists only its own sections.
\addtocontents{toc}{\protect\setcounter{tocdepth}{-1}}
\maketitle
\section{Introduction}
\label{sec:intro}

Physical artificial intelligence (Physical AI) requires perceiving, modeling, and acting upon the real world through continuous multi-sensor feedback~\cite{zhang2026physicalai,wu2024physicalscene}.
As physical interactions unfold over time, they produce complementary visual, auditory, and tactile signals~\cite{li2022seehearfeel,eguchi2024cluster,devillard2025havdb} rooted in a single physical cause: a micro-slip at contact, for instance, simultaneously generates acoustic emission and tactile vibration.
This synchronous co-occurrence provides a natural signal for cross-modal learning~\cite{owens2016ambient}, binding disparate sensor streams into a unified physical event~\cite{wallace2014tbw,diederich2015window,zhu2026eventnative}.
Multimodal aligners learn this correspondence without manual labels via contrastive instance discrimination~\cite{wu2018instance,radford2021clip}, mapping concurrent streams into a shared embedding space.

Prevailing architectures compute this similarity by compressing each temporal recording into a single pooled vector, or \emph{point}, whether trained with pairwise CLIP-style losses~\cite{radford2021clip} or multimodal geometric objectives~\cite{cicchetti2025gram,na2026hypergram,liu2025pmrl,tang2026barybind,cicchetti2025triangle}.
Much like bag-of-words behavior~\cite{yuksekgonul2023bow}, single-frame biases~\cite{buch2022atp,lei2023singleframe}, and order confusion~\cite{bagad2023testoftime} in vision--language models, point representations allow contrastive learners to rely on static shortcuts~\cite{geirhos2020shortcut}.
Standard evaluation against random negatives rarely exposes this limitation, as static context alone suffices to separate unrelated recordings~\cite{yuksekgonul2023bow}.

In physical interaction, however, agents repeatedly act under the same macroscopic \emph{condition}, for instance grasping the same object or sliding across the same surface~\cite{eguchi2024cluster,openx2024,khazatsky2024droid}.
Each execution shares this stationary setup while producing distinct moment-to-moment fluctuations in sound and touch, yielding what we call \emph{twin events}.
Distinguishing such pairs raises a fundamental question for Physical AI: 

\vbox{\textbf{\emph{Does multimodal alignment capture the physical dynamics within an interaction, or collapse to a static shared condition?}}}

Evaluating against the shared capture clock reveals that prevailing aligners fail this test: given the audio and visual streams of a stroke, they retrieve the haptic recording of its twin instead of the true co-occurring partner (\cref{fig:teaser}).
We trace this failure to temporal pooling and prove that, as recordings grow, pooled points converge to a limit governed solely by the condition, regardless of training loss or score geometry, a phenomenon we term \emph{condition collapse} (\cref{prop:pooling}).
Pooling also erases temporal order: a physically impossible, time-shuffled recording yields the exact same vector and scores closer than a real twin event.

What pooling averages out are the synchronous cross-modal fluctuations that uniquely identify an execution.
Through the Koopman operator~\cite{koopman1931hamiltonian}, we decompose sensor streams into a stationary \emph{condition layer}, a \emph{drive layer} shared along the path, and an \emph{event layer} specific to each interaction (\cref{fig:teaser}, middle).
Viewed within a single modality, the event layer resembles unstructured noise. Compared across modalities, its synchronous co-fluctuations provide verifiable evidence of a shared physical event.
We formalize this cross-modal covariance under a shared drive as the \emph{event gap} and prove that it captures the exact advantage of a time-locked trajectory score over a twin event (\cref{thm:levels}).

As observations grow, point accuracy therefore stalls while zero-mean \emph{trajectories} continue to accumulate event evidence (\cref{prop:stall,fig:scaling}).
To exploit this property, we introduce \textbf{CounterPoint}, a lightweight plug-in branch that removes each modality's temporal mean and scores cross-modal agreement at matching instants (\cref{fig:framework}).
Across controlled and free-hand texture sliding, CounterPoint substantially boosts physical event alignment across state-of-the-art (SOTA) aligners, with gains that follow the event gap of each modality pair (\cref{fig:mechanism}). 

To summarize, our contributions are:
\begin{itemize}
    \item \looseness=-1 We show that the prevailing point-based multimodal alignment paradigm fails to ground physical interactions: as sequence length grows, temporally pooled \emph{points} undergo \emph{condition collapse}, converging to the stationary task condition regardless of loss function or score geometry and rendering \emph{twin events} indistinguishable.
    \item We formalize the physical coupling erased by pooling as the \emph{event gap}, the instantaneous cross-modal covariance of transient contact fluctuations under a shared drive. Whereas within-modality pooling cancels out these zero-mean fluctuations, time-locked \emph{trajectories} accumulate their cross-modal covariance over time, guaranteeing asymptotic event separation whenever the gap is positive.
    \item To capture this trajectory evidence in practice, we introduce \textbf{CounterPoint}, a lightweight ($0.4\%$ add-on parameters) branch that removes each modality's temporal mean and scores frame-level cross-modal agreement. Across four SOTA aligners, CounterPoint turns unreliable visual retrieval on free-hand interactions (Recall@1, or R@1, $\le 10.1\%$) into reliable physical event alignment (at least $93.7\%$) and more than doubles unseen audio--haptic retrieval (from at most $27.4\%$ to at least $56.0\%$), with gains across modalities directly tracking the physical event gap.
\end{itemize}
\section{Related Work}
\label{sec:related}

\paragraph{Pooled multimodal alignment.}
Contrastive alignment typically matches modalities via the cosine similarity of temporally pooled embeddings~\cite{radford2021clip}.
Recent extensions generalize this pairwise formulation to joint multi-way alignment using Gram volumes~\cite{cicchetti2025gram}, hyperbolic variants~\cite{na2026hypergram}, singular values~\cite{liu2025pmrl}, Wasserstein barycenters~\cite{tang2026barybind}, or triangular areas~\cite{cicchetti2025triangle}.
Despite their distinct geometric objectives, these methods operate on time-aggregated representations that are largely invariant to temporal permutations, rendering a shuffled sequence nearly indistinguishable from the original.
Moreover, while prior theoretical analyses characterize failure modes such as dimensional collapse~\cite{jing2022dimensional} and the modality gap~\cite{liang2022gap}, the representational consequences of temporal pooling on event discriminability remain unexamined.

\paragraph{Token-level and temporal alignment.}
Temporally aware multimodal methods vary in how they establish cross-modal correspondence across time.
Multi-vector retrieval~\cite{song2019pvse,reddy2025videocolbert} matches each token to its closest cross-modal counterpart regardless of timestamp, discarding temporal co-occurrence.
Audio--visual synchronization learns embeddings whose similarity peaks at the true offset~\cite{chung2016outoftime,chung2019perfectmatch} or treats time-shifted audio as negatives~\cite{owens2018multisensory,korbar2018avts}, while sequence alignment matches video steps to text~\cite{yang2023tempclr,lin2024norton}.
These formulations assume an unknown offset or ordering and construct negatives via temporal shifts rather than distinct instances recorded under the same condition.
FATE~\cite{guan2026fate} and CAV-MAE Sync~\cite{araujo2025cavmae} evaluate co-occurring frames directly, yet their objectives retain each stream's temporal mean and FATE still relies on within-recording shifts for negatives.
Similarly, MoBind~\cite{nguyen2026mobind} aligns inertial signals with video poses and observes that global pooling obscures fine-grained events, though without characterizing this bottleneck.

\paragraph{Hard negatives and dynamical invariants.}
Beyond temporal alignment, two other directions address instance discrimination and sequence dynamics.
Same-class hard negatives sharpen contrastive learning~\cite{khosla2020supcon,robinson2021hard}, yet grouping by semantic category does not isolate distinct events occurring under identical recording conditions and external drives.
From a dynamical systems perspective, Koopman analyses compare trajectories through their limiting statistics~\cite{mezic2004comparison}, disentangle sequential factors~\cite{berman2023koopman}, and inform recent predictive architectures~\cite{ruizmorales2025koopman}.
However, these formulations focus on the time-invariant component of the dynamics captured by temporal pooling, leaving open how transient fluctuations around that stationary baseline align across modalities.
\section{Points Collapse, Trajectories Scale}
\label{sec:theory}

\subsection{Points Collapse to the Condition}
\label{sec:collapse}

We first explain why temporal pooling collapses event-specific representations into the shared underlying condition.
Consider modalities $m\in\mathcal{M}$ (e.g., haptic H, audio A, visual V) linked solely by a shared clock.
Physical event alignment demands instance-level discrimination, requiring a model to retrieve the exact same event across modalities even among near-identical candidates~\cite{wu2018instance}.
A standard point aligner encodes modality $m$ into per-timestep tokens $h_m(t)\in\mathbb{R}^{d}$, aggregates them over time into a unit vector $z_m$, and optimizes the encoders contrastively via pairwise similarity.

To formalize this, let $\theta$ denote the condition capturing the stationary factors that govern a recording's long-run statistics (e.g., texture, speed, force, and direction).
Over a window where $\theta$ holds, we model the system's evolution as a measure-preserving flow $S_t$ on a state space $X_\theta$ with invariant measure $\mu_\theta$~\cite{walters1982introduction}.
An event corresponds to an initial state $x\in X_\theta$, and modality $m$ observes its trajectory at time $t$ through a square-integrable observable $g_m:X_\theta\to\mathbb{R}^k$, yielding $g_m(S_t x)$.
We call another initial state $x'\in X_\theta$ under the same condition a \emph{twin event}.
The theory thus treats $g_m$ as the per-instant features and the pooled vector as a function of their empirical distribution, as in mean pooling or Deep Sets~\cite{zaheer2017deepsets}.

\begin{figure*}[t]
\centering
\includegraphics[width=\linewidth]{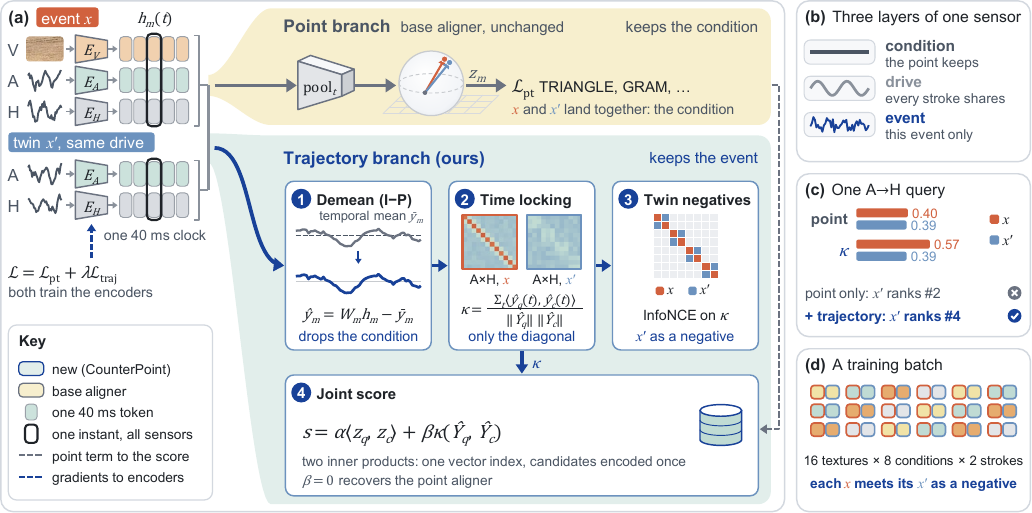}
\caption{\textbf{CounterPoint adds event evidence beside the point.} (a)~Shared encoders feed the point branch and a trajectory branch that removes temporal means and compares the same instants. Twin negatives refine the trajectory, and both scores enter retrieval. (b)~The condition, drive and event layers. (c)~The trajectory lowers the twin's rank in an A$\to$H query. (d)~Paired strokes supply training negatives.}
\label{fig:framework}
\end{figure*}

\begin{theorem}[Condition collapse]
\label{prop:pooling}
Let $S_t$ be ergodic and let $\nu_T^x=\frac{1}{T}\int_0^T\delta_{g_m(S_t x)}\,dt$ be the occupation measure of modality $m$ over a window of length $T$.
For any representation $z_T(x)=\Psi(\nu_T^x)$ with $\Psi$ continuous in the weak topology, $z_T(x)\to\Psi\big((g_m)_\#\mu_\theta\big)$ as $T\to\infty$ for $\mu_\theta$-almost every $x$.
\end{theorem}

\noindent All proofs are in \cref{sec:proofs}. Neither the contrastive loss nor the score geometry affects \cref{prop:pooling}, since they govern the map $\Psi$ rather than its input measure $\nu_T^x$.
Moreover, temporal order is lost even at finite windows $T$. For any permutation $\sigma$ of equal-length time blocks in $[0,T)$,
\begin{equation}
  \nu_T^{x,\sigma}=\nu_T^{x}\ \Longrightarrow\ z_T^{\sigma}(x)=z_T(x).
  \label{eq:shuffle}
\end{equation}
Since an event $x$ and its twin $x'$ converge to the exact same condition-dependent limit in \cref{prop:pooling}, longer windows inevitably collapse their representations together,
\begin{equation}
  \cos\big(z_T(x),\,z_T(x')\big)\ \xrightarrow{\,T\to\infty\,}\ 1 .
  \label{eq:collapse}
\end{equation}

\subsection{The Event Lives in the Dynamics}
\label{sec:layers}

To recover what temporal pooling erases, we introduce two projections that decompose each observable into condition, drive, and event layers.
First, we isolate the condition using the Koopman operator $(U_t g)(x)=g(S_t x)$~\cite{koopman1931hamiltonian,mezic2005spectral}.
Let $P$ project orthogonally onto the $U_t$-invariant functions, with $I$ denoting the identity operator.
By the mean ergodic theorem~\cite{vonneumann1932proof}, $Pg_m$ is the limiting time average of $g_m$, collapsing under ergodicity to the condition-dependent constant $\mathbb{E}_{\mu_\theta}[g_m]$, or the \emph{condition layer}.
Its orthogonal complement $\hat g_m=(I-P)g_m$ forms the \emph{dynamics layer}, which retains all temporal variations that distinguish an event from its twin.

The second projection refines the dynamics layer further.
Strokes following the same path share a common \emph{drive}, namely the external excitation imposed by the path.
We formalize this drive as a dynamical factor $u=\pi(x)$ with distribution $\gamma$, assuming that events are conditionally independent given $u$~\cite{glasner2003joinings}.
For example, a twin event shares the drive when a probe replays the exact path, whereas a free hand movement shares only the condition.
Conditioning on $u$, together with $P$, decomposes every observable into three mutually orthogonal layers,
\begin{equation}
  g_m=\underbrace{Pg_m}_{\text{condition}}+\underbrace{\mathbb{E}[\hat g_m\mid u]}_{\text{drive}}+\underbrace{\hat g_m-\mathbb{E}[\hat g_m\mid u]}_{\text{event}} ,
  \label{eq:threelayers}
\end{equation}
where the \emph{drive layer} captures variations reproduced across all events sharing the drive (e.g., passing over a ridge), and the \emph{event layer} isolates residual fluctuations unique to that specific event (e.g., a micro-slip).

Since the drive $u$ is latent, we measure event agreement directly between dynamics layers and rely on repeated strokes under the same drive to isolate the drive's contribution.
Specifically, we define the \emph{time-locked consistency} between a query modality $q$ on event $x$ and a candidate modality $c$ on event $\tilde x$ as the cosine similarity of their aligned dynamics trajectories over the window,
\begin{equation}
  \kappa(x,\tilde x)=\frac{\int_0^T \hat g_q(S_t x)\cdot\hat g_c(S_t \tilde x)\,dt}
  {\big(\int_0^T\lVert\hat g_q(S_t x)\rVert^2\,dt\big)^{1/2}\big(\int_0^T\lVert\hat g_c(S_t \tilde x)\rVert^2\,dt\big)^{1/2}} ,
  \label{eq:kappa}
\end{equation}
where the window length $T$ is left implicit.

\begin{theorem}[Three levels of agreement]
\label{thm:levels}
Let $\bar\rho_{qc}$ and $\bar\rho^{u}_{qc}$ denote the $L^2(\mu_\theta)$ correlations of the dynamics layers and the drive layers, respectively, both normalized by the $L^2(\mu_\theta)$ norms $\lVert\cdot\rVert_{\mu_\theta}$ of the dynamics layers.
Under standard mixing assumptions on the flow and drive, time-locked consistency converges almost surely to three hierarchical levels,
\begin{equation}
  \kappa(x,x)\to\bar\rho_{qc},\qquad \kappa(x,x')\to\bar\rho^{u}_{qc},\qquad \kappa(x,\tilde x)\to 0,
  \label{eq:levels}
\end{equation}
for the same event, a same-drive twin event, and an event from a different drive or condition.
\end{theorem}

\noindent By ergodic averaging, these time-locked products converge to expectations.
Conditional independence reduces same-drive twin agreement purely to the drive layer, while the consistency of independent recordings vanishes under centering and mixing.
Applying the law of total covariance, the gap between the top two levels isolates the \emph{event gap}, the normalized conditional covariance given the drive,
\begin{equation}
  \bar\rho^{e}_{qc}=\bar\rho_{qc}-\bar\rho^{u}_{qc}=\frac{\int \operatorname{Cov}\big(\hat g_q,\hat g_c\mid u\big)\,d\gamma}{\lVert\hat g_q\rVert_{\mu_\theta}\,\lVert\hat g_c\rVert_{\mu_\theta}} .
  \label{eq:gap}
\end{equation}
This event gap measures the cross-modal alignment of the event layers and equals the exact asymptotic advantage of $\kappa$ for a true match over a same-drive twin.
The event gap thus predicts which modality pairs can separate same-drive twins: it is positive when both event layers respond to the same contact fluctuations, and zero when either modality has no event layer, such as a camera capturing only the path of a replayed stroke.

\paragraph{Remark (the same instants).}
Consider scoring two recordings by taking the maximum $\kappa$ over a finite set of time offsets that includes zero.
If synchronized modalities of a single event achieve maximal alignment at zero offset, applying \cref{thm:levels} offset by offset preserves the same-event limit $\bar\rho_{qc}$ while allowing the same-drive twin limit only to grow, meaning temporal realignment never widens the event gap.

\subsection{Points Stall, Trajectories Converge}
\label{sec:finite}

We now compare how point and trajectory representations scale with window length $T$ in terms of \emph{event accuracy}, defined as the probability of ranking an event $x$ above a same-drive twin $x'$.
For trajectories, we define the score margin $\Delta_T=\kappa(x,x)-\kappa(x,x')$, giving an event accuracy of $\Pr[\Delta_T>0]$.

\begin{theorem}[Points stall, trajectories converge]
\label{prop:stall}
Fix the encoders under standard moment and mixing assumptions~\cite{bradley2005mixing}.
(i)~Unless the cross-modal point fluctuations are perfectly coupled, point event accuracy saturates at a constant $p^\ast<1$.
(ii)~If the event gap is strictly positive, then for all $T$,
\begin{equation}
  \Pr\big[\Delta_T\le 0\big]\le K/T,
  \label{eq:bound}
\end{equation}
with $K\propto(\bar\rho^{e}_{qc})^{-2}$.
\end{theorem}

\noindent This contrast stems from a fundamental difference in perturbation order.
For points, cross-modal alignment pulls the limiting unit vectors together, where cosine similarity peaks and first-order variations vanish.
Because event cues enter only at second order, they decay at the same $1/T$ rate as finite-window noise, locking accuracy at $p^\ast<1$.
For trajectories, event cues enter at first order: the margin $\Delta_T$ converges to the positive event gap $\bar\rho^{e}_{qc}>0$ while its variance decays as $1/T$, directly yielding \cref{eq:bound} with a constant that tightens as the event gap widens.
\section{CounterPoint: From Points to Trajectories}
\label{sec:method}

\paragraph{Demeaned, time-locked trajectories.}
The trajectory branch models the dynamics layer in \cref{sec:theory} by removing the condition from pre-pooling tokens and comparing the residual trajectories at matching frames (\cref{fig:framework}).
A linear projection $W_m\in\mathbb{R}^{k\times d}$ maps each token to $y_m(t)=W_mh_m(t)$, the learned counterpart of $g_m$, in a shared $k$-dimensional space.
The temporal mean $\bar y_m$ estimates the condition layer $Pg_m$, and CounterPoint leaves it to the base aligner.
Subtracting it applies the empirical $I-P$ over time within each recording, keeping the dynamics in the centered trajectory $\hat{Y}_m=(\hat y_m(1),\dots,\hat y_m(T))\in\mathbb{R}^{T\times k}$ with $\hat y_m(t)=y_m(t)-\bar y_m$.
We score two centered trajectories by the time-locked consistency of \cref{eq:kappa}, with sums over the $T$ frames in place of the integrals, and write it $\kappa(\hat{Y}_q,\hat{Y}_c)$.
Consistent with the same-instants remark in \cref{sec:layers}, the score evaluates matching frames $t$ rather than optimizing over temporal offsets as in lag-maximized correlation~\cite{chung2016outoftime}.

\paragraph{Training against twin events.}
The training objective contrasts matching pairs against twin events under the same drive whenever they are present, as these negatives directly resolve the event gap (\cref{eq:levels}).
Writing $\mathcal{L}_{\mathrm{traj}}^{qc}$ for the symmetric InfoNCE loss~\cite{oord2018cpc} over time-locked consistency scores on modality pair $(q,c)$, and $\mathcal{L}_{\mathrm{pt}}$ for the base aligner loss, CounterPoint optimizes their weighted sum across all trained modality pairs $\mathcal{P}$,
\begin{equation}
\mathcal{L}=\mathcal{L}_{\mathrm{pt}}+\frac{\lambda}{|\mathcal{P}|}\sum_{(q,c)\in\mathcal{P}}\mathcal{L}_{\mathrm{traj}}^{qc}.
\label{eq:loss}
\end{equation}

\paragraph{Joint point and trajectory score.}
For a query across modalities $Q\subseteq\mathcal{M}\setminus\{c\}$ and a candidate $n$ in modality $c$, CounterPoint combines the point cosine, which captures the condition, and the trajectory cosine, which separates events inside it, with weights $\alpha_q,\beta_q$,
\begin{equation}
s(Q,n)=\sum_{q\in Q}\Big[\alpha_q\,\langle z_q, z_c^{(n)}\rangle+\beta_q\,\kappa\big(\hat{Y}_q,\hat{Y}_c^{(n)}\big)\Big].
\label{eq:score}
\end{equation}
A single trained model evaluates any query subset.
Setting $\beta_q=0$ recovers the pure point score, while any point aligner exposing time-ordered tokens can adopt the trajectory branch without modification.

\section{Experiments}
\label{sec:exp}

In this section, we evaluate CounterPoint and its theoretical predictions across four questions: whether points collapse to the condition (\textbf{RQ1}), how much CounterPoint improves event retrieval (\textbf{RQ2}), whether trajectories follow our theory (\textbf{RQ3}), and what each component contributes (\textbf{RQ4}).

\begin{table}[!b]
\centering
\caption{\textbf{CounterPoint improves retrieval across aligners and datasets.} R@1 and ACC in \%. Avg is the mean of the three tasks. Superscripts: percentage-point gains over the base aligner. Bold/underline: best/second best within each dataset.}
\label{tab:main}
\footnotesize
\setlength{\tabcolsep}{2.6pt}
\begin{adjustbox}{max width=\linewidth}
\begin{tabular}{llllll}
\toprule
 & AV$\to$H & HV$\to$A & HA$\to$V & ACC & Avg \\
\midrule
\multicolumn{6}{l}{\textit{Cluster}} \\
\midrule
GRAM~\cite{cicchetti2025gram} & 25.7 & 25.5 & 3.2 & 58.2 & 18.1 \\
HyperGRAM~\cite{na2026hypergram} & 25.8 & 25.9 & 3.3 & 58.3 & 18.4 \\
PMRL~\cite{liu2025pmrl} & 24.2 & 24.7 & 3.1 & 56.9 & 17.3 \\
TRIANGLE~\cite{cicchetti2025triangle} & 27.3 & 27.4 & \underline{3.4} & 58.5 & 19.4 \\
\midrule
GRAM\,+\,CP    & 56.3\textsuperscript{\textcolor[HTML]{D1613E}{+30.6}} & 56.5\textsuperscript{\textcolor[HTML]{D1613E}{+30.9}} & 3.3\textsuperscript{\textcolor[HTML]{D1613E}{+0.1}} & 80.8\textsuperscript{\textcolor[HTML]{D1613E}{+22.6}} & 38.7\textsuperscript{\textcolor[HTML]{D1613E}{+20.6}} \\
HyperGRAM\,+\,CP& \underline{57.1}\textsuperscript{\textcolor[HTML]{D1613E}{+31.3}} & \underline{56.8}\textsuperscript{\textcolor[HTML]{D1613E}{+30.9}} & \underline{3.4}\textsuperscript{\textcolor[HTML]{D1613E}{+0.1}} & 80.8\textsuperscript{\textcolor[HTML]{D1613E}{+22.5}} & \underline{39.1}\textsuperscript{\textcolor[HTML]{D1613E}{+20.8}} \\
PMRL\,+\,CP    & 56.0\textsuperscript{\textcolor[HTML]{D1613E}{+31.8}} & 55.9\textsuperscript{\textcolor[HTML]{D1613E}{+31.2}} & \underline{3.4}\textsuperscript{\textcolor[HTML]{D1613E}{+0.3}} & \underline{81.0}\textsuperscript{\textcolor[HTML]{D1613E}{+24.1}} & 38.4\textsuperscript{\textcolor[HTML]{D1613E}{+21.1}} \\
TRIANGLE\,+\,CP& \textbf{58.4}\textsuperscript{\textcolor[HTML]{D1613E}{+31.2}} & \textbf{58.0}\textsuperscript{\textcolor[HTML]{D1613E}{+30.6}} & \textbf{3.5}\textsuperscript{\textcolor[HTML]{D1613E}{+0.0}} & \textbf{81.1}\textsuperscript{\textcolor[HTML]{D1613E}{+22.6}} & \textbf{40.0}\textsuperscript{\textcolor[HTML]{D1613E}{+20.6}} \\
\midrule
\multicolumn{6}{l}{\textit{HAVdb}} \\
\midrule
GRAM & 14.0 & 5.6 & 10.1 & 69.3 & 9.9 \\
HyperGRAM & 12.4 & 5.4 & 7.8 & 65.6 & 8.5 \\
PMRL & 10.5 & 4.6 & 7.4 & 64.9 & 7.5 \\
TRIANGLE & 12.4 & 5.5 & 8.2 & 66.3 & 8.7 \\
\midrule
GRAM\,+\,CP    & \textbf{97.7}\textsuperscript{\textcolor[HTML]{D1613E}{+83.7}} & \textbf{73.5}\textsuperscript{\textcolor[HTML]{D1613E}{+67.9}} & \textbf{94.8}\textsuperscript{\textcolor[HTML]{D1613E}{+84.7}} & \textbf{97.3}\textsuperscript{\textcolor[HTML]{D1613E}{+28.1}} & \textbf{88.7}\textsuperscript{\textcolor[HTML]{D1613E}{+78.8}} \\
HyperGRAM\,+\,CP& \textbf{97.7}\textsuperscript{\textcolor[HTML]{D1613E}{+85.3}} & \underline{73.3}\textsuperscript{\textcolor[HTML]{D1613E}{+67.9}} & \underline{94.0}\textsuperscript{\textcolor[HTML]{D1613E}{+86.2}} & \underline{97.2}\textsuperscript{\textcolor[HTML]{D1613E}{+31.6}} & \underline{88.3}\textsuperscript{\textcolor[HTML]{D1613E}{+79.8}} \\
PMRL\,+\,CP    & 96.9\textsuperscript{\textcolor[HTML]{D1613E}{+86.4}} & 72.4\textsuperscript{\textcolor[HTML]{D1613E}{+67.8}} & 93.8\textsuperscript{\textcolor[HTML]{D1613E}{+86.4}} & 96.6\textsuperscript{\textcolor[HTML]{D1613E}{+31.7}} & 87.7\textsuperscript{\textcolor[HTML]{D1613E}{+80.2}} \\
TRIANGLE\,+\,CP& \underline{97.2}\textsuperscript{\textcolor[HTML]{D1613E}{+84.8}} & 71.5\textsuperscript{\textcolor[HTML]{D1613E}{+66.0}} & 93.7\textsuperscript{\textcolor[HTML]{D1613E}{+85.5}} & 96.8\textsuperscript{\textcolor[HTML]{D1613E}{+30.5}} & 87.4\textsuperscript{\textcolor[HTML]{D1613E}{+78.8}} \\
\bottomrule
\end{tabular}
\end{adjustbox}
\end{table}

\begin{figure}[t]
\centering
\includegraphics[width=\linewidth]{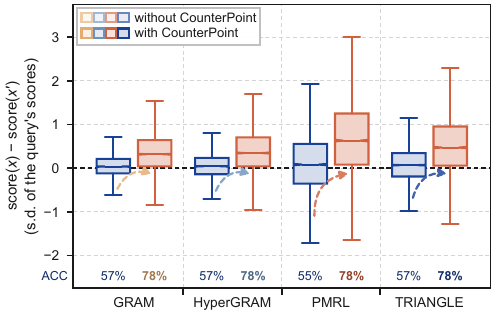}
\caption{\textbf{CounterPoint shifts twin margins above zero.} A$\to$H on one Cluster fold. Boxes show the score margin of $x$ over its twin in standard deviations of candidate scores per query, where positive values favor $x$. Blue and orange denote models without and with CounterPoint, with event accuracy listed at the bottom.}
\label{fig:margindist}
\end{figure}

\begin{table}[t]
\centering
\caption{\textbf{Gains generalize to unseen conditions.} Cluster Avg R@1 (\%) with categories, speeds, or directions held out. Superscripts: percentage-point gains. Bold: best. Held-out textures: \cref{tab:main}.}
\label{tab:heldout}
\footnotesize
\setlength{\tabcolsep}{4pt}
\begin{adjustbox}{max width=\linewidth}
\begin{tabular}{llll}
\toprule
Held out & Category & Speed & Direction \\
\midrule
GRAM         & 17.0 & 8.5 & 2.0 \\
HyperGRAM    & 16.7 & 8.7 & 2.0 \\
PMRL         & 15.9 & 8.6 & 2.0 \\
TRIANGLE     & 17.4 & 9.4 & 2.1 \\
\midrule
GRAM\,+\,CP     & 37.3\textsuperscript{\textcolor[HTML]{D1613E}{+20.3}} & 27.2\textsuperscript{\textcolor[HTML]{D1613E}{+18.7}} & 10.3\textsuperscript{\textcolor[HTML]{D1613E}{+8.3}} \\
HyperGRAM\,+\,CP& 37.4\textsuperscript{\textcolor[HTML]{D1613E}{+20.7}} & 27.7\textsuperscript{\textcolor[HTML]{D1613E}{+19.0}} & 10.5\textsuperscript{\textcolor[HTML]{D1613E}{+8.5}} \\
PMRL\,+\,CP     & 37.0\textsuperscript{\textcolor[HTML]{D1613E}{+21.1}} & \textbf{29.8}\textsuperscript{\textcolor[HTML]{D1613E}{+21.2}} & 9.9\textsuperscript{\textcolor[HTML]{D1613E}{+7.9}} \\
TRIANGLE\,+\,CP & \textbf{38.1}\textsuperscript{\textcolor[HTML]{D1613E}{+20.6}} & 29.7\textsuperscript{\textcolor[HTML]{D1613E}{+20.4}} & \textbf{10.8}\textsuperscript{\textcolor[HTML]{D1613E}{+8.8}} \\
\bottomrule
\end{tabular}
\end{adjustbox}
\end{table}

\subsection{Experimental Setup}
\label{sec:exp_setup}

\paragraph{Datasets.}
\looseness=-1
Two texture-sliding datasets instantiate the two regimes of our three-level agreement theorem (\cref{thm:levels}).
\emph{Cluster Haptic Texture}~\cite{eguchi2024cluster} tests the shared-drive regime: a robot replays probe strokes across 118 textures (9 categories) under 8 directions, 5 speeds, and 2 forces, so twin events share both the condition and the drive.
Here, A is audio, H is haptic vibration, and V is the texture scan along the logged path, which varies only with the drive.
We hold out textures, categories, speeds, or directions (5, 9, 2, and 1 folds).
\emph{HAVdb}~\cite{devillard2025havdb} tests the shared-condition regime: 10 subjects slide a hand across 10 textures, where A is audio, H combines vibration and force, and V tracks the fingertip.
A twin event, the same window from another slide by the same subject on the same texture, shares only the condition.
We hold out subjects (5 folds).
All splits use 3 seeds with 1\,s windows on Cluster and 16\,s windows on HAVdb unless stated otherwise (\cref{sec:data}).

\paragraph{Metrics and baselines.}
Synchronized recordings define ground-truth correspondence, and retrieval ranks candidates by representation similarity.
We report R@1 over all test events, twin events included (\%, ties counted as misses), for AV$\to$H, HV$\to$A, and HA$\to$V.
Avg is their mean, while A--H averages the first two.
To measure within-condition discrimination, event accuracy (ACC) tracks the fraction of test events scored above their twin event, averaged over the three tasks (chance 50\%, \cref{sec:finite}).
We pair CounterPoint (+\,CP) with four point aligners of distinct geometries: GRAM~\cite{cicchetti2025gram}, HyperGRAM~\cite{na2026hypergram}, PMRL~\cite{liu2025pmrl}, and TRIANGLE~\cite{cicchetti2025triangle}, all trained with their official losses on shared encoders emitting one token every 40\,ms.
CP shares these encoders and the point loss, adding a trajectory branch (0.4\% parameters) with $k=64$, $\lambda=0.5$, and joint point and trajectory weights (\cref{eq:score}) fitted on a development split (\cref{sec:impl}).
Its heads omit bias terms, as subtracting the temporal mean cancels any constant offset, and candidates are encoded once into a standard vector index, since both score terms depend only on candidate vectors.
On Cluster, we also replace the trajectory score in TRIANGLE\,+\,CP with the scores of prior temporal alignment methods.
We test the instantaneous cosine of CAV-MAE Sync~\cite{araujo2025cavmae}, which retains the temporal mean, both alone and with the FATE offset loss~\cite{guan2026fate} scaled to 1\,s.
On demeaned trajectories, we evaluate the maximum cosine over shifts up to 160\,ms~\cite{chung2016outoftime}, soft-DTW~\cite{cuturi2017softdtw,yang2023tempclr}, and~MaxSim~\cite{khattab2020colbert,reddy2025videocolbert}.

\subsection{Point Aligners Stop at the Condition (RQ1)}
\label{sec:exp_collapse}

\begin{figure}[t]
\centering
\includegraphics[width=\linewidth]{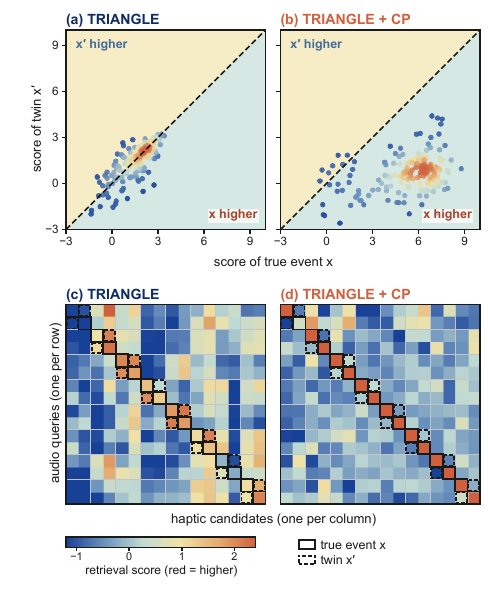}
\caption{\textbf{CounterPoint separates twin events that point representations conflate.} A$\to$H on HAVdb for TRIANGLE without (a,\,c) and with (b,\,d) CounterPoint. (a,\,b)~Each dot marks one query, with points below the diagonal indicating that $x$ outscores its twin $x'$. (c,\,d)~Score matrices for one subject, where solid and dashed boxes mark $x$ and $x'$, and red indicates higher scores.}
\label{fig:visual}
\end{figure}

\looseness=-1
Point aligners capture the condition level but fail to distinguish individual events, confirming condition collapse (\cref{prop:pooling}).
On Cluster, all four geometries fall within three percentage points in A--H R@1 and stay below 59\% event accuracy (\cref{tab:main}), with median twin margins of only 0.04 to 0.08 standard deviations versus 0.32 to 0.63 under CounterPoint (\cref{fig:margindist}).
On HAVdb, TRIANGLE assigns nearly identical scores to an event and its twin, scoring the entire $2\times2$ texture block high (\cref{fig:visual}a,\,c), whereas CounterPoint breaks this conflation and isolates a single cell per event (\cref{fig:visual}b,\,d).

\paragraph{Trained point encoders discard temporal order.}
\looseness=-1
Despite positional encodings, TRIANGLE exhibits permutation invariance (\cref{eq:shuffle}).
Shuffling 40\,ms input blocks, either jointly or independently, leaves its audio and haptic embeddings at 1.000 cosine similarity to the unshuffled originals, far above their similarity to twin events (0.909 and 0.874).
Discriminative event cues reside in the temporal order discarded by pooling, which point geometry alone cannot recover.

\begin{figure}[t]
\centering
\includegraphics[width=\linewidth]{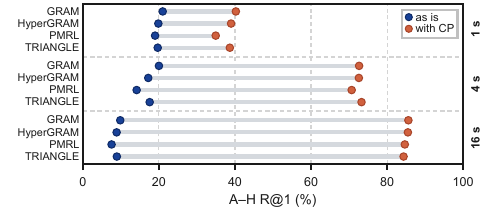}
\caption{\textbf{Points collapse while trajectories scale.} HAVdb A--H R@1 for GRAM, HyperGRAM, PMRL, and TRIANGLE without (blue) and with (orange) CounterPoint across 1, 4, and 16\,s training windows at a fixed candidate count.}
\label{fig:scaling}
\end{figure}

\begin{table}[t]
\centering
\caption{\textbf{Demeaned, time-locked trajectories outperform alternative temporal alignment methods.} Cluster results (\%). Alternatives replace the trajectory score of TRIANGLE\,+\,CP, while TRIANGLE is point-only. Below the middle rule: demeaned trajectories. A--H averages AV$\to$H and HV$\to$A.}
\label{tab:temporal}
\footnotesize
\setlength{\tabcolsep}{3.5pt}
\begin{adjustbox}{max width=\linewidth}
\begin{tabular}{lllll}
\toprule
Score & Instants matched & A--H & Avg & ACC \\
\midrule
TRIANGLE (point) & pooled & 27.4 & 19.4 & 58.5 \\
No $I-P$ (CAV-MAE Sync~\cite{araujo2025cavmae}) & same & 29.0 & 20.4 & 59.6 \\
FATE~\cite{guan2026fate} & same & 27.8 & 19.6 & 58.4 \\
\midrule
MaxSim~\cite{khattab2020colbert,reddy2025videocolbert} & any & 27.0 & 19.1 & 57.7 \\
Soft-DTW~\cite{cuturi2017softdtw,yang2023tempclr} & monotone warp & 37.3 & 26.0 & 65.1 \\
Best shift~\cite{chung2016outoftime} & shift $\le$\,160\,ms & 47.6 & 32.9 & 74.1 \\
\rowcolor[HTML]{E3EFEC}
CounterPoint & same & \textbf{58.2} & \textbf{40.0} & \textbf{81.1} \\
\bottomrule
\end{tabular}
\end{adjustbox}
\end{table}

\paragraph{Points collapse while trajectories scale.}
\looseness=-1 As window duration increases from 1 to 16\,s on HAVdb at a fixed candidate count, every point aligner loses over half its A--H R@1, while TRIANGLE\,+\,CP climbs from 38.6\% to 84.3\% (\cref{fig:scaling}), confirming condition collapse (\cref{eq:collapse}).
Even at 1\,s, the trajectory branch nearly doubles point retrieval accuracy.

\subsection{CounterPoint Improves Event Retrieval (RQ2)}
\label{sec:exp_main}

\paragraph{Consistent gains across aligners.}
CounterPoint boosts event retrieval across every aligner and dataset.
While leaving the base loss and geometry intact, it more than doubles A--H R@1 on Cluster from at most 27.4\% to at least 56.0\%, lifting event accuracy from below 59\% to roughly 81\% (\cref{tab:main}).
On HAVdb, Avg R@1 rises from at most 9.9\% to at least 87.4\%, and event accuracy reaches about 97\%.
This larger gain on HAVdb reflects that free-hand twins share only the condition while following independent dynamics.

\begin{figure}[t]
\centering
\includegraphics[width=\linewidth]{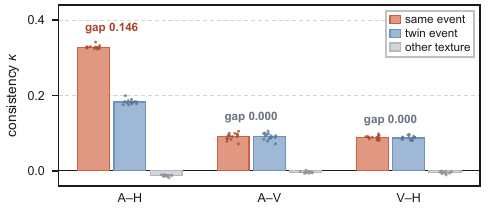}
\caption{\textbf{Only audio and haptics separate shared-drive twins.} Cluster consistency for the same event $\kappa(x,x)$, its twin $\kappa(x,x')$, and another texture $\kappa(x,\tilde x)$. Bars show means and dots show individual runs. The first two levels separate for A--H but coincide for path-only visual pairs.}
\label{fig:mechanism}
\end{figure}
\begin{figure}[t]
\centering
\includegraphics[width=\linewidth]{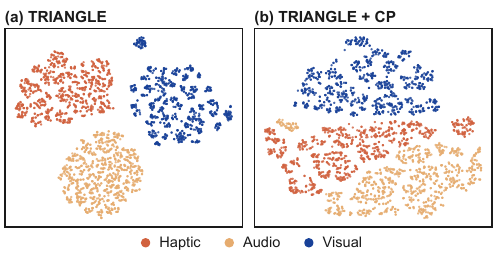}
\caption{\textbf{Audio and haptics align while path-only vision stays apart.} t-SNE~\cite{vandermaaten2008tsne} of 1,000 Cluster test events for (a)~TRIANGLE points and (b)~TRIANGLE\,+\,CP unit points concatenated with unit flattened trajectories.}
\label{fig:tsnemod}
\end{figure}
\paragraph{Gains transfer to unseen conditions.}
These improvements generalize beyond seen conditions.
On Cluster, every aligner gains 8 to 21 percentage points in Avg R@1 across held-out splits (\cref{tab:heldout}).
Specifically, TRIANGLE\,+\,CP roughly doubles the R@1 of TRIANGLE on unseen textures and categories, triples it on speeds, and increases it fivefold on directions from 2.1\% to 10.8\%.

\paragraph{Alternative temporal alignment methods fall short on shared-drive twins.}
\looseness=-1 Standard temporal aggregation struggles when twins share the same drive.
On Cluster, MaxSim stays near the point baseline, while soft-DTW and best shift reach only 37.3\% and 47.6\% A--H R@1, below the 58.2\% of CounterPoint and with lower event accuracy (\cref{tab:temporal}).
Similarly, CAV-MAE Sync without $I-P$ (29.0\%) and the FATE offset loss (27.8\%) remain close to the point baseline.

\paragraph{Minimal overhead and preserved semantics.}
The trajectory branch adds only 0.7\% to peak training memory, while querying 1,000 events against $10^6$ candidates takes 145\,ms versus 34\,ms for points alone.
Crucially, CounterPoint leaves the underlying point representation intact.
Linear probes for category, texture, direction, speed, and force show no statistically significant change (smallest Holm-corrected $p=0.18$), and development event accuracy remains stable between 80.4\% and 83.1\% across $\lambda\in\{0.1,0.25,1.0\}$ and $k\in\{16,32,128\}$.

\begin{table}[t]
\centering
\caption{\textbf{Each component contributes a distinct cue.} TRIANGLE\,+\,CP on Cluster (\%). Time-lock rows shift or permute candidate trajectories at evaluation, while remaining rows omit twin negatives or the point score. Bold denotes the best result.}
\label{tab:ablation}
\footnotesize
\setlength{\tabcolsep}{3.5pt}
\begin{adjustbox}{max width=\linewidth}
\begin{tabular}{llll}
\toprule
Variant & A--H & Avg & ACC \\
\midrule
TRIANGLE & 27.4 & 19.4 & 58.5 \\
\rowcolor[HTML]{E3EFEC}
TRIANGLE\,+\,CP & \textbf{58.2} & \textbf{40.0} & 81.1 \\
\midrule
\multicolumn{4}{l}{\textit{Without time lock}} \\
\quad candidates shifted by 40\,ms (one token) & 32.3 & 22.7 & 60.7 \\
\quad candidates shifted by 160\,ms & 29.0 & 20.4 & 59.1 \\
\quad candidate tokens permuted & 29.0 & 20.5 & 59.2 \\
\midrule
\multicolumn{4}{l}{\textit{Without twin negatives}} \\
\quad no twin event in any training batch & 55.7 & 38.3 & 78.8 \\
\midrule
\multicolumn{4}{l}{\textit{Without the point score}} \\
\quad demeaned, trajectory-only & 45.8 & 31.2 & \textbf{81.7} \\
\bottomrule
\end{tabular}
\end{adjustbox}
\end{table}

\begin{figure}[t]
\centering
\includegraphics[width=\linewidth]{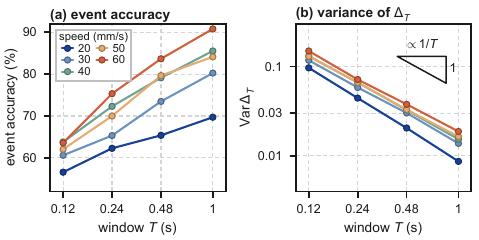}
\caption{\textbf{Longer windows raise accuracy and reduce margin variance.} Fixed 1\,s Cluster encoder evaluated on central tokens across five speeds, showing (a)~event accuracy versus window $T$ and (b)~variance of the event versus twin margin $\Delta_T$, where the triangle marks the $1/T$ reference slope.}
\label{fig:law}
\end{figure}

\subsection{Trajectories Follow the Theory (RQ3)}
\label{sec:exp_levels}

\paragraph{Trajectory gains track the empirical event gap.}
We define the empirical event gap on held-out data as the mean same-event consistency $\kappa(x,x)$ minus the mean twin consistency $\kappa(x,x')$.
On Cluster, learned consistency follows our three levels of agreement (\cref{thm:levels}, \cref{fig:mechanism}), where audio and haptics show an event gap of 0.146 while the path-only visual control remains near zero.
Retrieval gains directly mirror this gap (\cref{tab:main}).
On Cluster, CounterPoint lifts AV$\to$H and HV$\to$A by over 30 percentage points but HA$\to$V by at most 0.3, whereas on HAVdb it boosts HA$\to$V by at least 84.7 percentage points by capturing unconstrained motion dynamics.
The learned geometry reflects this split, as TRIANGLE audio and haptic embeddings converge under CounterPoint while the visual scan stays distinct (\cref{fig:tsnemod}).

\paragraph{Empirical margins follow the theoretical bound.}
Fixed-encoder margins adhere closely to our trajectory error bound (\cref{eq:bound}).
On central tokens of the fixed 1\,s Cluster encoder from 0.12 to 1\,s across speeds, event accuracy rises (\cref{fig:law}a), while the mean margin shifts by at most 0.013 and its variance decays as $1/T$ (\cref{fig:law}b).
Faster strokes sustain a wider margin, rising from 0.05 at 20\,mm/s to 0.15 at 60\,mm/s with higher accuracy, consistent with $K\propto(\bar\rho^{e}_{qc})^{-2}$.

\subsection{Component Ablations (RQ4)}
\label{sec:exp_ablation}
Each component plays a distinct role, as demeaning and time locking preserve event evidence, twin negatives sharpen it, and the point score anchors cross-condition retrieval (\cref{tab:temporal,tab:ablation}).
Disrupting temporal alignment at test time collapses the gain, where a single 40\,ms token shift costs 26 percentage points and a 160\,ms shift or permutation erases nearly all improvement.
Twin negatives further sharpen the trajectory, though CounterPoint still achieves 55.7\% A--H R@1 without them.
Finally, the trajectory branch alone achieves the highest event accuracy, while combining it with the point score lifts A--H R@1 from 45.8\% to 58.2\%.

\section{Conclusion}
\label{sec:conclusion}

\looseness=-1
In this work, we revisit multimodal alignment through the lens of physical interaction and show that compressing temporal recordings into pooled \emph{points} induces \emph{condition collapse}, causing representations to converge to a stationary limit regardless of training loss or embedding geometry. Temporal pooling averages out transient contact dynamics, whereas unpooled zero-mean \emph{trajectories} accumulate synchronous cross-modal fluctuations over time to capture the \emph{event gap}.
Guided by this analysis, we introduce CounterPoint, a principled trajectory alignment module that projects out stationary condition components to measure time-locked~cross-modal covariance.
Across SOTA aligners on controlled and free-hand interactions, CounterPoint outperforms prior temporal alignment methods, showing that grounding Physical AI requires moving beyond static point embeddings to preserve the coupled dynamics of trajectories.
\FloatBarrier
{
    \small
    \bibliographystyle{ieeenat_fullname}
    \bibliography{main}
}

\startsupplement
\addtocontents{toc}{\protect\setcounter{tocdepth}{2}}

\begin{strip}
The main paper addresses the alignment of multimodal observations of a single physical event for Physical AI. It advances three claims: pooled points collapse to the underlying condition, the event resides in the dynamics, and time-locked trajectories recover it.
This supplement supports each claim in turn and can be read as a self-contained document. It restates the problem setting, the three theorems, and the method before building upon them. Lettered sections and the figures, tables and equations numbered with the prefix S belong to this supplement, while all other numbers, such as \cref{tab:main}, refer to the main paper.
\cref{sec:proofs} proves the three theorems, and \cref{sec:scaling} examines the properties of the recordings on which their assumptions rest and tests the trajectory bound.
\cref{sec:data,sec:impl} describe the two datasets and every detail needed to reproduce the results.
\cref{sec:more} extends the four research questions of the experiments, \cref{sec:related-ext} places the work in its wider literature, and \cref{sec:impact} discusses scope, outlook and broader impact.

\setcounter{tocdepth}{2}
\tableofcontents
\end{strip}

%% =====================================================================
\section{Proofs}
\label{sec:proofs}

This section begins by restating the problem setting (\cref{sec:setting}), followed by the formal statements and proofs of the three theorems in the exact order and numbering of the main paper: condition collapse (\cref{sec:proof-collapse}), the three levels of agreement alongside the same-instants corollary (\cref{sec:proof-levels}), and point stalling versus trajectory convergence, including the Gaussian fluctuations of the trajectory margin (\cref{sec:proof-stall}).
Two short results then connect these theorems to our method. Twin events identify the event gap (\cref{sec:proof-repeats}), and the demeaned cosine that CounterPoint computes converges to the levels predicted by the theory (\cref{sec:proof-cosine}).
\subsection{Settings and notations}
\label{sec:setting}

\paragraph{Events and observations.}
Consider a set of modalities $m\in\mathcal{M}$ (such as haptic H, audio A, and visual V) synchronized solely by a shared clock.
The results below hold for any pair of modalities without restricting what the sensors measure.
Let $\theta$ denote the \emph{condition} capturing the stationary factors that govern a recording's long-run statistics, including texture, speed, force, and direction.
Over a window where $\theta$ holds, we model the system's evolution as a measure-preserving flow $S_t$ on a state space $X_\theta$ with invariant measure $\mu_\theta$~\cite{walters1982introduction}.
An \emph{event} corresponds to an initial state $x\in X_\theta$, and another initial state $x'\in X_\theta$ under the same condition is called a \emph{twin event}.
Modality $m$ observes the trajectory at time $t$ through a square-integrable observable $g_m\in L^2(\mu_\theta,\mathbb{R}^k)$, producing the signal $g_m(S_tx)$.

\looseness=-1 Standard point aligners encode modality $m$ into per-timestep tokens $h_m(t)\in\mathbb{R}^{d}$, aggregate them over time into a unit vector $z_m$ (the \emph{point}), and train encoders contrastively via pairwise similarity.
In our framework, $g_m$ represents these per-instant features, and temporal pooling acts on their empirical distribution over a window of length $T$, as in mean pooling or Deep Sets~\cite{zaheer2017deepsets}.
Specifically, the pooled representation is $z_T(x)=\Psi(\nu_T^x)$, where $\Psi$ is any map from distributions to representations and $\nu_T^x=\frac1T\int_0^T\delta_{g_m(S_tx)}\,dt$ is the occupation measure of modality $m$.

\paragraph{The three layers.}
Let $(U_tg)(x)=g(S_tx)$ be the Koopman operator, $P$ the orthogonal projection of $L^2(\mu_\theta)$ onto the $U_t$-invariant functions, and $I$ the identity operator.
By the mean ergodic theorem~\cite{vonneumann1932proof}, $Pg_m$ is the limiting time average of $g_m$. Under ergodicity, it collapses to the condition-dependent constant $\mathbb{E}_{\mu_\theta}[g_m]$, which we call the \emph{condition layer}.
Its orthogonal complement $\hat g_m=(I-P)g_m$ forms the \emph{dynamics layer}, representing the temporal variations that distinguish an event from its twin.

Strokes following the same path share a common \emph{drive}, defined as the external excitation imposed by the path.
We formalize this drive as a dynamical factor $u=\pi(x)$ with distribution $\gamma$, assuming that events are conditionally independent given $u$~\cite{glasner2003joinings}.
For instance, a twin event shares the drive when a probe replays the exact same path, whereas a freehand movement shares only the condition.
Writing $\mathbb{E}[f\mid u]$ for the conditional expectation of $f$ given $u$ under $\mu_\theta$, we decompose each observable into three mutually orthogonal layers in $L^2(\mu_\theta)$:
\begin{equation}
  g_m=\underbrace{Pg_m}_{\text{condition}}+\underbrace{\mathbb{E}[\hat g_m\mid u]}_{\text{drive}}+\underbrace{\hat g_m-\mathbb{E}[\hat g_m\mid u]}_{\text{event}} .
  \label{supp:eq-threelayers}
\end{equation}
Here, the \emph{drive layer} isolates variations reproduced across events sharing the drive (such as passing over a ridge), while the \emph{event layer} captures residual fluctuations specific to a single event (such as a micro-slip).
To verify orthogonality, note that the condition layer is constant while the remaining two layers have zero mean:
\begin{equation}
\begin{aligned}
  \int\hat g_m\,d\mu_\theta&=\mathbb{E}_{\mu_\theta}[g_m]-Pg_m=0,\\
  \int\mathbb{E}[\hat g_m\mid u]\,d\gamma&=\int\hat g_m\,d\mu_\theta=0 .
\end{aligned}
\end{equation}
Moreover, because the drive layer is a function of $u$, conditioning on $u$ inside the inner product shows that the drive and event layers are orthogonal:
\begin{equation}
\begin{aligned}
  &\big\langle\mathbb{E}[\hat g_m\mid u],\ \hat g_m-\mathbb{E}[\hat g_m\mid u]\big\rangle\\
  &\quad=\int\mathbb{E}[\hat g_m\mid u]\cdot\big(\mathbb{E}[\hat g_m\mid u]-\mathbb{E}[\hat g_m\mid u]\big)\,d\gamma=0 .
\end{aligned}
\end{equation}

\paragraph{Time-locked consistency.}
We define the \emph{time-locked consistency} between a query modality $q$ on event $x$ and a candidate modality $c$ on event $\tilde x$ as the cosine similarity between their aligned dynamics trajectories over the window:
\begin{equation}
  \kappa(x,\tilde x)=\frac{\int_0^T \hat g_q(S_t x)\cdot\hat g_c(S_t \tilde x)\,dt}
  {\big(\int_0^T\lVert\hat g_q(S_t x)\rVert^2dt\big)^{1/2}\big(\int_0^T\lVert\hat g_c(S_t \tilde x)\rVert^2dt\big)^{1/2}} .
  \label{supp:eq-kappa}
\end{equation}

\looseness=-1 \paragraph{Notations.}
\looseness=-1 For a measure $\mu$, $L^p(\mu)$ denotes the space of functions $f$ satisfying $\int\lVert f\rVert^p\,d\mu<\infty$.
We write $\langle f,h\rangle=\int f\cdot h\,d\mu_\theta$ and $\lVert f\rVert^2=\langle f,f\rangle$ for the inner product and norm on $L^2(\mu_\theta)$, making the measure explicit as $\lVert f\rVert_{\mu_\theta}$ when needed and dropping $\theta$ when only one condition is involved.
In Euclidean space $\mathbb{R}^k$, $\cdot$ denotes the dot product and $\lVert\cdot\rVert$ the Euclidean norm.
Probabilities $\Pr$, expectations $\mathbb{E}$, and variances $\operatorname{Var}$ are taken over events drawn from $\mu_\theta$, while $\Rightarrow$ denotes weak convergence of measures (and convergence in distribution for random vectors).
A bar over a correlation $\rho$ indicates normalization by the norms of the dynamics layers, as in $\bar\rho_{qc}=\langle\hat g_q,\hat g_c\rangle/(\lVert\hat g_q\rVert\,\lVert\hat g_c\rVert)$.
Finally, $g_\#\mu$ denotes the pushforward of $\mu$ under $g$, giving the distribution of $g(x)$ for $x\sim\mu$, and $\delta_v$ denotes the Dirac measure at $v$, respectively.

\subsection{Theorem~\ref{supp:thm-collapse}: condition collapse}
\label{sec:proof-collapse}

\begin{theorem}[Condition collapse]
\label{supp:thm-collapse}\leavevmode\ignorespaces
Let $S_t$ be ergodic and let $\nu_T^x=\frac{1}{T}\int_0^T\delta_{g_m(S_t x)}\,dt$ be the occupation measure of modality $m$ over a window of length $T$.
For any representation $z_T(x)=\Psi(\nu_T^x)$ with $\Psi$ continuous in the weak topology, $z_T(x)\to\Psi\big((g_m)_\#\mu_\theta\big)$ as $T\to\infty$ for $\mu_\theta$-almost every $x$.
\end{theorem}

\noindent Our proof relies on the following two conditions.

\paragraph{Assumptions.}
\begin{itemize}
\item[(P1)] \emph{Ergodicity.} The flow $S_t$ is ergodic with respect to $\mu_\theta$.
\item[(P2)] \emph{Continuity.} There exists a countable family $\mathcal F$ of functions $f$ on $\mathbb{R}^k$, satisfying $f\circ g_m\in L^1(\mu_\theta)$, such that $\Psi(\nu_n)\to\Psi(\nu)$ whenever $\int f\,d\nu_n\to\int f\,d\nu$ for every $f\in\mathcal F$.
\end{itemize}
In particular, continuity in the weak topology satisfies (P2) via a countable, convergence-determining family of bounded Lipschitz functions on $\mathbb{R}^k$. Likewise, moment-based pooling (such as mean pooling) satisfies (P2) using its coordinate feature functions.

\noindent This formulation reveals two immediate consequences for pooled point representations.
First, temporal order is discarded at every window length. For any permutation $\sigma$ of equal-length time blocks in $[0,T)$,
\begin{equation}
  \nu_T^{x,\sigma}=\nu_T^{x}\ \Longrightarrow\ z_T^{\sigma}(x)=z_T(x),
  \label{supp:eq-shuffle}
\end{equation}
where the superscript $\sigma$ denotes the occupation measure and representation of the block-permuted signal.
Second, an event and its twin converge to the same condition-dependent limit, causing their representations to align as the window grows:
\begin{equation}
  \cos\big(z_T(x),\,z_T(x')\big)\ \xrightarrow{\,T\to\infty\,}\ 1 ,
  \label{supp:eq-collapse}
\end{equation}
where $\cos$ denotes cosine similarity, defined as the inner product of two vectors divided by the product of their Euclidean norms.

\begin{proof}
\emph{Step 1 (ergodic limit for a single test function).}
Fix $f\in\mathcal F$. By definition of the occupation measure,
\begin{equation}
  \int f\,d\nu_T^x=\frac1T\int_0^T f\big(g_m(S_tx)\big)\,dt .
  \label{eq:occ-test}
\end{equation}
Since $f\circ g_m\in L^1(\mu)$ by (P2) and $S_t$ is ergodic by (P1), Birkhoff's ergodic theorem~\cite{birkhoff1931proof} yields a $\mu$-null set $N_f$ such that, for every $x\notin N_f$,
\begin{equation}
\begin{aligned}
  \frac1T\int_0^T f\big(g_m(S_tx)\big)\,dt\ &\longrightarrow\ \int f\circ g_m\,d\mu\\
  &=\int f\,d\big((g_m)_\#\mu\big),
\end{aligned}
  \label{eq:birkhoff-test}
\end{equation}
where the equality follows from the change-of-variables formula for pushforward measures.

\emph{Step 2 (simultaneous convergence across $\mathcal F$).}
Because $\mathcal F$ is countable, the union $N=\bigcup_{f\in\mathcal F}N_f$ is a countable union of null sets and is therefore $\mu$-null.
For all $x\notin N$, combining \cref{eq:occ-test,eq:birkhoff-test} gives
\begin{equation}
  \int f\,d\nu_T^x\ \longrightarrow\ \int f\,d\big((g_m)_\#\mu\big)\qquad\text{for every } f\in\mathcal F .
\end{equation}

\emph{Step 3 (convergence of the representation).}
Applying the continuity property (P2) for every $x\notin N$ yields
\begin{equation}
  z_T(x)=\Psi(\nu_T^x)\ \longrightarrow\ \Psi\big((g_m)_\#\mu\big) . \qedhere
\end{equation}
\end{proof}

\noindent\textbf{Proof of shuffle invariance (\cref{supp:eq-shuffle}).}
Partition $[0,T)$ into $L$ equal intervals $B_i=[(i-1)T/L,\,iT/L)$ and let $\nu_{B_i}^x=\frac LT\int_{B_i}\delta_{g_m(S_tx)}\,dt$ denote the occupation measure on block $i$.
Averaging over all blocks recovers the full-window occupation measure:
\begin{equation}
  \nu_T^x=\frac1T\sum_{i=1}^L\int_{B_i}\delta_{g_m(S_tx)}\,dt=\frac1L\sum_{i=1}^L\nu_{B_i}^x .
\end{equation}
Applying the permutation $\sigma$ places block $\sigma(i)$ into the $i$-th time slot, yielding the occupation measure
\begin{equation}
  \nu_T^{x,\sigma}=\frac1L\sum_{i=1}^L\nu_{B_{\sigma(i)}}^x=\frac1L\sum_{i=1}^L\nu_{B_i}^x=\nu_T^x ,
\end{equation}
since permutation merely reorders the sum.
Applying $\Psi$ to both sides gives $z_T^\sigma(x)=z_T(x)$.

\noindent\textbf{Proof of condition collapse (\cref{supp:eq-collapse}).}
Let $z^\ast=\Psi\big((g_m)_\#\mu_\theta\big)$, which depends solely on the condition $\theta$.
By \cref{supp:thm-collapse}, for $\mu_\theta$-almost every event $x$ and twin event $x'$,
\begin{equation}
  z_T(x)\to z^\ast\qquad\text{and}\qquad z_T(x')\to z^\ast .
\end{equation}
Points are unit vectors (\cref{sec:setting}), so their limit $z^\ast$ is a unit vector as well. Cosine similarity is therefore continuous at $(z^\ast,z^\ast)$, which implies
\begin{equation}
  \cos\big(z_T(x),z_T(x')\big)\ \longrightarrow\ \cos(z^\ast,z^\ast)=1 .
\end{equation}

\subsection{Theorem~\ref{supp:thm-levels}: three levels of agreement}
\label{sec:proof-levels}

Let $\pi:X\to\mathcal U$ be the factor map onto the drive system $(\mathcal U,\gamma,R_t)$, where $R_t$ denotes the drive flow satisfying $\pi(S_tx)=R_t\pi(x)$.
A pair of same-drive twin events follows the relative product law $\Lambda=\mu\times_{\mathcal U}\mu$, defined by
\begin{equation}
  \int f(x)\,h(x')\,d\Lambda(x,x')=\int_{\mathcal U} \mathbb{E}[f\mid u]\,\mathbb{E}[h\mid u]\,d\gamma(u).
  \label{eq:relprod}
\end{equation}
Relative weak mixing of $S$ over $\mathcal U$ means that the product flow $S\times S$, which evolves a pair as $(S_tx,S_tx')$, is ergodic with respect to $\Lambda$~\cite{furstenberg1977ergodic}.

\begin{theorem}[Three levels of agreement]
\label{supp:thm-levels}\leavevmode\ignorespaces
Let $\bar\rho_{qc}$ and $\bar\rho^{u}_{qc}$ denote the $L^2(\mu_\theta)$ correlations of the dynamics layers and the drive layers, respectively, both normalized by the $L^2(\mu_\theta)$ norms $\lVert\cdot\rVert_{\mu_\theta}$ of the dynamics layers.
Under standard mixing assumptions on the flow and drive, time-locked consistency converges almost surely to three hierarchical levels,
\begin{equation}
  \kappa(x,x)\to\bar\rho_{qc},\qquad \kappa(x,x')\to\bar\rho^{u}_{qc},\qquad \kappa(x,\tilde x)\to 0,
  \label{supp:eq-levels}
\end{equation}
for the same event, a same-drive twin event, and an event from a different drive or condition.
\end{theorem}

\noindent Our proof relies on the following conditions, which make the mixing assumptions on the flow and drive precise.

\paragraph{Assumptions.}
\begin{itemize}
\item[(M1)] \emph{Mixing relative to the drive.} The relative product $\Lambda$ is ergodic under the product flow $S\times S$, meaning that $S_t$ is weakly mixing relative to the drive.
\item[(M2)] \emph{No periodic component.} Every coordinate of the query dynamics layer $\hat g_q$ is orthogonal in $L^2(\mu_\theta)$ to every eigenfunction $e$ of the flow, meaning every complex function satisfying $e(S_tx)=\mathrm{e}^{\mathrm{i}\omega t}e(x)$ for some real frequency $\omega$ (where $\mathrm{i}$ is the imaginary unit).
\item[(M3)] \emph{Independent events.} An event $\tilde x$ from a different drive or condition is drawn independently of $x$.
\end{itemize}
Both (M1) and (M2) hold whenever the flow is weakly mixing relative to a weakly mixing drive, because the total flow is then weakly mixing~\cite{furstenberg1977ergodic} and admits only constant eigenfunctions.

Explicitly, the top two asymptotic levels in \cref{supp:eq-levels} are given by
\begin{equation}
\begin{aligned}
  \bar\rho_{qc}&=\frac{\rho_{qc}}{\lVert\hat g_q\rVert\,\lVert\hat g_c\rVert}, & \rho_{qc}&=\langle\hat g_q,\hat g_c\rangle,\\
  \bar\rho^{u}_{qc}&=\frac{\rho^{u}_{qc}}{\lVert\hat g_q\rVert\,\lVert\hat g_c\rVert}, & \rho^{u}_{qc}&=\int_{\mathcal U}\mathbb{E}[\hat g_q\mid u]\cdot\mathbb{E}[\hat g_c\mid u]\,d\gamma .
\end{aligned}
  \label{eq:rhos}
\end{equation}
The difference between these two levels defines the \emph{event gap}, which equals the normalized conditional covariance given the drive:
\begin{equation}
  \bar\rho^{e}_{qc}=\bar\rho_{qc}-\bar\rho^{u}_{qc}=\frac{\int \operatorname{Cov}\big(\hat g_q,\hat g_c\mid u\big)\,d\gamma}{\lVert\hat g_q\rVert\,\lVert\hat g_c\rVert} ,
  \label{supp:eq-gap}
\end{equation}
\looseness=-1 where $\operatorname{Cov}(f,h\mid u)=\mathbb{E}[f\cdot h\mid u]-\mathbb{E}[f\mid u]\cdot\mathbb{E}[h\mid u]$.
We denote its unnormalized counterpart by $\rho^{e}_{qc}=\rho_{qc}-\rho^{u}_{qc}$.
Physically, the event gap quantifies cross-modal alignment at the event layer. It is positive when both modalities capture the same fine-grained contact fluctuations, and vanishes when either modality lacks an event layer (such as a camera recording only the trajectory of a replayed stroke).

\begin{proof}
The three limits correspond to ergodic averages under three distinct joint laws for $(x,\tilde x)$.
We first establish two preliminary properties of the flow, evaluate the unnormalized limit for each law in turn, and finally normalize.
Throughout, let $a_m(x)=\lVert\hat g_m(x)\rVert$ denote the pointwise Euclidean norm of the dynamics layer, so that $\int a_m^2\,d\mu=\lVert\hat g_m\rVert^2$.

\emph{Step 0 (preliminary flow properties).}
First, (M1) implies that $S_t$ is ergodic with respect to $\mu$.
Any $S_t$-invariant function $f$ lifts to the map $(x,x')\mapsto f(x)$, which is invariant under $S\times S$ and therefore constant $\Lambda$-almost everywhere. Because the first marginal of $\Lambda$ is $\mu$, $f$ must be constant $\mu$-almost everywhere.
Second, $\Lambda$ is invariant under $S\times S$.
Combining $\pi(S_tx)=R_t\pi(x)$ with the invariance of $\mu$ yields $\mathbb{E}[f\circ S_t\mid u]=\mathbb{E}[f\mid R_tu]$, which gives
\begin{equation}
\begin{aligned}
  \int f(S_tx)\,h(S_tx')\,d\Lambda
  &=\int_{\mathcal U}\mathbb{E}[f\mid R_tu]\,\mathbb{E}[h\mid R_tu]\,d\gamma(u)\\
  &=\int_{\mathcal U}\mathbb{E}[f\mid u]\,\mathbb{E}[h\mid u]\,d\gamma(u)\\
  &=\int f(x)\,h(x')\,d\Lambda .
\end{aligned}
\end{equation}
Here, the outer equalities follow from \cref{eq:relprod}, and the middle equality holds because $\gamma$ is invariant under $R_t$.

\emph{Step 1 (the same event).}
Applying the Cauchy--Schwarz inequality in $\mathbb{R}^k$ and then in $L^2(\mu)$ shows that the integrand is integrable:
\begin{equation}
  \int\lvert\hat g_q\cdot\hat g_c\rvert\,d\mu\le\int a_q\,a_c\,d\mu\le\lVert\hat g_q\rVert\,\lVert\hat g_c\rVert<\infty .
\end{equation}
Since $S_t$ is ergodic, Birkhoff's ergodic theorem ensures that for $\mu$-almost every $x$,
\begin{equation}
  \frac1T\int_0^T\hat g_q(S_tx)\cdot\hat g_c(S_tx)\,dt\ \longrightarrow\ \int\hat g_q\cdot\hat g_c\,d\mu=\rho_{qc} .
  \label{eq:lim-same}
\end{equation}

\emph{Step 2 (a twin event under the same drive).}
Here $(x,x')$ follows $\Lambda$, and we examine $\Xi(x,x')=\hat g_q(x)\cdot\hat g_c(x')$.
This function is $\Lambda$-integrable because
\begin{equation}
\begin{aligned}
  \int\lvert\Xi\rvert\,d\Lambda
  &\le\int a_q(x)\,a_c(x')\,d\Lambda\\
  &=\int_{\mathcal U}\mathbb{E}[a_q\mid u]\,\mathbb{E}[a_c\mid u]\,d\gamma\\
  &\le\Big(\int_{\mathcal U}\mathbb{E}[a_q\mid u]^2\,d\gamma\Big)^{1/2}\Big(\int_{\mathcal U}\mathbb{E}[a_c\mid u]^2\,d\gamma\Big)^{1/2}\\
  &\le\Big(\int_{\mathcal U}\mathbb{E}[a_q^2\mid u]\,d\gamma\Big)^{1/2}\Big(\int_{\mathcal U}\mathbb{E}[a_c^2\mid u]\,d\gamma\Big)^{1/2}\\
  &=\lVert\hat g_q\rVert\,\lVert\hat g_c\rVert .
\end{aligned}
\end{equation}
These inequalities follow sequentially from Cauchy--Schwarz in $\mathbb{R}^k$, the definition of $\Lambda$, Cauchy--Schwarz in $L^2(\gamma)$, the conditional Jensen inequality, and the law of total expectation.
Since $\Lambda$ is $S\times S$-invariant by Step~0 and ergodic under $S\times S$ by (M1), Birkhoff's theorem yields, for $\Lambda$-almost every pair,
\begin{equation}
  \frac1T\int_0^T\hat g_q(S_tx)\cdot\hat g_c(S_tx')\,dt\ \longrightarrow\ \int\Xi\,d\Lambda .
\end{equation}
Expanding the dot product across the $k$ coordinates (indexed by superscript $(i)$) and applying \cref{eq:relprod} term by term evaluates this limit:
\begin{equation}
\begin{aligned}
  \int\Xi\,d\Lambda
  &=\sum_{i=1}^k\int\hat g_q^{(i)}(x)\,\hat g_c^{(i)}(x')\,d\Lambda\\
  &=\sum_{i=1}^k\int_{\mathcal U}\mathbb{E}\big[\hat g_q^{(i)}\mid u\big]\,\mathbb{E}\big[\hat g_c^{(i)}\mid u\big]\,d\gamma\\
  &=\int_{\mathcal U}\mathbb{E}[\hat g_q\mid u]\cdot\mathbb{E}[\hat g_c\mid u]\,d\gamma=\rho^{u}_{qc} .
\end{aligned}
  \label{eq:lim-twin}
\end{equation}

\emph{Step 3 (a different drive or condition).}
By (M3), two events $(x,\tilde x)$ from the same condition under different drives follow the product law $\mu\times\mu$, which is invariant under $S\times S$.
Moreover, $\Xi$ is square-integrable under $\mu\times\mu$:
\begin{equation}
  \int\Xi^2\,d(\mu\times\mu)\le\int a_q^2\,d\mu\int a_c^2\,d\mu=\lVert\hat g_q\rVert^2\,\lVert\hat g_c\rVert^2 .
\end{equation}
By Birkhoff's theorem, the time average of $\Xi$ converges almost surely to $\bar\Xi$, the orthogonal projection of $\Xi$ onto the $S\times S$-invariant subspace of $L^2(\mu\times\mu)$.
This invariant subspace is spanned in the $L^2$ closure by products $\overline{e(x)}\,e'(\tilde x)$ of flow eigenfunctions sharing the same frequency~\cite{walters1982introduction,glasner2003joinings}, where the overbar denotes complex conjugation.
By (M2), every coordinate of $\hat g_q$ is orthogonal to $e$, making $\Xi$ orthogonal to every such product:
\begin{equation}
\begin{aligned}
  &\int\!\!\int\hat g_q(x)\cdot\hat g_c(\tilde x)\,e(x)\,\overline{e'(\tilde x)}\,d\mu(x)\,d\mu(\tilde x)\\
  &\quad=\sum_{i=1}^k\Big(\int\hat g_q^{(i)}\,e\,d\mu\Big)\Big(\int\hat g_c^{(i)}\,\overline{e'}\,d\mu\Big)=0 .
\end{aligned}
\end{equation}
Consequently, $\bar\Xi=0$, and almost surely
\begin{equation}
  \frac1T\int_0^T\hat g_q(S_tx)\cdot\hat g_c(S_t\tilde x)\,dt\ \longrightarrow\ 0 .
  \label{eq:lim-other}
\end{equation}
\looseness=-1 For two distinct conditions $\theta\neq\theta'$, the pair follows $\mu_\theta\times\mu_{\theta'}$, and the identical argument holds with $e$ and $e'$ chosen as eigenfunctions of the query and candidate flows, respectively.

\emph{Step 4 (normalization).}
Applying Step~1 to the self-pairs $(q,q)$ and $(c,c)$ shows that, for $\mu$-almost every state,
\begin{equation}
\begin{aligned}
  \frac1T\int_0^T\lVert\hat g_q(S_tx)\rVert^2dt&\ \longrightarrow\ \lVert\hat g_q\rVert^2,\\
  \frac1T\int_0^T\lVert\hat g_c(S_t\tilde x)\rVert^2dt&\ \longrightarrow\ \lVert\hat g_c\rVert^2 .
\end{aligned}
\end{equation}
In all three regimes, the candidate $\tilde x$ has marginal distribution $\mu$ (or $\mu_{\theta'}$ under a different condition), so both denominator limits hold almost surely.
Dividing \cref{eq:lim-same,eq:lim-twin,eq:lim-other} by the product of their square roots yields \cref{supp:eq-levels}:
\begin{equation}
\begin{aligned}
  \kappa(x,x)&\ \longrightarrow\ \frac{\rho_{qc}}{\lVert\hat g_q\rVert\,\lVert\hat g_c\rVert}=\bar\rho_{qc},\\
  \kappa(x,x')&\ \longrightarrow\ \frac{\rho^{u}_{qc}}{\lVert\hat g_q\rVert\,\lVert\hat g_c\rVert}=\bar\rho^{u}_{qc},\\
  \kappa(x,\tilde x)&\ \longrightarrow\ 0 .
\end{aligned}
\end{equation}

\emph{Step 5 (the event gap).}
Conditioning on the drive $u$ inside the integral defining $\rho_{qc}$ and applying the definition of conditional covariance gives
\begin{equation}
\begin{aligned}
  \rho_{qc}&=\int\hat g_q\cdot\hat g_c\,d\mu=\int_{\mathcal U}\mathbb{E}[\hat g_q\cdot\hat g_c\mid u]\,d\gamma\\
  &=\int_{\mathcal U}\operatorname{Cov}(\hat g_q,\hat g_c\mid u)\,d\gamma+\int_{\mathcal U}\mathbb{E}[\hat g_q\mid u]\cdot\mathbb{E}[\hat g_c\mid u]\,d\gamma\\
  &=\int_{\mathcal U}\operatorname{Cov}(\hat g_q,\hat g_c\mid u)\,d\gamma+\rho^{u}_{qc} .
\end{aligned}
\end{equation}
Subtracting $\rho^{u}_{qc}$ and dividing by $\lVert\hat g_q\rVert\,\lVert\hat g_c\rVert$ establishes \cref{supp:eq-gap}.
Finally, suppose $\hat g_q$ has no event layer, meaning it depends solely on $u$.
Then $\mathbb{E}[\hat g_q\mid u]=\hat g_q$ and $\mathbb{E}[\hat g_q\cdot\hat g_c\mid u]=\hat g_q\cdot\mathbb{E}[\hat g_c\mid u]$, so
\begin{equation}
  \operatorname{Cov}(\hat g_q,\hat g_c\mid u)=\hat g_q\cdot\mathbb{E}[\hat g_c\mid u]-\hat g_q\cdot\mathbb{E}[\hat g_c\mid u]=0 ,
\end{equation}
and the event gap vanishes. By symmetry, it also vanishes when the candidate modality lacks an event layer.
\end{proof}

\noindent \Cref{supp:thm-levels} compares recordings at identical timestamps. The following corollary formalizes the same-instants remark from the main paper, proving that zero offset achieves the widest event gap among all scores permitted to shift time.
For a time offset $s\in\mathbb{R}$, let
\begin{equation}
  \kappa_s(x,\tilde x)=\frac{\int_0^T \hat g_q(S_t x)\cdot\hat g_c(S_{t+s} \tilde x)\,dt}
  {\big(\int_0^T\lVert\hat g_q(S_t x)\rVert^2dt\big)^{1/2}\big(\int_0^T\lVert\hat g_c(S_{t+s} \tilde x)\rVert^2dt\big)^{1/2}}
  \label{supp:eq-kappa-s}
\end{equation}
\looseness=-1 denote the consistency between the query and the candidate shifted by $s$, so that $\kappa_0=\kappa$. Positive offsets read the candidate later in time, while negative offsets read it earlier.

\begin{corollary}[The same instants]
\label{supp:cor-instants}\leavevmode\ignorespaces
Let $\mathcal{S}$ be a finite set of time offsets containing zero, and let the assumptions of \cref{supp:thm-levels} hold.
Define the offset-dependent levels
\begin{equation}
\begin{aligned}
  \bar\rho_{qc}(s)&=\frac{\langle\hat g_q,\,U_s\hat g_c\rangle}{\lVert\hat g_q\rVert\,\lVert\hat g_c\rVert},\\
  \bar\rho^{u}_{qc}(s)&=\frac{\int_{\mathcal U}\mathbb{E}[\hat g_q\mid u]\cdot\mathbb{E}[U_s\hat g_c\mid u]\,d\gamma}{\lVert\hat g_q\rVert\,\lVert\hat g_c\rVert},
\end{aligned}
  \label{supp:eq-rho-s}
\end{equation}
so that $\bar\rho_{qc}(0)=\bar\rho_{qc}$ and $\bar\rho^{u}_{qc}(0)=\bar\rho^{u}_{qc}$.
Assume that two modalities from the same event align best at zero offset, meaning $\bar\rho_{qc}(s)\le\bar\rho_{qc}$ for every $s\in\mathcal{S}$.
Then, almost surely,
\begin{equation}
\begin{aligned}
  \max_{s\in\mathcal{S}}\kappa_s(x,x)&\ \longrightarrow\ \bar\rho_{qc},\\
  \max_{s\in\mathcal{S}}\kappa_s(x,x')&\ \longrightarrow\ \max_{s\in\mathcal{S}}\bar\rho^{u}_{qc}(s)\ \ge\ \bar\rho^{u}_{qc},
\end{aligned}
  \label{supp:eq-max-levels}
\end{equation}
and the separation between these two limits is at most the event gap $\bar\rho^{e}_{qc}$.
\end{corollary}

\begin{proof}
\emph{Step 1 (time offset as a shifted observable).}
Fix $s\in\mathcal{S}$. By definition of the Koopman operator, $\hat g_c(S_{t+s}\tilde x)=(U_s\hat g_c)(S_t\tilde x)$, so $\kappa_s$ coincides with the consistency in \cref{supp:eq-kappa} for the shifted candidate observable $U_sg_c$.
Because $\mu$ is invariant, $U_sg_c\in L^2(\mu)$.
Moreover, $U_s$ preserves the constant $Pg_c$, so the dynamics layer of $U_sg_c$ is
\begin{equation}
  (I-P)\,U_sg_c=U_sg_c-Pg_c=U_s(g_c-Pg_c)=U_s\hat g_c ,
\end{equation}
and its norm equals $\lVert\hat g_c\rVert$ by invariance of $\mu$:
\begin{equation}
  \lVert U_s\hat g_c\rVert^2=\int\lVert\hat g_c(S_sx)\rVert^2\,d\mu=\int\lVert\hat g_c\rVert^2\,d\mu=\lVert\hat g_c\rVert^2 .
\end{equation}

\emph{Step 2 (convergence at a fixed offset).}
Because (M1) and (M2) depend only on the flow and the query observable, they remain unaffected by shifting the candidate.
Applying \cref{supp:thm-levels} to the pair $(g_q,U_sg_c)$ therefore yields, almost surely,
\begin{equation}
  \kappa_s(x,x)\ \longrightarrow\ \bar\rho_{qc}(s)\qquad\text{and}\qquad\kappa_s(x,x')\ \longrightarrow\ \bar\rho^{u}_{qc}(s) .
\end{equation}

\emph{Step 3 (maximizing over offsets).}
Because $\mathcal{S}$ is finite, the convergence in Step~2 holds simultaneously for all $s\in\mathcal{S}$ almost surely, and the maximum of finitely many convergent sequences converges to the maximum of their limits:
\begin{equation}
\begin{aligned}
  \max_{s\in\mathcal{S}}\kappa_s(x,x)&\ \longrightarrow\ \max_{s\in\mathcal{S}}\bar\rho_{qc}(s),\\
  \max_{s\in\mathcal{S}}\kappa_s(x,x')&\ \longrightarrow\ \max_{s\in\mathcal{S}}\bar\rho^{u}_{qc}(s) .
\end{aligned}
\end{equation}
Since $0\in\mathcal{S}$ and $\bar\rho_{qc}(s)\le\bar\rho_{qc}(0)=\bar\rho_{qc}$ for all $s\in\mathcal{S}$, the first limit equals $\bar\rho_{qc}$, while the second limit is bounded below by $\bar\rho^{u}_{qc}(0)=\bar\rho^{u}_{qc}$. This establishes \cref{supp:eq-max-levels}.

\emph{Step 4 (bound on the gap).}
Taking the difference between the two limits in \cref{supp:eq-max-levels} gives
\begin{equation}
  \bar\rho_{qc}-\max_{s\in\mathcal{S}}\bar\rho^{u}_{qc}(s)\ \le\ \bar\rho_{qc}-\bar\rho^{u}_{qc}=\bar\rho^{e}_{qc} . \qedhere
\end{equation}
\end{proof}

\noindent Searching over time offsets therefore preserves the same-event score while potentially inflating the same-drive twin score, confirming that temporal realignment cannot widen the event gap.

\subsection{Twin events identify the event gap}
\label{sec:proof-repeats}

Comparing an event with itself and with a same-drive twin cancels the drive and leaves the event-specific covariance.

\begin{proposition}[Twin events identify the event gap]
\label{supp:prop-twin}\leavevmode\ignorespaces
Under the assumptions of \cref{supp:thm-levels}, for $\Lambda$-almost every same-drive twin pair $(x,x')$,
\begin{equation}
  \frac1T\int_0^T\hat g_q(S_tx)\cdot\big(\hat g_c(S_tx)-\hat g_c(S_tx')\big)\,dt\ \longrightarrow\ \rho^{e}_{qc} .
  \label{eq:twin-estimate}
\end{equation}
\end{proposition}

\begin{proof}
Subtracting the twin-event limit in \cref{eq:lim-twin} from the same-event limit in \cref{eq:lim-same} yields
\begin{equation}
\begin{aligned}
  &\frac1T\int_0^T\hat g_q(S_tx)\cdot\hat g_c(S_tx)\,dt-\frac1T\int_0^T\hat g_q(S_tx)\cdot\hat g_c(S_tx')\,dt\\
  &\quad\longrightarrow\ \rho_{qc}-\rho^{u}_{qc}=\rho^{e}_{qc} ,
\end{aligned}
\end{equation}
which establishes \cref{eq:twin-estimate}.
\end{proof}

\subsection{The demeaned cosine}
\label{sec:proof-cosine}

We now connect the empirical score computed by CounterPoint to the three theoretical levels established above.
Operating at the shared clock instants $t=1,\dots,T$, CounterPoint applies a linear projection $W_m$ to each encoder token $h_m(t)$ to obtain $y_m(t)=W_mh_m(t)$, computes its temporal average
\begin{equation}
  \bar y_m=\frac1T\sum_{t=1}^Ty_m(t) ,
\end{equation}
\looseness=-1 and compares the demeaned trajectories $\hat y_m(t)=y_m(t)-\bar y_m$ via cosine similarity.
Thus, $\hat y_m(t)$ serves as the learned, empirically centered counterpart to the dynamics layer $\hat g_m$.

\begin{proposition}[Convergence of the demeaned cosine]
\label{supp:prop-cosine}\leavevmode\ignorespaces
Let $x_t$ denote the system state at clock instant $t$.
As in the setting of \cref{sec:setting}, where the per-instant features are observables of the state, suppose the encoder tokens are state observables $h_m(t)=\eta_m(x_t)$ with $\eta_m\in L^2(\mu)$, and that the discrete-time state sequence satisfies the assumptions of \cref{supp:thm-levels}.
Write $y_m=W_m\eta_m$ and $\hat y_m=(I-P)y_m$.
Here $\hat y_m$ without a time argument denotes the dynamics layer of $y_m$, while $\hat y_m(t)$ denotes its empirical, window-centered value.
Then the cosine similarity between the demeaned trajectories converges almost surely to the three levels in \cref{supp:eq-levels}, with $\hat y_m$ taking the place of $\hat g_m$.
\end{proposition}

\begin{proof}
\emph{Step 1 (algebraic identity for demeaning).}
Expanding the empirical inner product of the centered trajectories and substituting the definition of $\bar y_m$ yields
\begin{equation}
\begin{aligned}
  &\frac1T\sum_{t=1}^T\big(y_q(t)-\bar y_q\big)\cdot\big(y_c(t)-\bar y_c\big)\\
  &\quad=\frac1T\sum_{t=1}^Ty_q(t)\cdot y_c(t)-\bar y_q\cdot\frac1T\sum_{t=1}^Ty_c(t)\\
  &\qquad-\Big(\frac1T\sum_{t=1}^Ty_q(t)\Big)\cdot\bar y_c+\bar y_q\cdot\bar y_c\\
  &\quad=\frac1T\sum_{t=1}^Ty_q(t)\cdot y_c(t)-\bar y_q\cdot\bar y_c .
\end{aligned}
  \label{eq:uncentered}
\end{equation}

\emph{Step 2 (ergodic limits of the two terms).}
Because $y_m(t)=y_m(x_t)$ with $y_m\in L^2(\mu)$ and the state sequence is ergodic, Birkhoff's ergodic theorem for sequences implies that, almost surely,
\begin{equation}
\begin{aligned}
  \frac1T\sum_{t=1}^Ty_q(t)\cdot y_c(t)&\ \longrightarrow\ \langle y_q,y_c\rangle,\\
  \bar y_m&\ \longrightarrow\ \mathbb{E}[y_m]=Py_m .
\end{aligned}
\end{equation}
The right-hand side of \cref{eq:uncentered} therefore converges almost surely to $\langle y_q,y_c\rangle-Py_q\cdot Py_c$.

\emph{Step 3 (matching the dynamics-layer correlation).}
Since $Py_m$ equals the constant $\mathbb{E}[y_m]$, the cross inner products simplify to $\langle Py_q,y_c\rangle=Py_q\cdot\mathbb{E}[y_c]=Py_q\cdot Py_c$ and $\langle y_q,Py_c\rangle=Py_q\cdot Py_c$.
Expanding $\langle\hat y_q,\hat y_c\rangle$ gives
\begin{equation}
\begin{aligned}
  \langle\hat y_q,\hat y_c\rangle
  &=\langle y_q-Py_q,\ y_c-Py_c\rangle\\
  &=\langle y_q,y_c\rangle-Py_q\cdot Py_c-Py_q\cdot Py_c+Py_q\cdot Py_c\\
  &=\langle y_q,y_c\rangle-Py_q\cdot Py_c .
\end{aligned}
\end{equation}
\looseness=-1Combining this identity with Step~2 shows that the demeaned product of an event with itself converges to $\langle\hat y_q,\hat y_c\rangle$, which is the same-event level $\rho_{qc}$ for the observables $y_m$.

\emph{Step 4 (twin events and independent events).}
The algebraic identity in Step~1 holds for any candidate trajectory.
For a same-drive twin event, the pair $(x,x')$ follows $\Lambda$. Applying the ergodic theorem to the product system (as in Step~2 of \cref{supp:thm-levels}) gives $\frac1T\sum_{t=1}^T y_q(x_t)\cdot y_c(x'_t)\to\int y_q(x)\cdot y_c(x')\,d\Lambda$, while the individual means still converge to $Py_q$ and $Py_c$.
Decomposing $y_m=Py_m+\hat y_m$, applying \cref{eq:relprod}, and using the zero-mean property $\int\hat y_m\,d\mu=0$ yields
\begin{equation}
\begin{aligned}
  &\int y_q(x)\cdot y_c(x')\,d\Lambda\\
  =&\int_{\mathcal U}\big(Py_q+\mathbb{E}[\hat y_q\mid u]\big)
  \cdot\big(Py_c+\mathbb{E}[\hat y_c\mid u]\big)\,d\gamma\\
  =&Py_q\cdot Py_c +\int_{\mathcal U}\mathbb{E}[\hat y_q\mid u]\cdot\mathbb{E}[\hat y_c\mid u]\,d\gamma ,
\end{aligned}
\end{equation}
because the cross terms vanish as $Py_q\cdot\int\hat y_c\,d\mu=0$ and $Py_c\cdot\int\hat y_q\,d\mu=0$.
Subtracting $Py_q\cdot Py_c$ leaves $\rho^{u}_{qc}$, the drive-level correlation of $y_m$.

For an independent event $\tilde x$, expanding $y_m=Py_m+\hat y_m$ produces four time averages.
The constant term $Py_q\cdot Py_c$ cancels against the product of the empirical means, the two cross terms vanish almost surely by the ergodic theorem because $\hat y_q$ and $\hat y_c$ have zero mean, and the product of the dynamics layers converges to zero by Step~3 of \cref{supp:thm-levels}.

\emph{Step 5 (normalization).}
Setting $q=c$ in Step~3 shows that the empirical variances $\frac1T\sum_{t=1}^T\lVert\hat y_m(t)\rVert^2$ converge almost surely to $\lVert\hat y_m\rVert^2$ for the query, a twin candidate, and an independent candidate alike, since both marginals of $\Lambda$ equal $\mu$.
Dividing the asymptotic inner products by the product of the limiting norms recovers the three normalized levels in \cref{supp:eq-levels}, and subtracting the twin-event limit from the same-event limit yields $\bar\rho^{e}_{qc}$.
\end{proof}

\subsection{Theorem~\ref{supp:thm-stall}: points stall, trajectories converge}
\label{sec:proof-stall}

We define \emph{event accuracy} as the probability of ranking an event $x$ above a same-drive twin $x'$ over random draws of the pair $(x,x')$.
For trajectories, we define the score margin $\Delta_T=\kappa(x,x)-\kappa(x,x')$, which yields an event accuracy of $\Pr[\Delta_T>0]$.

\begin{theorem}[Points stall, trajectories converge]
\label{supp:thm-stall}\leavevmode\ignorespaces
Fix the encoders under standard moment and mixing assumptions~\cite{bradley2005mixing}.
(i)~Unless the cross-modal point fluctuations are perfectly coupled, point event accuracy saturates at a constant $p^\ast<1$.
(ii)~If the event gap is strictly positive, then for all $T$,
\begin{equation}
  \Pr\big[\Delta_T\le 0\big]\le K/T,
  \label{supp:eq-bound}
\end{equation}
with $K\propto(\bar\rho^{e}_{qc})^{-2}$.
\end{theorem}

\noindent The proof relies on the following technical conditions and yields an explicit expression for $K$.

\paragraph{Assumptions.}
\begin{itemize}
\item[(A0)] \emph{Smooth pooling.} Each point takes the Deep Sets form~\cite{zaheer2017deepsets} $z_m=\psi_m(\bar\phi_T)$, an instance of the representation $\Psi(\nu_T^x)$ of \cref{supp:thm-collapse}, where $\bar\phi_T=\frac1T\int_0^T\phi(S_tx)\,dt$ is the window average of a feature vector $\phi\in L^2(\mu,\mathbb{R}^p)$ of the state and $\psi_m$ is a twice continuously differentiable map into the unit sphere. Mean pooling is the case where $\phi$ stacks the observables and $\psi_m$ normalizes. The trajectory margin $\Delta_T$ has this smooth form by construction (\cref{eq:margin-five}).
\item[(A1)] \emph{Central limit theorem.} There exist limits $z_m^\ast$ and a centered Gaussian vector $(\zeta_q,\zeta_c,\zeta_{c'})$ such that $\sqrt T\,(z_m-z_m^\ast)\Rightarrow\zeta_m$ jointly for the query $q$, the candidate $c$, and its twin candidate $c'$.
\item[(A2)] \looseness=-1 \emph{Imperfect coupling.} The difference $\zeta_c-\zeta_q$ is not a deterministic linear function of $\zeta_q-\zeta_{c'}$. When the limits $z_q^\ast$ and $z_c^\ast$ differ, $z_q^\ast\cdot(\zeta_c-\zeta_{c'})$ is not almost surely zero.
\item[(A3)] \emph{Moments and memory.} The dynamics layers satisfy $\hat g_q,\hat g_c\in L^4(\mu)$, and the variance of each of the five normalized averages entering $\Delta_T$ is bounded by $\Gamma/T$ for a constant $\Gamma$ and all $T$. \Cref{supp:lem-var} provides a sufficient condition for this bound.
\end{itemize}
Part (i) holds under (A0) through (A2). When the limits align ($z_q^\ast=z_c^\ast$), the saturation accuracy is
\begin{equation}
  p^\ast=\Pr\big[\lVert\zeta_q-\zeta_{c'}\rVert^2>\lVert\zeta_q-\zeta_c\rVert^2\big] .
\end{equation}
Part (ii) holds under (A0) and (A3) with the constant
\begin{equation}
  K=\frac{125\,\Gamma}{(\bar\rho^{e}_{qc})^2} .
  \label{eq:K}
\end{equation}

\begin{lemma}[Integrable autocovariances yield the variance bound]
\label{supp:lem-var}\leavevmode\ignorespaces
Let $f_t$ be a stationary process with finite variance and autocovariance $c_f(\ell)$ at lag $\ell$.
If $c_f$ is absolutely integrable, then for every $T$,
\begin{equation}
  \operatorname{Var}\Big(\frac1T\int_0^Tf_t\,dt\Big)\le\frac1T\int_{-\infty}^{\infty}\lvert c_f(\ell)\rvert\,d\ell .
  \label{eq:varbound}
\end{equation}
Consequently, if $\hat g_q,\hat g_c\in L^4(\mu)$ and the autocovariances of the five normalized integrands of $\Delta_T$ are absolutely integrable, then (A3) holds with $\Gamma$ the largest of the five integrals.
\end{lemma}

\begin{proof}
The variance of a time average equals the double integral of the autocovariance, which depends on the two times solely through their lag $\ell=t-s$.
Integrating first over pairs $(s,t)\in[0,T]^2$ with fixed lag $\ell$, which span an interval of length $T-\lvert\ell\rvert$, gives
\begin{equation}
\begin{aligned}
  \operatorname{Var}\Big(\frac1T\int_0^Tf_t\,dt\Big)
  &=\frac{1}{T^2}\int_0^T\!\!\int_0^Tc_f(t-s)\,ds\,dt\\
  &=\frac{1}{T^2}\int_{-T}^{T}\big(T-\lvert\ell\rvert\big)\,c_f(\ell)\,d\ell\\
  &=\frac1T\int_{-T}^{T}\Big(1-\frac{\lvert\ell\rvert}{T}\Big)c_f(\ell)\,d\ell\\
  &\le\frac1T\int_{-\infty}^{\infty}\lvert c_f(\ell)\rvert\,d\ell ,
\end{aligned}
\end{equation}
where the inequality uses $0\le1-\lvert\ell\rvert/T\le1$ on $[-T,T]$.
For the second claim, the five integrands have finite variance whenever $\hat g_q,\hat g_c\in L^4(\mu)$ (as verified in Step~3 of Part~(ii) below) and are stationary because $\mu$ and $\Lambda$ are invariant. Applying \cref{eq:varbound} to each integrand yields the bound $\Gamma/T$ in (A3).
\end{proof}

\begin{proof}[Proof of \cref{supp:thm-stall}]
We establish the two parts separately.

\medskip\noindent\textbf{Part (i): points.}
By (A0), each point is $z_m=\psi_m(\bar\phi_T)$, where $\bar\phi_T$ is the window average of the feature vector $\phi$ and $\psi_m$ is a twice continuously differentiable map into the unit sphere.
The query $q$ and candidate $c$ are evaluated on event $x$, whereas the twin candidate $c'$ is evaluated on $x'$ using the same map $\psi_c$.
Let $\phi^\ast=\mathbb{E}[\phi]$ and $z_m^\ast=\psi_m(\phi^\ast)$.

\emph{Step 1 (Gaussian fluctuations).}
Assumption (A1) follows from a joint central limit theorem for the feature averages:
\begin{equation}
  \sqrt T\,\big(\bar\phi_T(x)-\phi^\ast,\ \bar\phi_T(x')-\phi^\ast\big)\ \Rightarrow\ (\chi,\chi') ,
\end{equation}
where $(\chi,\chi')$ is a centered Gaussian vector in which $\chi'$ shares the marginal law of $\chi$ because $x'$ has the same law as $x$.
A first-order Taylor expansion of $\psi_m$ at $\phi^\ast$ with Jacobian $D\psi_m$ gives
\begin{equation}
  \sqrt T\,(z_m-z_m^\ast)=D\psi_m(\phi^\ast)\,\sqrt T\,(\bar\phi_T-\phi^\ast)+o_P(1) ,
\end{equation}
where $o_P(1)$ converges to zero in probability. Consequently,
\begin{equation}
\begin{aligned}
  &\sqrt T\,\big(z_q-z_q^\ast,\ z_c-z_c^\ast,\ z_{c'}-z_c^\ast\big)\\
  &\quad\Rightarrow\ (\zeta_q,\zeta_c,\zeta_{c'})=\big(D\psi_q\,\chi,\ D\psi_c\,\chi,\ D\psi_c\,\chi'\big) .
\end{aligned}
  \label{eq:point-clt}
\end{equation}
This limit is a centered Gaussian vector $\mathcal N(0,\Sigma)$ in which $\zeta_{c'}$ has the same marginal distribution as $\zeta_c$.

\emph{Step 2 (fluctuations lie tangent to the sphere).}
Because $\lVert z_m\rVert=\lVert z_m^\ast\rVert=1$, expanding the squared norm yields
\begin{equation}
  1=\lVert z_m^\ast+(z_m-z_m^\ast)\rVert^2=1+2\,z_m^\ast\cdot(z_m-z_m^\ast)+\lVert z_m-z_m^\ast\rVert^2 .
\end{equation}
Multiplying through by $\sqrt T/2$ gives
\begin{equation}
  z_m^\ast\cdot\sqrt T\,(z_m-z_m^\ast)=-\frac{1}{2\sqrt T}\,\big\lVert\sqrt T\,(z_m-z_m^\ast)\big\rVert^2\ \longrightarrow\ 0
\end{equation}
in probability. Passing to the limit yields $z_m^\ast\cdot\zeta_m=0$, meaning every fluctuation lies in the tangent space of the sphere at $z_m^\ast$.

\emph{Step 3 (the point margin).}
A point score ranks event $x$ above its twin $x'$ when the margin
\begin{equation}
  \Delta^{\rm pt}_T=z_q\cdot z_c-z_q\cdot z_{c'}
\end{equation}
is positive, so point event accuracy equals $\Pr[\Delta^{\rm pt}_T>0]$.

\emph{Step 4 (aligned limits: second-order margin).}
A contrastive aligner pulls the asymptotic limits of the query and candidate together, so $z_q^\ast=z_c^\ast=z^\ast$.
Using the identity $\lVert z-z'\rVert^2=2-2\,z\cdot z'$ for unit vectors $z$ and $z'$, we rewrite the margin as
\begin{equation}
  \Delta^{\rm pt}_T=\tfrac12\big(\lVert z_q-z_{c'}\rVert^2-\lVert z_q-z_c\rVert^2\big) .
\end{equation}
Each squared distance scales as $1/T$. Specifically, because the three limits coincide,
\begin{equation}
  T\,\lVert z_q-z_c\rVert^2=\big\lVert\sqrt T\,(z_q-z^\ast)-\sqrt T\,(z_c-z^\ast)\big\rVert^2\ \Rightarrow\ \lVert\zeta_q-\zeta_c\rVert^2 ,
\end{equation}
and the analogous limit holds jointly for $c'$ by \cref{eq:point-clt} and the continuous mapping theorem. Thus,
\begin{equation}
  T\,\Delta^{\rm pt}_T\ \Rightarrow\ Z=\tfrac12\big(\lVert\zeta_q-\zeta_{c'}\rVert^2-\lVert\zeta_q-\zeta_c\rVert^2\big) .
  \label{eq:point-limit}
\end{equation}
Both the signal and the noise of the margin decay at rate $1/T$, leaving a limiting distribution $Z$ that is independent of $T$.

\emph{Step 5 (the limit is negative with positive probability).}
Set $D_1=\zeta_c-\zeta_q$ and $D_2=\zeta_q-\zeta_{c'}$, so that $Z=\tfrac12\big(\lVert D_2\rVert^2-\lVert D_1\rVert^2\big)$.
Since $(D_1,D_2)$ is jointly centered Gaussian, we can decompose
\begin{equation}
  D_1=M\,D_2+D_3
\end{equation}
for a deterministic matrix $M$ and a centered Gaussian vector $D_3$ independent of $D_2$.
By (A2), $D_3$ is not almost surely zero, so it has positive variance along at least one direction and its norm exceeds any threshold with positive probability. Conditioning on $D_2$ therefore gives
\begin{equation}
  \Pr\big[\lVert M\,D_2+D_3\rVert>\lVert D_2\rVert\ \big|\ D_2\big]>0
\end{equation}
\looseness=-1 for every realization of $D_2$. Taking expectations over $D_2$ yields $\Pr[\lVert D_1\rVert>\lVert D_2\rVert]>0$, which is equivalent to $\Pr[Z<0]>0$.

\emph{Step 6 (event accuracy converges to $p^\ast<1$).}
The limit $Z$ is a quadratic form of a Gaussian vector that does not vanish identically on the support of that vector by Step~5.
Because the zero set of a non-trivial polynomial has Lebesgue measure zero, $\Pr[Z=0]=0$.
The interval $(0,\infty)$ therefore has a $\Pr_Z$-null boundary, and the portmanteau theorem applied to \cref{eq:point-limit} gives
\begin{equation}
  \Pr\big[\Delta^{\rm pt}_T>0\big]=\Pr\big[T\,\Delta^{\rm pt}_T>0\big]\ \longrightarrow\ \Pr[Z>0]=p^\ast .
\end{equation}
By Step~5, $p^\ast=1-\Pr[Z<0]<1$, and the definition of $Z$ gives $p^\ast=\Pr\big[\lVert\zeta_q-\zeta_{c'}\rVert^2>\lVert\zeta_q-\zeta_c\rVert^2\big]$.

\emph{Step 7 (unaligned limits).}
If $z_q^\ast\neq z_c^\ast$, the margin is of first order. Writing $\Delta^{\rm pt}_T=z_q\cdot(z_c-z_{c'})$ and using $z_q\to z_q^\ast$ in probability gives
\begin{equation}
  \sqrt T\,\Delta^{\rm pt}_T=z_q\cdot\big(\sqrt T\,(z_c-z_c^\ast)-\sqrt T\,(z_{c'}-z_c^\ast)\big)\ \Rightarrow\ z_q^\ast\cdot(\zeta_c-\zeta_{c'}) .
\end{equation}
This limit is a centered Gaussian random variable that is not almost surely zero by (A2). It is therefore positive with probability $\tfrac12$, giving $p^\ast=\tfrac12<1$.

\medskip\noindent\textbf{Part (ii): trajectories.}
Let the event $x$ and its twin $x'$ follow the joint distribution $\Lambda$, and recall that $\lVert\hat g_m\rVert$ denotes the $L^2(\mu)$ norm of the dynamics layer.

\emph{Step 1 (the margin in terms of five averages).}
The trajectory margin depends on the window through five time averages, normalized by their asymptotic expectations:
\begin{equation}
\begin{aligned}
  n&=\frac{1}{\lVert\hat g_q\rVert\,\lVert\hat g_c\rVert\,T}\int_0^T\hat g_q(S_tx)\cdot\hat g_c(S_tx)\,dt,\\
  n'&=\frac{1}{\lVert\hat g_q\rVert\,\lVert\hat g_c\rVert\,T}\int_0^T\hat g_q(S_tx)\cdot\hat g_c(S_tx')\,dt,\\
  w_q&=\frac{1}{\lVert\hat g_q\rVert^2\,T}\int_0^T\lVert\hat g_q(S_tx)\rVert^2dt,\\
  w_c&=\frac{1}{\lVert\hat g_c\rVert^2\,T}\int_0^T\lVert\hat g_c(S_tx)\rVert^2dt,\\
  w_{c'}&=\frac{1}{\lVert\hat g_c\rVert^2\,T}\int_0^T\lVert\hat g_c(S_tx')\rVert^2dt .
\end{aligned}
  \label{eq:five}
\end{equation}
Dividing the numerator and denominator of \cref{supp:eq-kappa} by $\lVert\hat g_q\rVert\,\lVert\hat g_c\rVert\,T$ expresses the consistencies and the margin as
\begin{equation}
\begin{aligned}
  \kappa(x,x)&=\frac{n}{\sqrt{w_q\,w_c}},\qquad\kappa(x,x')=\frac{n'}{\sqrt{w_q\,w_{c'}}},\\
  \Delta_T&=\frac{1}{\sqrt{w_q}}\Big(\frac{n}{\sqrt{w_c}}-\frac{n'}{\sqrt{w_{c'}}}\Big) .
\end{aligned}
  \label{eq:margin-five}
\end{equation}

\emph{Step 2 (expectations).}
Because $\mu$ and $\Lambda$ are invariant, the expectation of each integrand is constant in $t$, so the expectation of each window average equals that of its integrand at $t=0$:
\begin{equation}
\begin{aligned}
  \mathbb{E}[n]&=\frac{1}{\lVert\hat g_q\rVert\,\lVert\hat g_c\rVert}\int\hat g_q\cdot\hat g_c\,d\mu=\bar\rho_{qc},\\
  \mathbb{E}[n']&=\frac{1}{\lVert\hat g_q\rVert\,\lVert\hat g_c\rVert}\int\hat g_q(x)\cdot\hat g_c(x')\,d\Lambda=\bar\rho^{u}_{qc},\\
  \mathbb{E}[w_q]&=\mathbb{E}[w_c]=\mathbb{E}[w_{c'}]=1 .
\end{aligned}
  \label{eq:five-means}
\end{equation}
Here, the second line follows from \cref{eq:lim-twin}, and the third uses the fact that both marginals of $\Lambda$ equal $\mu$.

\emph{Step 3 (variances).}
Each of the five integrands is square-integrable.
With $a_m(x)=\lVert\hat g_m(x)\rVert$, the Cauchy--Schwarz inequality bounds the second moment of the same-event product:
\begin{equation}
\begin{aligned}
  \int(\hat g_q\cdot\hat g_c)^2d\mu&\le\int a_q^2\,a_c^2\,d\mu\\
  &\le\Big(\int a_q^4\,d\mu\Big)^{1/2}\Big(\int a_c^4\,d\mu\Big)^{1/2}<\infty ,
\end{aligned}
\end{equation}
which is finite by (A3). For the twin product, the definition of $\Lambda$, Cauchy--Schwarz, and the conditional Jensen inequality yield the same bound:
\begin{equation}
\begin{aligned}
  \int a_q(x)^2\,a_c(x')^2\,d\Lambda&=\int_{\mathcal U}\mathbb{E}[a_q^2\mid u]\,\mathbb{E}[a_c^2\mid u]\,d\gamma\\
  &\le\Big(\int a_q^4\,d\mu\Big)^{1/2}\Big(\int a_c^4\,d\mu\Big)^{1/2} ,
\end{aligned}
\end{equation}
and the squared norms follow as the special case $q=c$.
All five averages therefore have finite variance, which (A3) bounds by $\Gamma/T$.

\emph{Step 4 (constructing a high-probability event).}
Set $\varepsilon=\bar\rho^{e}_{qc}/5$, and let $G$ denote the event on which every one of the five averages falls within $\varepsilon$ of its expectation in \cref{eq:five-means}.
By Cauchy--Schwarz (and conditional Jensen for the twin level), $\lvert\bar\rho_{qc}\rvert\le1$ and $\lvert\bar\rho^{u}_{qc}\rvert\le1$, so
\begin{equation}
  \bar\rho^{e}_{qc}=\bar\rho_{qc}-\bar\rho^{u}_{qc}\le2\qquad\text{and}\qquad\varepsilon\le\tfrac25 .
\end{equation}
On $G$, each denominator average $w$ satisfies $\lvert w-1\rvert<\varepsilon\le\tfrac25$, implying $w>\tfrac35$ and
\begin{equation}
\begin{aligned}
  \frac{1}{\sqrt w}&<\sqrt{\tfrac53}<1.3,\\
  \Big\lvert\frac{1}{\sqrt w}-1\Big\rvert&=\frac{\lvert1-w\rvert}{\sqrt w\,(1+\sqrt w)}<\varepsilon ,
\end{aligned}
\end{equation}
where the second inequality uses $\sqrt w\,(1+\sqrt w)>1$ for $w>\tfrac35$.
Consequently, for any level $r$ with $\lvert r\rvert\le1$ and any average $\tilde n$ with $\lvert\tilde n-r\rvert<\varepsilon$,
\begin{equation}
  \Big\lvert\frac{\tilde n}{\sqrt w}-r\Big\rvert\le\frac{\lvert\tilde n-r\rvert}{\sqrt w}+\lvert r\rvert\,\Big\lvert\frac{1}{\sqrt w}-1\Big\rvert<1.3\,\varepsilon+\varepsilon=2.3\,\varepsilon .
  \label{eq:ratio}
\end{equation}

\emph{Step 5 (positivity of the margin on $G$).}
Applying \cref{eq:ratio} to $(\tilde n,w,r)=(n,w_c,\bar\rho_{qc})$ and $(n',w_{c'},\bar\rho^{u}_{qc})$ shows that, on $G$,
\begin{equation}
  \frac{n}{\sqrt{w_c}}>\bar\rho_{qc}-2.3\,\varepsilon\qquad\text{and}\qquad\frac{n'}{\sqrt{w_{c'}}}<\bar\rho^{u}_{qc}+2.3\,\varepsilon .
\end{equation}
Subtracting these bounds and substituting into \cref{eq:margin-five} yields
\begin{equation}
  \sqrt{w_q}\,\Delta_T>\bar\rho^{e}_{qc}-4.6\,\varepsilon=5\,\varepsilon-4.6\,\varepsilon>0 .
\end{equation}
Since $w_q>0$ on $G$, it follows that $\Delta_T>0$ on $G$.

\emph{Step 6 (error probability bound).}
By Step~5, the failure event $\{\Delta_T\le0\}$ is contained in the complement $G^c$.
Denoting the five averages by $A_1,\dots,A_5$, $G^c$ is the union of the five events $\{\lvert A_i-\mathbb{E}[A_i]\rvert\ge\varepsilon\}$.
Applying the union bound, Chebyshev's inequality, and the variance bound from (A3) gives, for all $T$,
\begin{equation}
\begin{aligned}
  \Pr[\Delta_T\le0]\le\Pr[G^c]
  &\le\sum_{i=1}^5\Pr\big[\lvert A_i-\mathbb{E}[A_i]\rvert\ge\varepsilon\big]\\
  &\le\sum_{i=1}^5\frac{\operatorname{Var}(A_i)}{\varepsilon^2}
  \le\frac{5\,\Gamma}{\varepsilon^2\,T}=\frac{125\,\Gamma}{(\bar\rho^{e}_{qc})^2\,T} ,
\end{aligned}
\end{equation}
establishing \cref{supp:eq-bound} with the constant $K$ in \cref{eq:K}.
\end{proof}

\begin{proposition}[Asymptotic normality of the margin]
\label{supp:prop-fluct}\leavevmode\ignorespaces
Let $A_T=(n,n',w_q,w_c,w_{c'})$ be the vector of the five averages in \cref{eq:five}, and suppose it satisfies a central limit theorem,
\begin{equation}
  \sqrt T\,\big(A_T-\mathbb{E}[A_T]\big)\ \Rightarrow\ \mathcal N(0,\Sigma_A) ,
  \label{eq:clt-five}
\end{equation}
where $\Sigma_A$ is the long-run covariance matrix of the five integrands.
Then
\begin{equation}
  \sqrt T\,\big(\Delta_T-\bar\rho^{e}_{qc}\big)\ \Rightarrow\ \mathcal N(0,\Sigma_\Delta),\qquad
  \Sigma_\Delta=\nabla\Phi^\top\,\Sigma_A\,\nabla\Phi ,
\end{equation}
where $\Phi(v)=v_3^{-1/2}\big(v_1\,v_4^{-1/2}-v_2\,v_5^{-1/2}\big)$, $\nabla\Phi$ denotes its gradient evaluated at $\mathbb{E}[A_T]=(\bar\rho_{qc},\bar\rho^{u}_{qc},1,1,1)$, and $^\top$ denotes matrix transposition.
\end{proposition}

\begin{proof}
\emph{Step 1 (smoothness of the margin map).}
By \cref{eq:margin-five}, $\Delta_T=\Phi(A_T)$, and by \cref{eq:five-means},
\begin{equation}
  \Phi\big(\mathbb{E}[A_T]\big)=\bar\rho_{qc}-\bar\rho^{u}_{qc}=\bar\rho^{e}_{qc} .
\end{equation}
Because the last three coordinates of $\mathbb{E}[A_T]$ equal $1$ and are strictly positive, $\Phi$ is continuously differentiable in a neighborhood of $\mathbb{E}[A_T]$.

\emph{Step 2 (first-order Taylor expansion).}
Differentiability at $\mathbb{E}[A_T]$ gives
\begin{equation}
\begin{aligned}
  \Delta_T-\bar\rho^{e}_{qc}&=\nabla\Phi^\top\big(A_T-\mathbb{E}[A_T]\big)+R_T ,\\
  \lvert R_T\rvert&\le\omega\big(\lVert A_T-\mathbb{E}[A_T]\rVert\big)\,\lVert A_T-\mathbb{E}[A_T]\rVert ,
\end{aligned}
\end{equation}
for a modulus function $\omega$ satisfying $\omega(s)\to0$ as $s\to0$.

\emph{Step 3 (vanishing remainder).}
By \cref{eq:clt-five}, $\sqrt T\,\lVert A_T-\mathbb{E}[A_T]\rVert$ is stochastically bounded and $A_T-\mathbb{E}[A_T]\to0$ in probability. Hence
\begin{equation}
  \sqrt T\,\lvert R_T\rvert\le\omega\big(\lVert A_T-\mathbb{E}[A_T]\rVert\big)\,\sqrt T\,\lVert A_T-\mathbb{E}[A_T]\rVert\ \longrightarrow\ 0
\end{equation}
in probability.

\emph{Step 4 (limiting distribution).}
Multiplying the expansion in Step~2 by $\sqrt T$ gives
\begin{equation}
  \sqrt T\,\big(\Delta_T-\bar\rho^{e}_{qc}\big)=\nabla\Phi^\top\sqrt T\,\big(A_T-\mathbb{E}[A_T]\big)+\sqrt T\,R_T .
\end{equation}
The first term converges in distribution to $\nabla\Phi^\top\mathcal N(0,\Sigma_A)=\mathcal N(0,\nabla\Phi^\top\Sigma_A\nabla\Phi)$ by \cref{eq:clt-five}, while the remainder $\sqrt T\,R_T$ converges to zero in probability by Step~3. By Slutsky's theorem, the sum has the same Gaussian limit.
\end{proof}

\noindent Thus, the trajectory margin concentrates around the event gap $\bar\rho^{e}_{qc}$ with variance decaying as $1/T$. \Cref{sec:scaling} verifies both behaviors empirically on real recordings.

\section{Trajectory bound across windows and speeds}
\label{sec:scaling}

This section first examines two properties of the recorded signals on which the assumptions of the theory rest, then details the test of the bound of \cref{supp:eq-bound} reported in \cref{sec:exp_levels}, and examines how the event accuracy of a trajectory depends on the sliding speed.

\paragraph{The signals have short memory and agree at the same instants.}
We measure two properties of the signals before any model, on the 3,200 recordings of the Cluster development split (\cref{sec:data}), using the audio and vibration envelopes at 10\,ms steps with the mean envelope of each motion condition subtracted, the signals of the raw coupling in \cref{sec:more-rq1}.
First, the memory of each signal is short. The autocorrelation of the envelope falls below $1/e$ after 10.3\,ms for audio and 9.8\,ms for vibration, and the correlation between the averages of neighboring 40\,ms tokens is 0.40 and 0.17. The autocovariances thus decay well within the 1\,s window, the short memory on which the variance bound of (A3) and \cref{supp:lem-var} rests.
Second, the two modalities agree most at the same instants. The audio--vibration cross-correlation peaks at zero lag for the event (0.149) and for its twin event (0.071), and stays near zero at every lag for another texture. This is the property that \cref{supp:cor-instants} assumes, $\bar\rho_{qc}(s)\le\bar\rho_{qc}$, and the higher peak of the event shows an event gap already in the raw signals.

\begin{figure}[tbp]
\centering
\includegraphics[width=\linewidth]{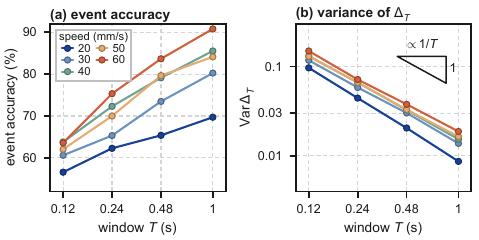}
\caption{\looseness=-1 \textbf{Longer windows raise accuracy and reduce margin variance} (\cref{fig:law} of the main paper, reproduced). The fixed 1\,s Cluster encoder evaluated on the central $T$ instants of its window at each of the five speeds. (a)~Event accuracy against the window. (b)~Variance of the margin $\Delta_T$ between the event and its twin event, where the triangle marks the $1/T$ reference slope.}
\label{supp:fig-law}
\end{figure}

\paragraph{A fixed encoder follows the bound.}
\Cref{supp:thm-stall} fixes the encoder, so we hold the 1\,s encoder fixed (3 seeds) on the Cluster development split and evaluate it at every window, by feeding it only the central $T$ instants with their positions (\cref{supp:fig-law}).
At every speed, the mean of $\Delta_T$ shifts by at most 0.013 from 0.12 to 1\,s, and its variance decays as $1/T$ (\cref{supp:fig-law}b).
These are the two ingredients of the proof of \cref{supp:thm-stall}, a margin centered near the event gap and a variance that decays as $1/T$.
Together they make the event accuracy rise with the window (\cref{supp:fig-law}a).

\paragraph{Faster strokes have a larger event gap.}
The mean of the margin grows with speed, from 0.05 at 20\,mm/s to 0.15 at 60\,mm/s.
Faster strokes are therefore more accurate at the same window (\cref{supp:fig-law}a), consistent with part (ii) of \cref{supp:thm-stall}, whose constant $K\propto(\bar\rho^{e}_{qc})^{-2}$ shrinks as the event gap grows.

%% =====================================================================

\section{Datasets}
\label{sec:data}

\begin{figure*}[tp]
\centering
\includegraphics[width=\linewidth]{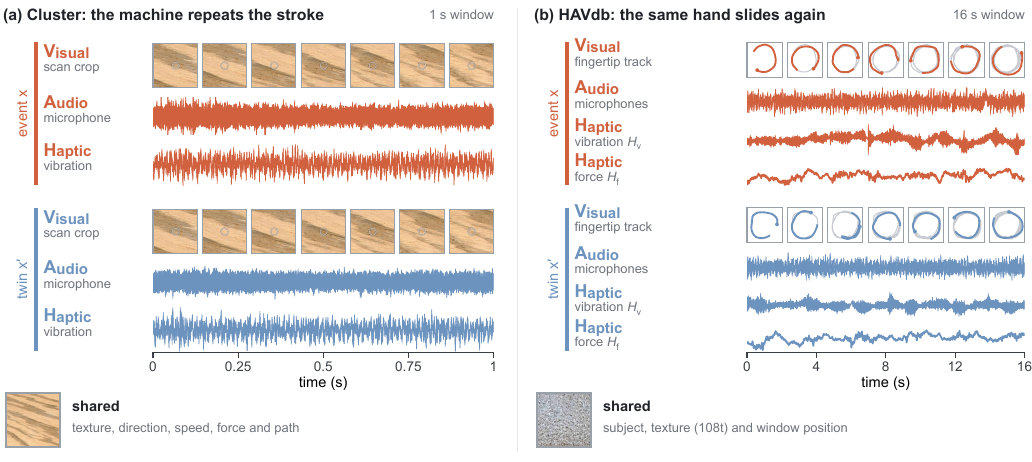}
\caption{\textbf{Every test event $x$ has a twin event $x'$ under identical conditions.} The recorded signals of an event $x$ (red) and of its twin event $x'$ (blue) on the shared clock, each modality drawn on one common scale for $x$ and $x'$. (a)~Cluster, one 1\,s window: the machine runs the same texture, direction, speed, force and path again. The visual stream is the crop of the texture scan that follows the probe, shown as seven of its frames. (b)~HAVdb, one 16\,s window at the same position in two standard slides of one subject on texture 108t. The visual stream is the fingertip tracked by the cameras as it slides over the texture, in image coordinates, shown in seven consecutive slices with the earlier track in grey. The haptic modality combines vibration \Hv{} and force \Hf. The swatch under each panel shows the shared texture, the whole Cluster scan in (a) and a central crop of the photograph of texture 108t taken by the tracking camera in (b). The two events agree in their condition layer and differ in their fine timing, which an aligner must resolve.}
\label{fig:data}
\end{figure*}

The two evaluation datasets instantiate the two theoretical regimes of \cref{supp:thm-levels} (\cref{tab:datasets}).
In Cluster, a motorized stage replays each stroke along a fixed path, placing twin events in the shared-drive regime.
In HAVdb, human subjects slide their fingertips freely, placing twin events in the shared-condition regime.
Both benchmarks include repeated trials under matching conditions (\cref{fig:data}), enabling direct measurement of the event gap.

\begin{table*}[tp]
\centering
\caption{\textbf{Overview of evaluation datasets.} Both benchmarks pair every event with twin recordings under matching conditions. Cluster enforces machine-controlled trajectories, whereas HAVdb captures unconstrained human hand motion.}
\label{tab:datasets}
\footnotesize
\setlength{\tabcolsep}{4pt}
\begin{adjustbox}{max width=\linewidth}
\begin{tabular}{lll}
\toprule
 & Cluster Haptic Texture~\cite{eguchi2024cluster} & HAVdb~\cite{devillard2025havdb} \\
\midrule
Actuator & three-axis machine with load cell & human hand, 10 subjects \\
Modalities & haptic vibration (H), audio (A), visual (V, scan crop) & haptic vibration (\Hv), force (\Hf), audio (A), visual (V, tracked fingertip) \\
Rates & A 44.1\,kHz, H $\approx$8.6\,kHz, V 30\,fps & A and \Hv{} 12.5\,kHz, \Hf{} 12.2\,kHz, V $\approx$30--60\,Hz \\
Surfaces & 118 textures, 9 categories & 10 textures \\
Conditions & 8 directions $\times$ 5 speeds $\times$ 2 forces & free sliding (standard slides) \\
Recordings & 18,880 & 500 slides, 300 of them standard \\
Twin event of an event & other recording, same texture and condition & same window, another standard slide, same subject and texture \\
Length & 1.2--4.5\,s plateau, 1\,s windows & 14--60\,s, windows of 1, 4, 16\,s \\
Clock & 40\,ms tokens & 40\,ms tokens \\
Split & textures (5 folds), categories, speeds, directions & subjects (5 folds) \\
Candidates & 3,040 or 3,200 per query & 243--267 per fold \\
\bottomrule
\end{tabular}
\end{adjustbox}
\end{table*}

\begin{figure*}[tp]
\centering
\includegraphics[width=\linewidth]{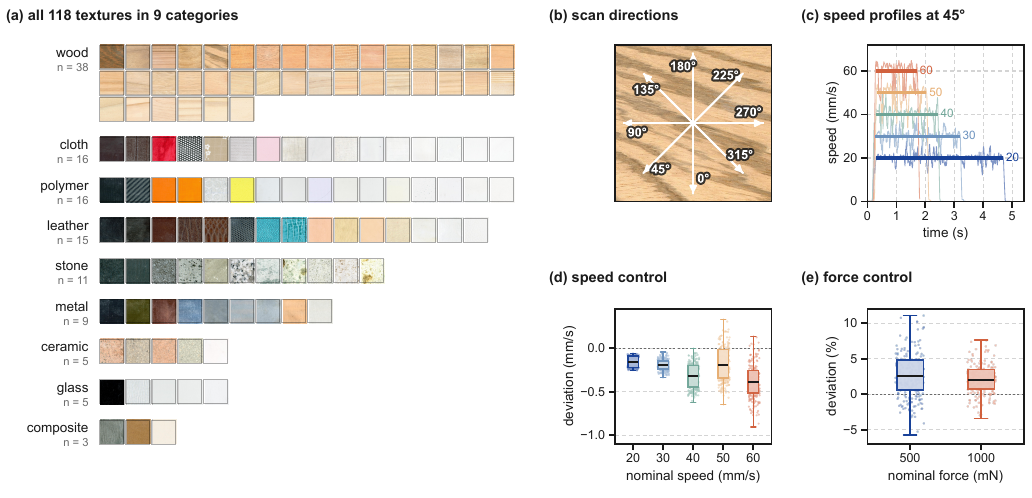}
\caption{\textbf{Cluster Haptic Texture, controlled motion.} (a)~All 118 textures in their recorded colors, the central part of each scan, grouped by category with the number of textures in each and ordered by brightness within a category. (b)~The 8 scan paths on one texture from the position log, each labeled at its end with its direction in the dataset. (c)~The speed profiles of the 5 speeds along 45\textdegree, with the constant-speed plateau drawn thick at its nominal value. (d,\,e)~The deviation of the measured speed and of the measured normal force from their nominal values over all 18,880 recordings, each box over a random sample of the recordings.}
\label{fig:datacluster}
\end{figure*}

\paragraph{Cluster Haptic Texture.}
A three-axis motorized stage presses a finger probe onto a $100\times100$\,mm texture sample and translates it along a 90\,mm straight path through the center (\cref{fig:datacluster}b).
Each of the 118 textures (\cref{fig:datacluster}a) is recorded twice across 8 directions, 5 speeds, and 2 normal forces, yielding $118\times8\times5\times2\times2=18{,}880$ recordings.
Mechanical repeatability is tight (\cref{fig:datacluster}d,\,e), with measured speed deviating from nominal by 0.26\,mm/s on average (1.0\,mm/s maximum) and measured force exceeding nominal by 2.6\%.
Following initial acceleration, the probe maintains a constant-speed plateau lasting 1.2 to 4.5\,s (\cref{fig:datacluster}c), from which we extract 1\,s windows.
We call each combination of direction, speed, and force a \emph{motion condition} ($8\times5\times2=80$ per texture), so the condition $\theta$ of a window specifies its texture and motion condition.

During each stroke, a microphone captures denoised audio at 44.1\,kHz, and a three-axis accelerometer mounted near the probe records vibration at roughly 8.6\,kHz on the same clock as the load cell.
Although the dataset lacks a camera stream, it provides a registered high-resolution scan of every surface. We synthesize the visual modality V at 30\,fps as a $30\times30$\,mm crop of this scan centered on the logged probe position, with the finger rendered as a ring.
Because this stream encodes the surface appearance and nominal trajectory without capturing event-specific contact fluctuations, \cref{supp:thm-levels} predicts a zero event gap. Empirically, its event gap with audio and with haptics is 0.000 (\cref{tab:regimes}), and CounterPoint changes HA$\to$V R@1 by at most 0.3 percentage points (\cref{tab:main}).

\paragraph{Cluster splits.}
\looseness=-1 We define four evaluation splits on Cluster, denoted S1 through S4, all sharing a fixed 20-texture \emph{development split} used for model selection and score-weight fitting:
\begin{itemize}
\item \textbf{S1 (held-out textures):} The remaining 98 textures are partitioned into 5 folds of 19 or 20 test textures across all speeds and directions. Unless noted otherwise, all Cluster experiments use S1.
\item \textbf{S2 (held-out categories):} Whole material categories are held out in a 9-fold cross-validation scheme (one fold per category).
\item \textbf{S3 (held-out speeds):} Using the 98 textures, models are evaluated on a single unseen speed, either 40\,mm/s (speed interpolation) or 60\,mm/s (speed extrapolation).
\item \textbf{S4 (held-out directions):} Using the 98 textures, models are trained on the four axial directions and tested on the four diagonal directions (1 fold).
\end{itemize}
Every Cluster metric labeled \emph{development} is computed on the 20-texture development split.

\begin{figure*}[tp]
\centering
\includegraphics[width=\linewidth]{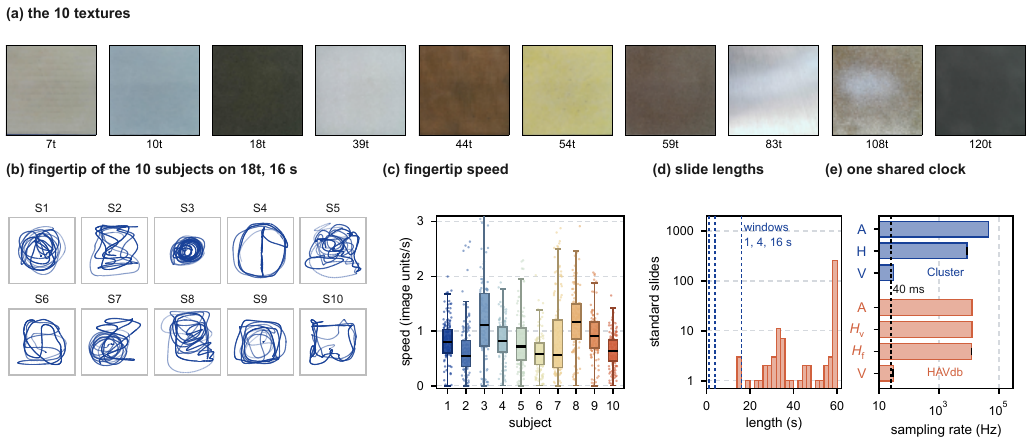}
\caption{\textbf{HAVdb, free motion.} (a)~The 10 textures. (b)~The tracked fingertip of all 10 subjects on texture 18t over the first 16\,s of one standard slide each, in image coordinates at one common scale, with the opacity of the path growing with time. (c)~The fingertip speed of each subject over all of its standard slides, in image units per second. (d)~The lengths of the standard slides on the shared clock and our three windows. (e)~The sampling rates of the sensors of both datasets, as the median over recordings with the 5th to 95th percentiles where they vary, all resampled to one 40\,ms clock.}
\label{fig:datahavdb}
\end{figure*}
\paragraph{HAVdb.}
Ten human subjects slide a fingertip across 10 textures (\cref{fig:datahavdb}a) for five trials per texture, consisting of three standard free-sliding trials, one constant-force trial, and one constant-speed trial.
Audio and fingertip vibrations (at the nail and phalanx) are recorded at 12.5\,kHz via two microphones and accelerometers, while a six-axis force-torque sensor beneath the surface samples at 12.2\,kHz.
Simultaneously, two overhead 4K cameras track two optical markers on the fingertip, providing the visual stream V.
Models train on all trials from the training subjects, and twin events are paired across the three standard trials, most of which span the full 60\,s duration (\cref{fig:datahavdb}d).
Because subjects draw freehand loops (\cref{fig:datahavdb}b) at speeds that vary both within and across trials (\cref{fig:datahavdb}c), twin events share the subject, texture, and window index without sharing an identical trajectory.
Unlike the path-only visual stream in Cluster, the tracked fingertip in HAVdb follows actual hand motion and therefore carries genuine event-level dynamics.

\paragraph{HAVdb evaluation windows.}
We group the 10 subjects into five pairs for cross-validation. Fold $k$ evaluates on one subject pair, uses the preceding pair as its development split, and trains on the remaining six subjects.
While training windows are sampled at random offsets, every development or test slide lasting at least 50\,s yields three evenly spaced evaluation windows from start to finish at each tested window length.
Consequently, the candidate pool size remains constant across window lengths (roughly 260 candidates per fold at 1, 4, and 16\,s), matching the fixed-candidate protocol of \cref{fig:scaling} so that longer windows enrich temporal context without altering retrieval difficulty.
For evaluation, the twin of a window is the window at the same temporal index within another standard trial of the same subject and texture.

\paragraph{Resampling to the shared clock.}
Because sensor sampling rates span three orders of magnitude (\cref{fig:datahavdb}e), we first extract frame-level features on each modality's native clock and then resample them onto a common 40\,ms grid.
Alignment across modalities relies solely on this shared clock, ensuring that token $t$ spans the exact same 40\,ms interval in every stream.

\section{Implementation details}
\label{sec:impl}

This section provides the full experimental and algorithmic details for \cref{sec:exp_setup}, covering the CounterPoint formulation (\cref{alg:counterpoint}), shared backbone encoders and training schedules (\cref{tab:hparams}), temporal alignment baselines, and evaluation protocols.

\begin{table}[tbp]
\centering
\caption{\textbf{Hyperparameters.} Shared across all aligners unless marked as CounterPoint (CP).}
\label{tab:hparams}
\footnotesize
\setlength{\tabcolsep}{4pt}
\begin{adjustbox}{max width=\linewidth}
\begin{tabular}{ll}
\toprule
Parameter & Value \\
\midrule
Token step & 40\,ms \\
Token dimension $d$ & 256 \\
Point dimension & 512 \\
Encoder & 4 layers, width 256, linear attention \\
Instants $T$ (CP) & 25 per second of window \\
Trajectory dimension $k$ (CP) & 64 \\
Temperature $\tau$ (CP) & learnable, init.\ 0.07 \\
Loss weight $\lambda$ (CP) & 0.5 \\
Optimizer, learning rate & AdamW, $3\times10^{-4}$ \\
Weight decay, warm-up & 0.05, 10\% \\
Batch size & 256 \\
Gradient clipping & norm 2 \\
Epochs & 40 \\
Seeds per fold & 3 \\
\bottomrule
\end{tabular}
\end{adjustbox}
\end{table}

\paragraph{CounterPoint formulation.}
CounterPoint applies to any set of modalities synchronized on a shared clock.
A standard point aligner encodes modality $m$ into per-timestep tokens $h_m(t)\in\mathbb{R}^d$ and pools them into a unit vector $z_m$.
CounterPoint retains this point representation and introduces a parallel trajectory branch.
Specifically, a bias-free linear projection $W_m\in\mathbb{R}^{k\times d}$ maps each token to $y_m(t)=W_mh_m(t)$, serving as the learned counterpart of $g_m$ in a shared $k$-dimensional space.
While the temporal mean $\bar y_m$ estimates the condition layer $Pg_m$ (already captured by $z_m$), subtracting it applies the empirical projection $I-P$ within each recording. This yields the centered trajectory $\hat Y_m=(\hat y_m(1),\dots,\hat y_m(T))\in\mathbb{R}^{T\times k}$ with $\hat y_m(t)=y_m(t)-\bar y_m$.

\looseness=-1 We compare two centered trajectories using the discrete form of the time-locked consistency in \cref{supp:eq-kappa}, written $\kappa(\hat Y_q,\hat Y_c)$.
\looseness=-1 Letting $\mathcal{L}^{qc}_{\text{traj}}$ denote the symmetric InfoNCE loss over time-locked consistency scores for modality pair $(q,c)$, and $\mathcal{L}_{\text{pt}}$ denote the base aligner loss, CounterPoint minimizes the joint objective across all trained modality pairs $\mathcal{P}$:
\begin{equation}
  \mathcal{L}=\mathcal{L}_{\text{pt}}+\frac{\lambda}{|\mathcal{P}|}\sum_{(q,c)\in\mathcal{P}}\mathcal{L}^{qc}_{\text{traj}} .
  \label{supp:eq-loss}
\end{equation}
At inference, given a query across modalities $Q\subseteq\mathcal{M}\setminus\{c\}$ and a candidate $n$ in modality $c$, CounterPoint scores each candidate by combining the point cosine similarity (capturing condition identity) with the trajectory consistency (resolving within-condition events):
\begin{equation}
  s(Q,n)=\sum_{q\in Q}\Big[\alpha_q\,\langle z_q, z_c^{(n)}\rangle+\beta_q\,\kappa\big(\hat Y_q,\hat Y_c^{(n)}\big)\Big] ,
  \label{supp:eq-score}
\end{equation}
where $\alpha_q,\beta_q$ are scalar weights fitted on development data. Setting $\beta_q=0$ recovers the pure point baseline. We denote an aligner trained and scored with CounterPoint by $+$\,CP.

\paragraph{Tasks and metrics.}
Synchronized multi-modal recordings define the ground-truth pairings.
We evaluate three cross-modal retrieval tasks that query one target modality from the remaining two, namely AV$\to$H, HV$\to$A, and HA$\to$V (where A denotes audio, H haptics, and V the visual stream).
Recall at 1 (R@1) is computed across all test candidates, including twin events, with ties counted as misses. We report the mean R@1 across all three tasks (Avg) as well as the mean across the audio--haptic tasks (A--H).
To quantify within-condition discrimination directly, event accuracy (\eacc\ in tables) measures the fraction of test events ranked above their same-drive or same-condition twin event (where random guessing yields 50\%).
The tables report event accuracy as the mean over AV$\to$H and HV$\to$A. On Cluster, the path-only visual target serves as the zero-gap control of \cref{supp:thm-levels} and stays at chance. Averaged over all three tasks, event accuracy of TRIANGLE rises from 55.7\% to 70.7\% with CounterPoint on Cluster and from 68.3\% to 97.4\% on HAVdb.
When a table or figure specifies a single task or sensor pair, metrics are reported for that subset alone.
We evaluate four base point aligners: GRAM~\cite{cicchetti2025gram}, HyperGRAM~\cite{na2026hypergram}, PMRL~\cite{liu2025pmrl}, and TRIANGLE~\cite{cicchetti2025triangle}.

\paragraph{Encoders and tokenization.}
All aligners share identical modality encoders and differ solely in their geometric alignment loss, isolating the effect of the representation.
Each encoder projects input tokens linearly into $d=256$ dimensions, prepends a learnable class token, adds learned positional embeddings, and applies a 4-layer pre-norm Transformer with linear attention (4 heads, MLP expansion ratio 4, dropout 0.1).
On Cluster, audio and haptic tokens group 4 consecutive log-spectrogram frames computed with a 10\,ms hop, yielding one token per 40\,ms step. Audio spectrograms use 64 mel bands, while haptic spectrograms use 128 frequency bins up to 4\,kHz per accelerometer axis.
Visual tokens combine $2\times16\times16$ spatiotemporal tubelets extracted over 8 frames at $112\times112$ resolution with phase-correlation motion surfaces computed between consecutive frames.
On HAVdb, all modalities are tokenized analogously onto the same 40\,ms clock.
For every modality, the pooled point vector $z_m\in\mathbb{R}^{512}$ is obtained by linearly projecting the final class token and normalizing to unit length.

\paragraph{Training and batch construction.}
Every baseline is trained using its official loss and scoring rule on identical data splits and optimization schedules (\cref{tab:hparams}).
Models train for 40 epochs across 3 random seeds per fold using AdamW ($\beta=(0.9,0.98)$, weight decay 0.05, learning rate $3\times10^{-4}$ with a 10\% linear warm-up and cosine decay) and gradient norm clipping at 2.0. The checkpoint achieving the highest development score is selected for testing.

Batch composition reflects the structure of each dataset. On Cluster, each batch of 256 samples comprises 16 textures $\times$ 8 motion conditions $\times$ 2 twin recordings, ensuring that every event encounters its twin as a hard negative during training. For the ablation without twin negatives (\cref{supp:tab-ablation}), we instead sample 16 distinct motion conditions per texture using a single randomly chosen recording from each twin pair, preserving the batch size and same-texture negative count while removing all twin pairs. On HAVdb, each batch of 256 windows pairs two slides for each of 128 random draws of a subject and texture at random temporal offsets, with 48 steps per epoch.

\begin{algorithm}[t]
\caption{CounterPoint on top of a point aligner}
\label{alg:counterpoint}
\small
\begin{algorithmic}[1]
\Require batches of synchronized events, encoders $E_m$ and point loss $\mathcal{L}_{\text{pt}}$ of the base aligner, heads $W_m$
\Ensure trained $E_m$ and $W_m$, weights $\alpha_q,\beta_q$
\For{each batch}
  \State $H_m\gets E_m(\text{signal}_m)$ on the shared clock, $\forall m$
  \State $z_m\gets\text{pool}(H_m)$ \Comment{unchanged}
  \State $\hat Y_m\gets H_mW_m^\top$ minus its temporal mean \Comment{$I-P$}
  \State $\kappa_{ij}\gets\kappa(\hat Y_q^{(i)},\hat Y_c^{(j)})$, $\forall(q,c)\in\mathcal{P}$ \Comment{\cref{supp:eq-kappa}}
  \State update $E_m,W_m$ on $\mathcal{L}_{\text{pt}}+\frac{\lambda}{|\mathcal{P}|}\sum\mathcal{L}^{qc}_{\text{traj}}$ \Comment{\cref{supp:eq-loss}}
\EndFor
\State fit $\alpha_q,\beta_q$ on held-out data
\State \textbf{Retrieval:} index $(z_c^{(n)},\hat Y_c^{(n)})$ once, then rank by $s(Q,n)$ \Comment{\cref{supp:eq-score}}
\end{algorithmic}
\end{algorithm}

\paragraph{Trajectory branch and score calibration.}
\cref{alg:counterpoint} outlines training and retrieval for the trajectory branch.
The linear head $W_m$ operates on the sequence of $T$ output tokens prior to pooling ($T=25$ for a 1\,s window and $T=400$ for a 16\,s window) and projects them to $k=64$ dimensions.
The symmetric trajectory InfoNCE loss $\mathcal{L}_{\text{traj}}^{qc}$ uses label smoothing of 0.1, a learnable temperature $\tau$ initialized at 0.07, and weight $\lambda=0.5$ in \cref{supp:eq-loss}. Sensitivity analyses over $\lambda$, $k$, and computational overhead are reported in \cref{sec:more-rq2}.
At retrieval time, each candidate stores $z_c$ alongside its Frobenius-normalized flattened trajectory $\hat Y_c/\lVert\hat Y_c\rVert_F$, allowing \cref{supp:eq-score} to be evaluated as a sum of inner products in a standard vector index.
To calibrate the combination weights $\alpha_q,\beta_q$ in \cref{supp:eq-score}, we first standardize the point and trajectory similarities by their mean and standard deviation over 2,000 training events, and then fit $\alpha_q,\beta_q$ via L-BFGS to maximize the softmax log-likelihood of the ground-truth target on the development split.

\paragraph{Temporal alignment baselines.}
All temporal alignment baselines in \cref{tab:temporal} are trained and evaluated under the exact same protocol as TRIANGLE\,+\,CP (identical encoders, point loss, loss weight $\lambda=0.5$, calibrated score weights, 5 folds, and 3 seeds), modifying only the trajectory similarity function inside $\mathcal{L}_{\text{traj}}^{qc}$ and \cref{supp:eq-score}:
\begin{itemize}
\item \textbf{Best shift} takes the maximum trajectory cosine similarity across temporal offsets of up to $\pm 4$ tokens ($\pm 160$\,ms), evaluated over the overlapping frames.
\item \textbf{MaxSim} normalizes each token representation to unit length, matches every token in one modality to its maximum-similarity counterpart anywhere in the other modality, averages across tokens, and symmetrizes over both directions.
\item \textbf{Soft-DTW} computes the negative soft dynamic time warping discrepancy (with smoothing 0.1 and pairwise token cost $1-\cos$), normalized by sequence length $T$.
\item \textbf{CAV-MAE Sync} retains the temporal mean in the head output and averages the per-instant cosine similarity across aligned timestamps.
\item \textbf{FATE} uses the same uncentered per-instant cosine score as CAV-MAE Sync and adds its temporal offset classification loss with relative weight 2. Specifically, a central anchor span from one modality is classified against shifted candidate spans from the other modality using soft targets $\cos^2(\pi j/2T_0)$ for shifts $\lvert j\rvert\le T_0$ and zero otherwise, symmetrized across both directions. On the 1\,s windows of Cluster, we scale the geometry proportionally to a 13-token anchor (0.52\,s) evaluated over single-token shifts up to $\pm 6$ tokens with $T_0=3$ tokens (120\,ms).
\end{itemize}

\paragraph{Statistical testing.}
Every comparison pairs CounterPoint with its corresponding base aligner on the exact same fold and seed.
We report means across folds and seeds, ranges across fold means, and Holm-corrected $p$-values over the 12 comparisons per test set (4 aligners $\times$ 3 metrics).
Significance is evaluated via two-sided one-sample $t$-tests on the fold-averaged paired differences. Because splits S3 and S4 contain fewer than 3 folds, their paired units are individual fold $\times$ seed runs. Full test statistics are provided in \cref{sec:more-rq2}.

\paragraph{Measurement protocols.}
We compute three diagnostic measurements on held-out splits to validate our theoretical predictions:
\begin{itemize}
\item \textbf{Empirical event gap (\cref{sec:exp_levels} and \cref{fig:mechanism}):} Defined as the difference between mean same-event consistency $\kappa(x,x)$ and mean twin-event consistency $\kappa(x,x')$, measured across the development textures of all 5 texture folds and 3 seeds (15 runs). For each query event $x$, we evaluate cross-modal consistency against the synchronized target of the same event, its twin event $x'$, and a randomly sampled event $\tilde x$ from a different texture under the identical motion condition. On A--H, these three levels average 0.328, 0.183, and $-0.013$, yielding an event gap of 0.146. On A--V (0.091, 0.091, $-0.004$) and V--H (0.088, 0.087, $-0.005$), the event gap is 0.000.
\item \textbf{Temporal order test (\cref{sec:exp_collapse}):} Evaluated on the development textures of Cluster fold 1 and the 16\,s test windows of HAVdb fold 1 (averaged over 3 seeds). Each input stream is partitioned into its underlying 40\,ms token blocks (and corresponding visual frames) and reordered using a random permutation per event, applied either identically across all modalities (jointly) or independently per modality. We measure the cosine similarity between the point embeddings of the shuffled and unshuffled inputs, comparing it against the cosine similarity between an unshuffled event and its twin.
\item \textbf{Twin score margins (\cref{fig:margindist}):} Computed over all test queries in Cluster fold 1 across 3 seeds. For each audio query, the margin equals the score of the ground-truth event minus that of its twin event, normalized by the standard deviation of scores across all test candidates for that query so that different aligners share a common scale.
\end{itemize}

\section{Additional results}
\label{sec:more}

The results below address the four research questions of \cref{sec:exp} in sequence, with each paragraph opening on its core empirical finding.
\Cref{sec:more-rq1} traces condition collapse from a synthetic ergodic flow to real multi-modal embeddings.
\Cref{sec:more-rq2} demonstrates that CounterPoint yields consistent gains across folds, recall cut-offs, query subsets, and hyperparameter settings with minimal overhead.
\Cref{sec:more-rq3} examines the three correlation levels and the event gap across individual events, training epochs, window lengths, sensor pairs, and dataset regimes.
\Cref{sec:more-rq4} isolates the contribution of each model component through targeted ablations.

%% .....................................................................
\subsection{Point aligners stop at the condition (RQ1)}
\label{sec:more-rq1}

\begin{figure*}[tp]
\centering
\includegraphics[width=\linewidth]{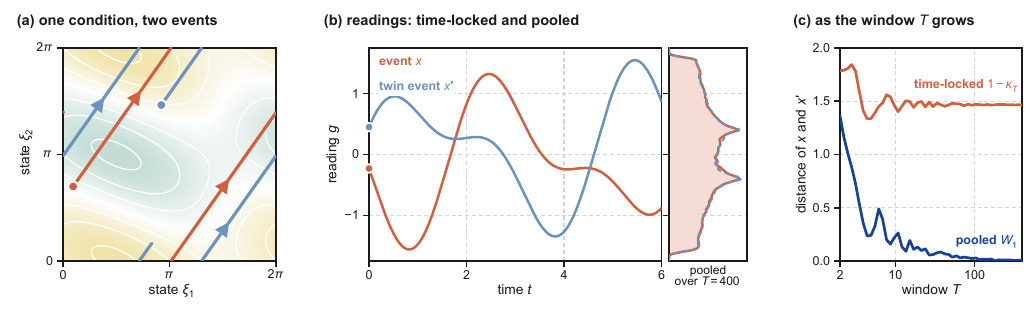}
\caption{\textbf{Condition collapse on an ergodic flow, computed.} The flow and the sensor of \cref{supp:eq-torus}. (a)~The state space of one condition, the torus cut open to the square $[0,2\pi)^2$, shaded by the sensor $g$ (yellow high, teal low) with white level lines. The event $x$ (red) and its twin event $x'$ (blue) start at the dots and move along parallel orbits of slope $\sqrt2$, over the window of (b). (b)~Left: the readings of $x$ and $x'$ at the same instants differ throughout the window. Right: pooled over $T=400$, on the same value axis, their distributions coincide. (c)~As the window grows, the pooled distance $W_1$ of \cref{supp:eq-torus-w1} goes to zero, while the time-locked distance $1-\kappa_T$ of \cref{supp:eq-torus-kappa} stays above 1.3.}
\label{fig:torus}
\end{figure*}

\begin{figure*}[tp]
\centering
\includegraphics[width=\linewidth]{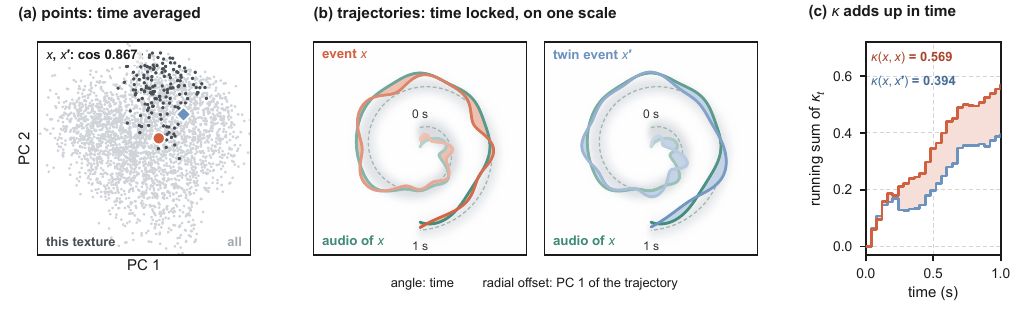}
\caption{\textbf{A Koopman view of one event $x$ and its twin event $x'$.} (a)~Points, the condition layer: the first two principal components of the haptic points of all 3,200 development recordings (light grey), with the 160 recordings of this texture in dark grey. The points of $x$ (red) and $x'$ (blue) land almost on one location. (b)~Trajectories, the dynamics layer: the demeaned trajectories of \cref{supp:eq-empirical-split} on a time spiral, drawn twice on one scale. The angle is time, clockwise from the top over the 1\,s window, the radius grows with time, and the radial offset from the dashed spiral is the first principal coordinate. Left: audio of $x$ (teal) with haptic of $x$ (red). Right: the same audio of $x$ with haptic of $x'$ (blue). Lines darken as time advances, the filled band is the gap between audio and haptic at each instant, and the grey shading is the density of all trajectories of this texture. (c)~The running sum of the per-instant terms $\kappa_t$ of \cref{supp:eq-kappa-t} between the audio trajectory of $x$ and the haptic trajectory of $x$ (red) or of $x'$ (blue). The pair is the 75th percentile of the event gap in the development split.}
\label{fig:koopman}
\end{figure*}

\paragraph{Visualizing condition collapse on an ergodic flow.}
\Cref{fig:torus} illustrates condition collapse (\cref{supp:thm-collapse}) on a canonical ergodic system, the irrational winding flow
\begin{equation}
  S_t(\xi_1,\xi_2)=\big(\xi_1+t,\ \xi_2+\sqrt2\,t\big)\quad\text{on the torus }[0,2\pi)^2 ,
  \label{supp:eq-torus}
\end{equation}
whose unique invariant measure is uniform.
In torus coordinates $(\xi_1,\xi_2)$, a sensor observes $g(\xi_1,\xi_2)=\cos\xi_2+0.6\sin(\xi_1+\xi_2)$.
An event $x$ and its twin $x'$ correspond to two distinct initial states under the same condition, producing parallel trajectories across the unwrapped torus (\cref{fig:torus}a).
Because the two orbits occupy different states at any given instant $t$, the sensor records distinct time-domain waveforms (\cref{fig:torus}b).
\looseness=-1 Over a window of length $T$, an order-invariant point representation compares the empirical occupation measures via the 1-Wasserstein distance
\begin{equation}
  W_1(T)=\int_{\mathbb{R}}\big|F^x_T(s)-F^{x'}_T(s)\big|\,ds ,
  \label{supp:eq-torus-w1}
\end{equation}
where $F^x_T(s)=\frac1T\int_0^T\mathbf 1\big[g(S_tx)\le s\big]\,dt$ is the empirical cumulative distribution function of $x$ and $\mathbf 1[\cdot]$ denotes the indicator function.
\looseness=-1 In contrast, a trajectory representation compares the demeaned signals $\hat g_x(t)=g(S_tx)-\frac1T\int_0^Tg(S_sx)\,ds$ at matching instants through the cosine distance $1-\kappa_T$, where
\begin{equation}
  \kappa_T=\frac{\int_0^T\hat g_x(t)\,\hat g_{x'}(t)\,dt}{\Big(\int_0^T\hat g_x(t)^2\,dt\Big)^{1/2}\Big(\int_0^T\hat g_{x'}(t)^2\,dt\Big)^{1/2}} .
  \label{supp:eq-torus-kappa}
\end{equation}
Because both orbits cover the torus uniformly, their occupation measures converge to the same distribution (\cref{fig:torus}b, right) and $W_1(T)\to0$, causing any pooled representation to merge them.
Conversely, the time-locked distance $1-\kappa_T$ remains above 1.3 across all window lengths (\cref{fig:torus}c), preserving event identity in the temporal ordering.

\begin{figure*}[tp]
\centering
\includegraphics[width=\linewidth]{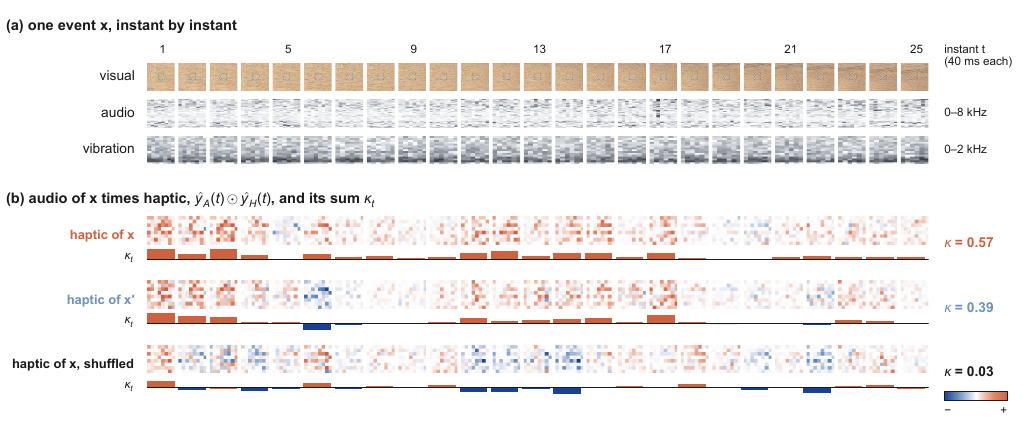}
\caption{\textbf{Agreement accumulates instant by instant.} The event $x$ of \cref{fig:koopman} over its 25 instants of 40\,ms. (a)~The window as recorded, one tile per instant: a square crop of the frame of the visual stream, the texture scan around the probe, and the log spectrograms of audio (0--8\,kHz) and vibration (0--2\,kHz) cut at the instant boundaries. (b)~For each haptic trajectory, the element-wise product $\hat y_A(t)\odot\hat y_H(t)$ with the audio trajectory of $x$ at each instant, as an $8\times8$ map of its 64 coordinates (red: agreement, blue: disagreement), normalized as in \cref{supp:eq-kappa-t}. Each map sums to $\kappa_t$, drawn as the bar below it, and the bars of a row sum to $\kappa$. The haptic trajectory of $x$ adds positive terms at most instants ($\kappa=0.57$), that of $x'$ at fewer ($0.39$), and $x$ shuffled, its haptic instants permuted, cancels out ($0.03$), although the shuffle keeps the set of instants, and with it every order-invariant point of \cref{supp:eq-shuffle}, unchanged.}
\label{fig:instant}
\end{figure*}

\paragraph{The Koopman decomposition on real recordings.}
Real sensor recordings show the same behavior.
\Cref{fig:koopman} examines a development event $x$ and its twin $x'$ through the empirical counterpart of the Koopman decomposition $g_m=Pg_m+\hat g_m$ from \cref{supp:eq-threelayers}.
Over a window of $T=25$ instants, the linear projection $y_m(t)$ splits into
\begin{equation}
  y_m(t)=\underbrace{\bar y_m}_{\text{condition layer}}+\underbrace{\hat y_m(t)}_{\text{dynamics layer}} ,
  \qquad
  \bar y_m=\frac1T\sum_{s=1}^{T}y_m(s) ,
  \label{supp:eq-empirical-split}
\end{equation}
where the temporal mean $\bar y_m$ estimates $Pg_m$ and the centered trajectory $\hat y_m(t)$ estimates $\hat g_m$.

\looseness=-1 Because pooled points discard temporal order (\cref{sec:exp_collapse}), they act as functions $\Psi(\nu_T^x)$ of the occupation measure and retain primarily the condition layer.
Within the development embedding space, the points of $x$ and $x'$ therefore land nearly on top of each other with a cosine similarity of 0.867 (\cref{fig:koopman}a).
\looseness=-1 The same collapse appears in the projected condition layer: across all 1,600 development pairs, the haptic temporal means $\bar y_H$ of an event and its twin achieve a median cosine similarity of 0.94, compared to 0.13 for a recording from another texture, matching \cref{supp:eq-collapse}.

Trajectories preserve temporal order and separate the two events.
Unrolled along a time spiral, the audio and haptic trajectories of $x$ track each other closely at every instant, whereas the haptic trajectory of $x'$ diverges, keeping the cross-modal band tight for $x$ and wide for $x'$ (\cref{fig:koopman}b).
Time-locked consistency sums this instant-by-instant alignment across the window,
\begin{equation}
  \kappa(\hat Y_A,\hat Y_H)=\sum_{t=1}^{T}\kappa_t ,
  \qquad
  \kappa_t=\frac{\langle\hat y_A(t),\hat y_H(t)\rangle}{\lVert\hat Y_A\rVert\,\lVert\hat Y_H\rVert} ,
  \label{supp:eq-kappa-t}
\end{equation}
where $\lVert\hat Y_m\rVert$ denotes the Frobenius norm of the full trajectory.
Comparing the audio trajectory of $x$ against both haptic candidates, positive terms accumulate steadily over the window for $x$ to reach $\kappa(x,x)=0.569$, while the twin event reaches only $\kappa(x,x')=0.394$ (\cref{fig:koopman}c).
This example sits at the 75th percentile of the event gap, and the trend holds across the entire development split (fold~1, seed~0). Scoring on the condition layer ($\cos(\bar y_A,\bar y_H)$) ranks $x$ above $x'$ on 57\% of pairs with a median margin of 0.011, whereas scoring on the dynamics layer ($\kappa$) ranks $x$ above $x'$ on 82\% of pairs with a median margin of 0.094. (Multi-run averages over 15 runs are reported in \cref{sec:exp_levels} and \cref{sec:impl}.)

\begin{figure}[tbp]
\centering
\includegraphics[width=\linewidth]{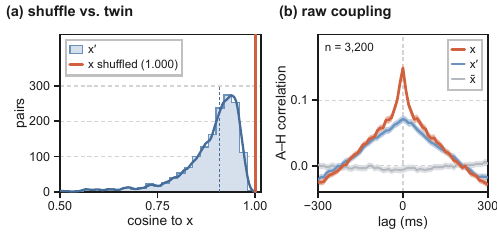}
\caption{\textbf{Pooling collapses to the condition, and the event survives in the dynamics.} The two measured sides of \cref{supp:thm-collapse}, whose computed case is \cref{fig:torus}.
(a)~In a trained TRIANGLE on Cluster, the haptic point cosine to $x$ of its twin event $x'$ as a histogram over the 1,600 development pairs (dashed: median), and of $x$ with its 25 blocks of 40\,ms shuffled, which stays at 1.000 for every pair, above every real $x'$.
(b)~Without any model, the cross-correlation of the audio envelope of $x$ with the haptic envelope of $x$, of $x'$ and of an event $\tilde x$ of another texture, at 10\,ms steps, each envelope with the mean envelope of its motion condition subtracted. Lines are the means over the 3,200 development queries, bands their 95\% confidence intervals. The coupling peaks at zero lag, highest for $x$, lower for $x'$, and absent for $\tilde x$.}
\label{fig:layers}
\end{figure}

\Cref{fig:instant} breaks down the time-locked consistency of this pair across all 25 instants, displaying each normalized element-wise product $\langle\hat y_A(t),\hat y_H(t)\rangle/(\lVert\hat Y_A\rVert\,\lVert\hat Y_H\rVert)$ as an $8\times8$ grid with its sum $\kappa_t$ plotted below.
The true event accumulates positive alignment across most frames, whereas its twin does so sporadically.
Permuting the haptic instants collapses the total consistency to 0.03 while keeping the set of instants unchanged, and with it the occupation measure that determines every order-invariant point (\cref{supp:eq-shuffle}).

\begin{figure}[tbp]
\centering
\includegraphics[width=\linewidth]{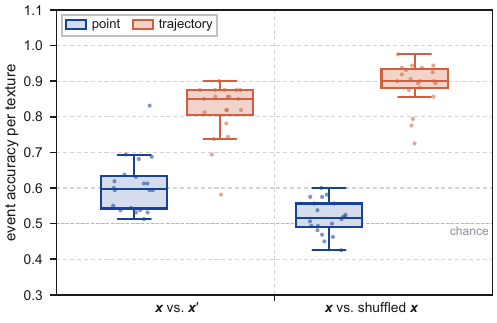}
\caption{\textbf{Condition collapse, measured.} Per development texture, how often an event $x$ outscores $x'$ (event accuracy) or its own time-shuffled copy, under the point (blue) and the trajectory (red) of TRIANGLE\,+\,CP. Each dot is one of the 20 development textures. The point stays near chance against the shuffled copy (52\%) and near 60\% against the twin event, while the trajectory of the same model rises far above both.}
\label{fig:collapse}
\end{figure}

\begin{figure}[tbp]
\centering
\includegraphics[width=\linewidth]{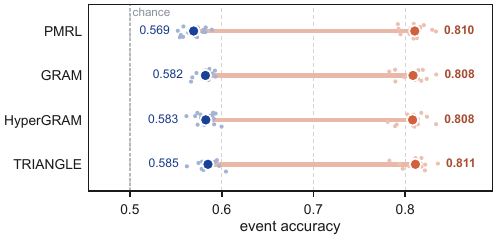}
\caption{\textbf{Escaping condition collapse.} Event accuracy, how often $x$ outscores $x'$, for each aligner (blue) and the same aligner\,+\,CP (red). Large dots and numbers: means. Small dots: the 15 runs (5 folds $\times$ 3 seeds). Dotted line: chance. Every point aligner stays below 0.59, and every one reaches 0.81 with CounterPoint.}
\label{fig:winbars}
\end{figure}

\begin{figure}[tbp]
\centering
\includegraphics[width=\linewidth]{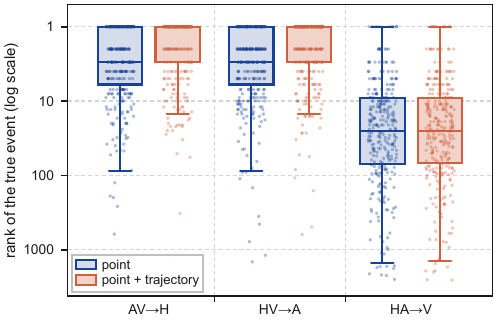}
\caption{\textbf{Ranks before and after the trajectory score.} Rank of the true event per query of TRIANGLE\,+\,CP, point alone (blue) and point\,+\,trajectory (red), 5 folds $\times$ 3 seeds. Each box covers all 47,040 queries of a task, and the dots are 300 of them drawn at random, the same queries in both boxes. For AV$\to$H and HV$\to$A, the median rank of the true event improves to 1 when the trajectory score is added. HA$\to$V, whose path-only visual stream has no event gap, is the control and keeps its rank.}
\label{fig:rankbox}
\end{figure}

% results/order_invariance.csv (Cluster S1 fold 1, development, 3 seeds) and results/havdb_order.csv (HAVdb fold 1, 16 s test, 3 seeds)
\begin{table*}[tp]
\centering
\caption{\textbf{Every point aligner discards the order of the instants.} Cosine of the point of a shuffled input to the point of the original input (Shuf.), with its 40\,ms blocks permuted jointly or independently across modalities (the two give the same values), against the cosine of the points of an event and of its twin event (Twin). Cluster: development textures of the first fold, audio and haptics, whose tokens each cover one 40\,ms block. HAVdb: 16\,s test windows of the first fold. Mean over 3 seeds.}
\label{tab:order}
\footnotesize
\setlength{\tabcolsep}{4pt}
\begin{adjustbox}{max width=\linewidth}
\begin{tabular}{lllllllll}
\toprule
 & \multicolumn{2}{l}{A} & \multicolumn{2}{l}{H or \Hv} & \multicolumn{2}{l}{\Hf} & \multicolumn{2}{l}{V} \\
\cmidrule(lr){2-3}\cmidrule(lr){4-5}\cmidrule(lr){6-7}\cmidrule(lr){8-9}
 & Shuf. & Twin & Shuf. & Twin & Shuf. & Twin & Shuf. & Twin \\
\midrule
\multicolumn{9}{l}{\textit{Cluster}} \\
\quad GRAM & 1.000 & 0.928 & 1.000 & 0.914 & & & & \\
\quad HyperGRAM & 1.000 & 0.919 & 1.000 & 0.897 & & & & \\
\quad PMRL & 1.000 & 0.842 & 1.000 & 0.694 & & & & \\
\quad TRIANGLE & 1.000 & 0.909 & 1.000 & 0.874 & & & & \\
\quad TRIANGLE\,+\,CP & 0.999 & 0.910 & 0.999 & 0.879 & & & & \\
\midrule
\multicolumn{9}{l}{\textit{HAVdb}} \\
\quad TRIANGLE & 1.000 & 0.716 & 1.000 & 0.768 & 1.000 & 0.750 & 1.000 & 0.632 \\
\quad TRIANGLE\,+\,CP & 1.000 & 0.679 & 1.000 & 0.720 & 1.000 & 0.712 & 0.999 & 0.581 \\
\bottomrule
\end{tabular}
\end{adjustbox}
\end{table*}

\paragraph{Separating condition and event layers within a single model.}
\Cref{fig:layers}a quantifies layer collapse inside a trained point aligner, while \cref{fig:layers}b isolates the event layer directly in raw sensor envelopes before any learning.
Scoring raw audio envelopes against vibration envelopes via zero-lag correlation ranks the true event above its twin on 64.6\% of development queries, whereas shifting candidates by 160\,ms drops accuracy to 50.6\% (chance).
\looseness=-1 \Cref{fig:collapse} compares point and trajectory scores per development texture in our trained model. The point score ranks an event above a time-shuffled copy of itself on 52\% of the 1,600 development pairs and above its twin on 61\%, whereas the trajectory score succeeds on 89\% and 82\%, respectively.

\begin{figure}[tbp]
\centering
\includegraphics[width=\linewidth]{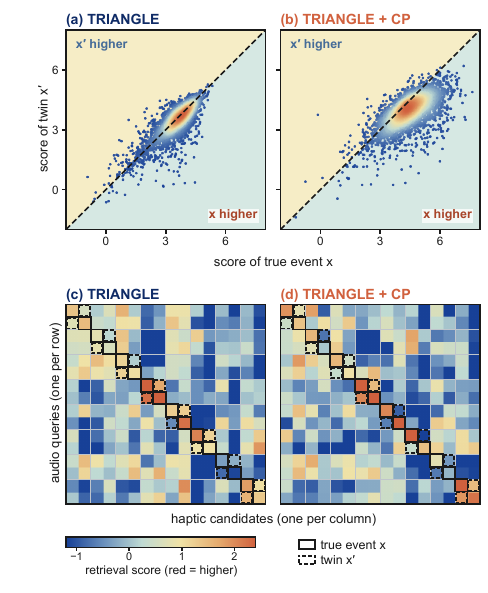}
\caption{\textbf{Points tie the true recording with its twin event on Cluster.} Here the twin event $x'$ is the same stroke made again by the machine on the same texture with the same path, speed and force. S1 test, fold~1. Panels as in \cref{fig:visual}, with the 3,200 audio-to-haptic queries in (a, b), colored by density, and one motion condition in (c, d). The point aligner keeps the queries on the diagonal and lights whole texture blocks. CounterPoint moves the cloud below the diagonal and lights the true recording.}
\label{fig:visualcluster}
\end{figure}

\begin{figure*}[tp]
\centering
\includegraphics[width=\linewidth]{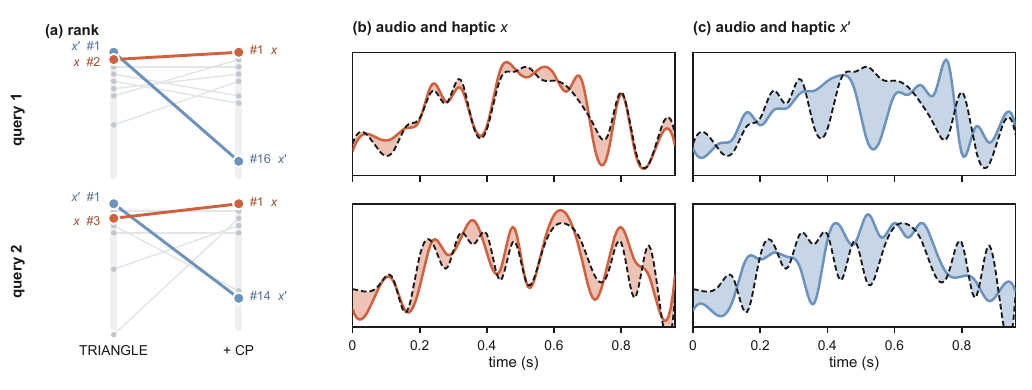}
\caption{\textbf{Qualitative A$\to$H retrieval.} Two audio queries of Cluster S1 fold~1 (seed~0) for which TRIANGLE ranks its twin event $x'$ first and TRIANGLE\,+\,CP ranks the event $x$ first (316 of the 547 queries in which TRIANGLE ranks $x'$ first, of which the two with the largest margins are shown). (a)~Rank of $x$ (red), $x'$ (blue) and the other candidates of either top~5 (grey). (b,\,c)~The audio trajectory of the query (dashed) and the haptic trajectory of $x$ or $x'$ on the first principal axis of the query trajectory, each scaled to unit maximum, the band between them filled.}
\label{fig:qualitative}
\end{figure*}

\begin{figure*}[tp]
\centering
\includegraphics[width=\linewidth]{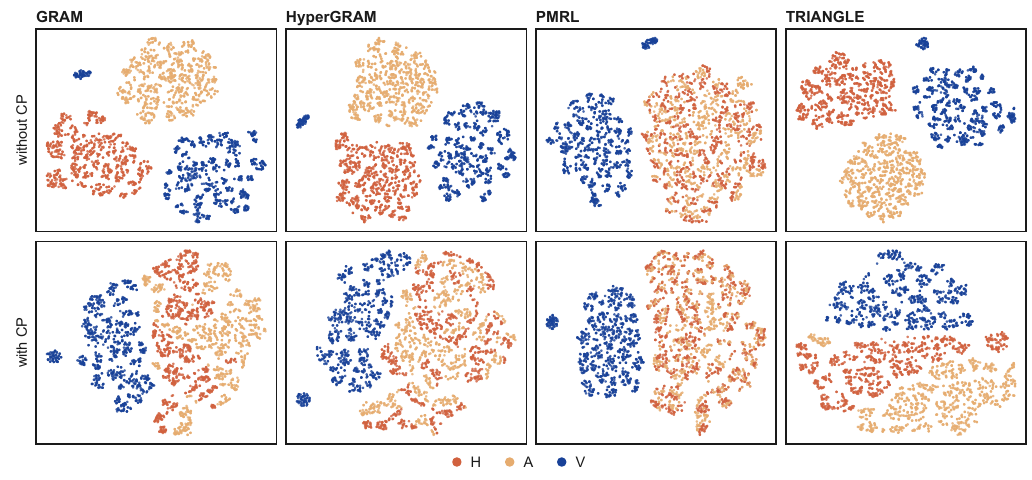}
\caption{\textbf{t-SNE of the multimodal representations on Cluster.} With CounterPoint (bottom), the audio and haptic representations of every aligner move next to each other, while the visual one, which observes only the motion path, stays apart.}
\label{fig:tsnecluster}
\end{figure*}

\begin{figure*}[tp]
\centering
\includegraphics[width=\linewidth]{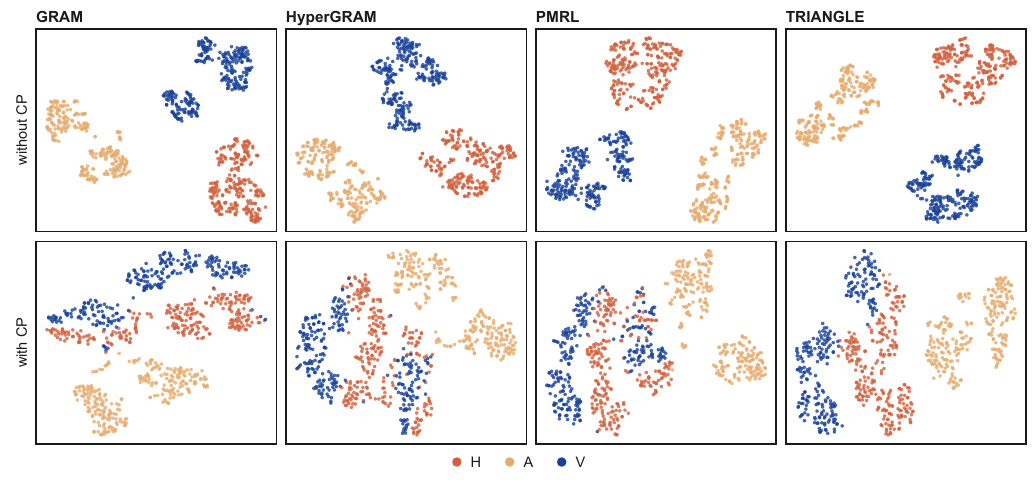}
\caption{\textbf{t-SNE of the multimodal representations on HAVdb.} As in \cref{fig:tsnecluster}, on all 267 test windows of 16\,s (fold~1, seed~0), where H combines vibration and force and V is the tracked fingertip. With CounterPoint (bottom), the haptic and visual representations of every aligner move next to each other. On HAVdb the tracked fingertip follows the free hand and carries its motion dynamics, and force and position form the sensor pair with the highest trajectory MRR (\cref{sec:more-rq3}).}
\label{fig:tsnehavdb}
\end{figure*}

\begin{figure*}[tp]
\centering
\includegraphics[width=\linewidth]{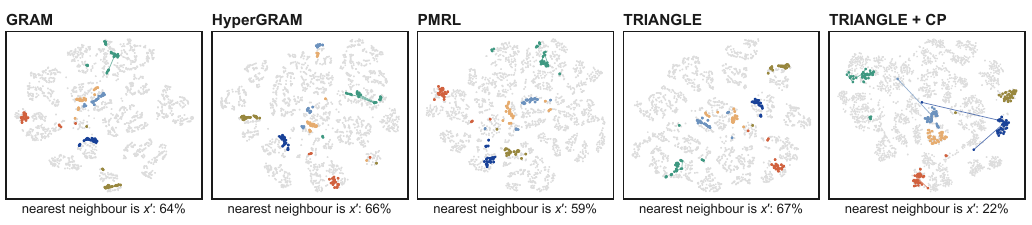}
\caption{\textbf{t-SNE of haptic representations.} Six motion conditions are colored, the rest are grey, and a segment joins every $x$ to its $x'$. Every point aligner forms one tight cluster per condition, in which an event $x$ and its twin event $x'$ nearly coincide. Below each panel: share of test events $x$ whose nearest haptic neighbor is $x'$. The trajectories of TRIANGLE\,+\,CP keep the clusters but separate the two (\cref{fig:tsnezoom} zooms into one cluster).}
\label{fig:tsne}
\end{figure*}

\begin{figure*}[tp]
\centering
\includegraphics[width=\linewidth]{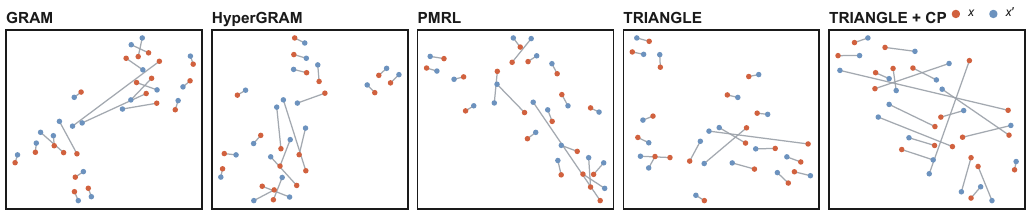}
\caption{\textbf{Inside one condition cluster of \cref{fig:tsne}.} t-SNE of the haptic representations of a single motion condition (20 held-out textures, $x$ and $x'$ of each, fold~1, seed~0), with $x$ in red and $x'$ in blue. A segment joins every $x$ to its $x'$. The evaluated point aligners place $x'$ close to $x$, illustrating their weak event separation. The trajectories of TRIANGLE\,+\,CP place them far apart, since their fluctuations differ.}
\label{fig:tsnezoom}
\end{figure*}

\paragraph{Order invariance holds across all point aligners.}
\Cref{tab:order} extends the token-shuffling test of \cref{sec:exp_collapse} to all four base aligners and to HAVdb.
Even though every encoder adds learned positional embeddings to its input tokens (\cref{sec:impl}), the resulting pooled points remain virtually invariant to temporal permutation.
On Cluster, shuffling token blocks either jointly across modalities or independently per modality leaves audio and haptic point embeddings at a cosine similarity of at least 0.999 with the unshuffled embedding, well above the similarity between an event and its twin (0.694 to 0.928).
On HAVdb, every modality (including the tracked fingertip) maintains a shuffled-to-original cosine similarity of at least 0.999, compared to at most 0.768 for twin events.
Because the point branch of TRIANGLE\,+\,CP exhibits the exact same invariance, CounterPoint preserves the point representation intact while capturing temporal structure exclusively through the trajectory branch.

\paragraph{Score distributions and retrieval rankings.}
\Cref{fig:winbars} shows that event accuracy remains below 59\% for all four point aligners, but rises to between 80.8\% and 81.1\% when augmented with CounterPoint.
In the per-query rank distributions (\cref{fig:rankbox}), adding the trajectory score brings the median rank of the ground-truth event to 1 on both AV$\to$H and HV$\to$A while leaving visual targets unchanged.
Score heatmaps (\cref{fig:visual} in the main paper and \cref{fig:visualcluster} on Cluster) illustrate this separation directly. On Cluster fold~1, TRIANGLE ranks the true recording above its twin on 58\% of A$\to$H queries (11 of 16 rows in the single-condition matrix of \cref{fig:visualcluster}c), whereas CounterPoint does so on 78\% of queries (15 of 16 rows in \cref{fig:visualcluster}d).
Inspecting individual queries confirms the mechanism: out of 3,200 A$\to$H queries in fold~1, TRIANGLE ranks the twin event first 547 times, and CounterPoint corrects 316 of those errors to rank the true event first.
\Cref{fig:qualitative} visualizes two such cases where the twin event favored by TRIANGLE drops out of the top ranks under CounterPoint as the true haptic trajectory tracks the audio query along the shared clock.

\paragraph{Geometry of the learned embeddings.}
\Cref{fig:tsnecluster} extends the modality-colored t-SNE projections of \cref{fig:tsnemod} to all four aligners on Cluster, and \cref{fig:tsnehavdb} shows the corresponding projections on HAVdb. \Cref{fig:tsne} visualizes haptic representations across the full test fold, and \cref{fig:tsnezoom} zooms into a single motion condition.
Every point aligner organizes embeddings into tight motion-condition clusters containing one sub-cluster per texture, inside which an event and its twin collapse onto nearly identical coordinates.
CounterPoint preserves this macro-level condition structure while separating twin pairs within each sub-cluster. Across the entire test set, a recording's twin is its nearest haptic neighbor only 22\% of the time under CounterPoint, compared to 59--67\% for point aligners.

%% .....................................................................
\FloatBarrier
\subsection{CounterPoint improves event retrieval (RQ2)}
\label{sec:more-rq2}

\begin{figure}[tbp]
\centering
\includegraphics[width=\linewidth]{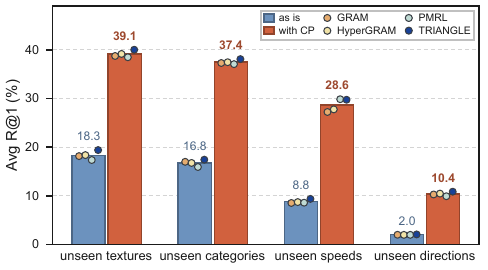}
\caption{\textbf{Generalization to unseen conditions.} Avg R@1 on Cluster with textures, categories, speeds or directions held out from training. Bars: mean of the four aligners as is (blue) and with CounterPoint (red). Dots: each aligner.}
\label{fig:splits}
\end{figure}
% generated by scripts/nov_tables.py; every cell is a percentage
\begin{table}[tbp]
\centering
\caption{\textbf{Generalization to unseen conditions on Cluster.} Avg R@1 (\%, mean of AV$\to$H, HV$\to$A and HA$\to$V) when whole textures, categories, speeds or directions are held out (5, 9, 2 and 1 folds, 3 seeds each). Superscripts: percentage-point gains over the base aligner.}
\label{tab:splits}
\footnotesize
\setlength{\tabcolsep}{3.3pt}
\begin{adjustbox}{max width=\linewidth}
\begin{tabular}{llllll}
\toprule
Held out & Texture & Category & Speed & Direction & Avg \\
\midrule
GRAM         & 18.1 & 17.0 & 8.5 & 2.0 & 11.4 \\
HyperGRAM    & 18.4 & 16.7 & 8.7 & 2.0 & 11.4 \\
PMRL         & 17.3 & 15.9 & 8.6 & 2.0 & 11.0 \\
TRIANGLE     & 19.4 & 17.4 & 9.4 & 2.1 & 12.1 \\
\midrule
GRAM\cpup    & 38.7\up{20.6} & 37.3\up{20.3} & 27.2\up{18.7} & 10.3\up{8.3} & 28.4\up{16.9} \\
HyperGRAM\cpup& 39.1\up{20.8} & 37.4\up{20.7} & 27.7\up{19.0} & 10.5\up{8.5} & 28.7\up{17.2} \\
PMRL\cpup    & 38.4\up{21.1} & 37.0\up{21.1} & \textbf{29.8}\up{21.2} & 9.9\up{7.9} & 28.8\up{17.8} \\
TRIANGLE\cpup& \textbf{40.0}\up{20.6} & \textbf{38.1}\up{20.6} & 29.7\up{20.4} & \textbf{10.8}\up{8.8} & \textbf{29.7}\up{17.6} \\
\bottomrule
\end{tabular}
\end{adjustbox}
\end{table}

% generated by scripts/nov_tables.py; every cell is a percentage
\begin{table*}[tp]
\centering
\caption{\textbf{Per-fold results on Cluster.} Avg R@1 (\%) on the test fold, mean over 3 seeds. Bold: best.
Category folds hold out one category (abbreviated), speed folds interpolate (40\,mm/s) or extrapolate (60\,mm/s), the direction fold trains on axial strokes and tests on diagonal ones.}
\label{tab:perfold-cluster}
\footnotesize
\setlength{\tabcolsep}{2.4pt}
\begin{adjustbox}{max width=\linewidth}
\begin{tabular}{llllllllllllllllll}
\toprule
 & \multicolumn{5}{l}{Texture (S1)} & \multicolumn{9}{l}{Category (S2)} & \multicolumn{2}{l}{Speed (S3)} & Dir.\ (S4) \\
\cmidrule(lr){2-6}\cmidrule(lr){7-15}\cmidrule(lr){16-17}\cmidrule(lr){18-18}
 & 1 & 2 & 3 & 4 & 5 & cera & clot & comp & glas & leat & meta & poly & ston & wood & interp. & extrap. & diag. \\
\midrule
GRAM         & 17.3 & 19.6 & 17.4 & 19.0 & 17.5 & 24.9 & 14.2 & 33.9 & 17.1 & 13.3 & 14.2 & 9.6 & 19.9 & 5.8 & 9.4 & 7.6 & 2.0 \\
HyperGRAM    & 17.9 & 20.1 & 17.6 & 19.4 & 16.8 & 24.6 & 13.8 & 33.1 & 16.8 & 13.5 & 14.4 & 9.4 & 19.4 & 5.8 & 10.1 & 7.3 & 2.0 \\
PMRL         & 16.7 & 18.6 & 16.5 & 18.2 & 16.6 & 22.8 & 12.8 & 32.1 & 15.3 & 13.3 & 13.8 & 9.6 & 18.2 & 5.2 & 9.6 & 7.6 & 2.0 \\
TRIANGLE     & 18.7 & 20.7 & 18.5 & 20.8 & 18.3 & 26.4 & 14.2 & 34.5 & 17.3 & 14.2 & 14.6 & 9.7 & 19.7 & 6.0 & 11.2 & 7.6 & 2.1 \\
\midrule
GRAM\cpup    & 35.6 & 39.6 & 39.0 & 41.3 & 38.0 & 49.2 & 34.9 & 52.1 & \textbf{40.2} & 26.1 & 37.6 & 25.0 & \textbf{47.3} & 23.0 & 24.5 & 29.9 & 10.3 \\
HyperGRAM\cpup& 36.0 & 40.3 & 39.8 & 41.7 & 37.8 & 50.1 & 34.9 & \textbf{52.9} & 39.7 & 26.8 & 37.1 & 24.6 & 47.0 & 23.7 & 25.8 & 29.7 & 10.5 \\
PMRL\cpup    & 35.8 & 39.5 & 38.6 & 40.9 & 37.5 & 49.0 & 34.9 & 51.1 & 38.9 & \textbf{27.6} & 36.6 & \textbf{25.4} & 46.4 & 23.5 & 26.7 & \textbf{32.9} & 9.9 \\
TRIANGLE\cpup& \textbf{37.3} & \textbf{41.3} & \textbf{40.1} & \textbf{42.8} & \textbf{38.4} & \textbf{52.1} & \textbf{36.0} & 52.2 & 39.4 & 27.2 & \textbf{38.0} & 25.3 & \textbf{47.3} & \textbf{25.1} & \textbf{27.7} & 31.8 & \textbf{10.8} \\
\bottomrule
\end{tabular}
\end{adjustbox}
\end{table*}

\begin{table*}[tp]
\centering
\caption{\textbf{Per-fold results on HAVdb.} Main task R@1 (\%, mean of AV$\to$H and HV$\to$A, H combining vibration and force) on the unseen subjects of each fold, mean over 3 seeds. Bold: best.}
\label{tab:perfold-havdb}
\footnotesize
\setlength{\tabcolsep}{2.6pt}
\begin{adjustbox}{max width=\linewidth}
\begin{tabular}{llllllllllllllll}
\toprule
 & \multicolumn{5}{l}{1\,s} & \multicolumn{5}{l}{4\,s} & \multicolumn{5}{l}{16\,s} \\
\cmidrule(lr){2-6}\cmidrule(lr){7-11}\cmidrule(lr){12-16}
 & 1 & 2 & 3 & 4 & 5 & 1 & 2 & 3 & 4 & 5 & 1 & 2 & 3 & 4 & 5 \\
\midrule
GRAM         & 21.4 & 13.3 & 22.5 & 23.6 & 24.1 & 17.7 & 10.3 & 22.8 & 23.4 & 25.8 & 10.4 & 4.5 & 12.2 & 10.9 & 11.2 \\
HyperGRAM    & 19.2 & 11.7 & 20.8 & 23.2 & 24.4 & 14.9 & 10.2 & 16.7 & 19.3 & 24.9 & 7.8 & 4.0 & 10.7 & 11.0 & 10.8 \\
PMRL         & 20.2 & 11.1 & 18.7 & 22.7 & 22.2 & 14.5 & 7.5 & 13.5 & 16.4 & 18.8 & 7.9 & 3.5 & 8.1 & 9.3 & 9.0 \\
TRIANGLE     & 20.3 & 11.2 & 19.7 & 23.4 & 23.8 & 15.9 & 8.9 & 17.1 & 19.9 & 26.1 & 10.1 & 4.0 & 9.1 & 10.7 & 10.8 \\
\midrule
GRAM\cpup    & \textbf{36.5} & \textbf{28.5} & \textbf{48.6} & \textbf{43.6} & \textbf{44.0} & 73.5 & 60.6 & \textbf{76.0} & 78.5 & 74.7 & 87.8 & \textbf{78.5} & \textbf{90.6} & 86.3 & 84.9 \\
HyperGRAM\cpup& \textbf{36.5} & 26.6 & 48.1 & 41.6 & 42.1 & 74.6 & \textbf{62.3} & 73.0 & \textbf{79.0} & 74.0 & 87.7 & 78.0 & 89.6 & \textbf{86.5} & 85.5 \\
PMRL\cpup    & 32.3 & 20.6 & 43.8 & 36.1 & 41.9 & 71.2 & 60.4 & 73.5 & 75.5 & 72.9 & \textbf{87.9} & 77.1 & 87.7 & 85.9 & 84.8 \\
TRIANGLE\cpup& 36.4 & 25.8 & 47.9 & 40.8 & 42.3 & \textbf{75.5} & 62.0 & 74.7 & 78.8 & \textbf{75.4} & 87.1 & 75.1 & 87.5 & 85.9 & \textbf{86.1} \\
\bottomrule
\end{tabular}
\end{adjustbox}
\end{table*}

% generated from results/significance.csv (scripts/significance.py); every cell is in percentage points
\begin{table*}[tp]
\centering
\caption{\textbf{Every gain is significant.} Paired difference of each aligner\,+\,CP and the same aligner on the same fold and seed, in percentage points: mean and 95\% $t$-interval across the units of each test set. The units are the folds, with the 3 seeds averaged, or the fold $\times$ seed runs for S3 and S4, which have fewer than 3 folds. Min: the smallest difference of any unit over the four aligners. $p$: the largest of the four Holm-corrected $p$-values, corrected over the 12 comparisons of each test set.}
\label{tab:signif}
\footnotesize
\setlength{\tabcolsep}{4pt}
\begin{adjustbox}{max width=\linewidth}
\begin{tabular}{lllllll}
\toprule
 & GRAM & HyperGRAM & PMRL & TRIANGLE & Min & $p$ \\
\midrule
\multicolumn{7}{l}{\textit{Cluster S1, textures} (5 folds)} \\
\quad A--H R@1 & 30.8 [27.7, 33.8] & 31.1 [27.9, 34.4] & 31.5 [29.0, 34.0] & 30.9 [28.4, 33.3] & 27.0 & $3\cdot10^{-5}$ \\
\quad Avg R@1 & 20.6 [18.6, 22.5] & 20.8 [18.6, 22.9] & 21.1 [19.4, 22.8] & 20.6 [18.9, 22.3] & 18.1 & $3\cdot10^{-5}$ \\
\quad \eacc & 22.6 [21.3, 23.9] & 22.5 [21.1, 23.9] & 24.1 [22.1, 26.1] & 22.6 [21.3, 23.9] & 21.6 & $3\cdot10^{-5}$ \\
\midrule
\multicolumn{7}{l}{\textit{Cluster S2, categories} (9 folds)} \\
\quad A--H R@1 & 30.3 [24.9, 35.8] & 30.9 [25.6, 36.2] & 31.7 [26.5, 36.9] & 30.9 [25.6, 36.3] & 18.9 & $6\cdot10^{-6}$ \\
\quad Avg R@1 & 20.3 [16.7, 23.9] & 20.7 [17.1, 24.3] & 21.1 [17.6, 24.7] & 20.6 [17.0, 24.3] & 12.8 & $6\cdot10^{-6}$ \\
\quad \eacc & 23.5 [20.9, 26.1] & 23.5 [20.9, 26.2] & 24.7 [22.1, 27.2] & 23.7 [21.1, 26.2] & 16.2 & $3\cdot10^{-7}$ \\
\midrule
\multicolumn{7}{l}{\textit{Cluster S3, speeds} (6 runs)} \\
\quad A--H R@1 & 27.9 [21.5, 34.4] & 28.3 [22.2, 34.4] & 31.6 [24.3, 38.8] & 30.3 [23.5, 37.1] & 20.8 & $6\cdot10^{-4}$ \\
\quad Avg R@1 & 18.7 [14.4, 23.0] & 19.0 [15.0, 23.1] & 21.2 [16.5, 26.0] & 20.4 [15.9, 24.8] & 13.9 & $6\cdot10^{-4}$ \\
\quad \eacc & 25.7 [21.7, 29.7] & 25.6 [21.6, 29.5] & 28.8 [25.6, 32.0] & 26.5 [22.3, 30.7] & 21.8 & $2\cdot10^{-4}$ \\
\midrule
\multicolumn{7}{l}{\textit{Cluster S4, directions} (3 runs)} \\
\quad A--H R@1 & 12.4 [8.6, 16.2] & 12.7 [8.6, 16.7] & 11.9 [9.5, 14.2] & 13.2 [10.6, 15.7] & 10.9 & 0.020 \\
\quad Avg R@1 & 8.3 [5.7, 10.8] & 8.5 [5.8, 11.2] & 7.9 [6.5, 9.4] & 8.8 [7.0, 10.5] & 7.3 & 0.020 \\
\quad \eacc & 19.6 [18.2, 21.1] & 19.9 [18.9, 20.9] & 19.6 [18.2, 21.0] & 20.1 [18.0, 22.2] & 19.0 & 0.005 \\
\midrule
\multicolumn{7}{l}{\textit{HAVdb, subjects} (5 folds)} \\
\quad A--H R@1 & 75.8 [73.2, 78.4] & 76.6 [73.3, 79.9] & 77.1 [73.8, 80.5] & 75.4 [72.0, 78.8] & 71.1 & $2\cdot10^{-6}$ \\
\quad Avg R@1 & 78.8 [77.1, 80.5] & 79.8 [77.1, 82.4] & 80.2 [77.0, 83.4] & 78.8 [75.7, 81.8] & 75.5 & $2\cdot10^{-6}$ \\
\quad \eacc & 28.1 [23.7, 32.4] & 31.6 [27.8, 35.5] & 31.7 [27.9, 35.5] & 30.5 [26.9, 34.0] & 24.0 & $7\cdot10^{-5}$ \\
\bottomrule
\end{tabular}
\end{adjustbox}
\end{table*}

\begin{figure}[tbp]
\centering
\includegraphics[width=\linewidth]{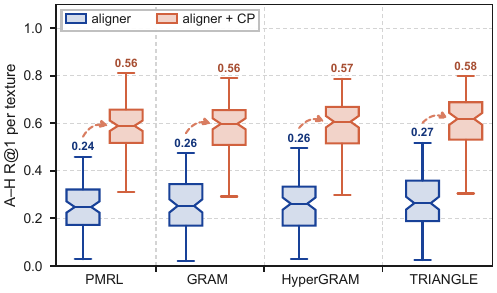}
\caption{\textbf{The gain holds for every held-out texture.} A--H R@1 per held-out texture (mean over 3 seeds) of each aligner (blue) and of the same aligner\,+\,CP (red). Numbers: means over the 98 held-out textures. Every aligner improves on each of the 98 textures, and its mean rises from 0.24--0.27 to 0.56--0.58.}
\label{fig:texturebox}
\end{figure}

% generated by scripts/nov_tables.py; every cell is a percentage
\begin{table}[tbp]
\centering
\caption{\textbf{Recall at 1, 5 and 10, and MRR (\%).} Cluster S1: mean of AV$\to$H and HV$\to$A over 5 folds $\times$ 3 seeds. HAVdb 16\,s: mean of AV$\to$H and HV$\to$A, H combining vibration and force over 5 folds $\times$ 3 seeds. Bold: best.}
\label{tab:recall}
\footnotesize
\setlength{\tabcolsep}{2.2pt}
\begin{adjustbox}{max width=\linewidth}
\begin{tabular}{lllllllll}
\toprule
 & \multicolumn{4}{l}{Cluster S1} & \multicolumn{4}{l}{HAVdb 16\,s} \\
\cmidrule(lr){2-5}\cmidrule(lr){6-9}
 & R@1 & R@5 & R@10 & MRR & R@1 & R@5 & R@10 & MRR \\
\midrule
GRAM         & 25.6 & 67.5 & 82.2 & 44.1 & 9.8 & 27.2 & 38.5 & 19.4 \\
HyperGRAM    & 25.9 & 67.7 & 82.1 & 44.3 & 8.9 & 24.5 & 35.6 & 17.6 \\
PMRL         & 24.5 & 64.8 & 79.8 & 42.4 & 7.5 & 21.8 & 31.5 & 15.7 \\
TRIANGLE     & 27.4 & 69.5 & 83.3 & 45.8 & 8.9 & 24.7 & 35.5 & 17.8 \\
\midrule
GRAM\cpup    & 56.4 & 82.9 & 90.1 & 68.2 & \textbf{85.6} & \textbf{89.8} & \textbf{91.4} & \textbf{87.7} \\
HyperGRAM\cpup& 57.0 & 83.5 & 90.5 & 68.7 & 85.5 & 89.3 & 91.0 & 87.4 \\
PMRL\cpup    & 56.0 & 81.5 & 88.9 & 67.4 & 84.7 & 89.0 & 90.5 & 86.8 \\
TRIANGLE\cpup& \textbf{58.2} & \textbf{84.4} & \textbf{91.2} & \textbf{69.8} & 84.3 & 88.7 & 90.2 & 86.5 \\
\bottomrule
\end{tabular}
\end{adjustbox}
\end{table}

\begin{figure*}[tp]
\centering
\includegraphics[width=\linewidth]{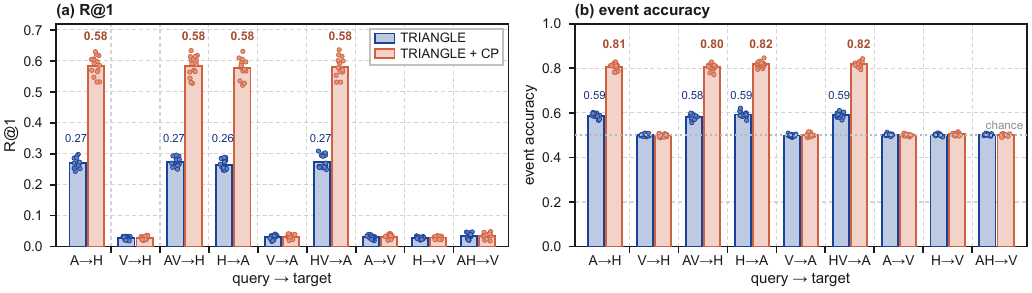}
\caption{\textbf{Retrieval from any subset of modalities.} Cluster S1 test, 5 folds $\times$ 3 seeds. Each bar pair is one query subset and target: TRIANGLE (blue) and TRIANGLE\,+\,CP (red), with weights fitted on development per query modality. Bars and numbers: means over the 15 runs. Dots: the runs. Subsets that contain the target are omitted. (a)~R@1. (b)~Event accuracy (dotted: chance). One model serves every subset. Audio and haptic targets gain 30.6 to 31.6 points of R@1 and reach an event accuracy of 80.5 to 81.8\%, and visual targets and visual-only queries, the path-only control, stay in place.}
\label{fig:anysubset}
\end{figure*}

\paragraph{Consistent gains across splits, folds, and held-out textures.}
\Cref{fig:splits} illustrates the four Cluster generalization splits from \cref{sec:exp_main}, and \cref{tab:splits} reports per-split performance for every aligner alongside the macro-average across splits.
\Cref{tab:perfold-cluster,tab:perfold-havdb} break down \cref{tab:main,tab:splits} and the HAVdb window scaling results (\cref{fig:scaling}) by individual fold. CounterPoint improves every base aligner across all 17 Cluster folds and all 15 HAVdb fold-by-window cells.
Paired statistical tests (\cref{tab:signif}, following \cref{sec:impl}) confirm that every confidence interval lies strictly above zero, with a minimum unit gain of 7.3 points in Avg R@1 (S4) and Holm-corrected $p \le 0.02$ across all test sets (including S4 with 3 runs).
\Cref{fig:texturebox} further shows consistent improvements across individual held-out textures.

% generated by scripts/nov_tables.py; every cell is a percentage
\begin{table}[tbp]
\centering
\caption{\textbf{Longer windows widen the gain on HAVdb.} Main task R@1 (\%, mean of AV$\to$H and HV$\to$A, H combining vibration and force) on unseen subjects for three window lengths. Superscripts: percentage-point gains over the base aligner. Bold: best.}
\label{tab:havdb}
\footnotesize
\setlength{\tabcolsep}{5pt}
\begin{adjustbox}{max width=\linewidth}
\begin{tabular}{llll}
\toprule
Window & 1\,s & 4\,s & 16\,s \\
\midrule
GRAM         & 21.0 & 20.0 & 9.8 \\
HyperGRAM    & 19.8 & 17.2 & 8.9 \\
PMRL         & 19.0 & 14.1 & 7.5 \\
TRIANGLE     & 19.7 & 17.6 & 8.9 \\
\midrule
GRAM\cpup    & \textbf{40.2}\up{19.2} & 72.6\up{52.6} & \textbf{85.6}\up{75.8} \\
HyperGRAM\cpup& 39.0\up{19.1} & 72.6\up{55.4} & 85.5\up{76.6} \\
PMRL\cpup    & 34.9\up{15.9} & 70.7\up{56.5} & 84.7\up{77.1} \\
TRIANGLE\cpup& 38.6\up{19.0} & \textbf{73.3}\up{55.7} & 84.3\up{75.4} \\
\bottomrule
\end{tabular}
\end{adjustbox}
\end{table}

% generated by scripts/nov_tables.py; every cell is a percentage
\begin{table*}[tp]
\centering
\caption{\textbf{Every task gains at every window on HAVdb.} R@1 of the tasks of \cref{tab:main} and \eacc\ in \% (chance 50) on unseen subjects, for models trained on windows of 1, 4 and 16\,s (5 folds $\times$ 3 seeds). H combines vibration and force. Superscripts: percentage-point gains over the base aligner. Bold/underline: best/second best.}
\label{tab:windows}
\footnotesize
\setlength{\tabcolsep}{2.4pt}
\begin{adjustbox}{max width=\linewidth}
\begin{tabular}{lllllllllllll}
\toprule
 & \multicolumn{4}{l}{1\,s} & \multicolumn{4}{l}{4\,s} & \multicolumn{4}{l}{16\,s} \\
\cmidrule(lr){2-5}\cmidrule(lr){6-9}\cmidrule(lr){10-13}
 & AV$\to$H & HV$\to$A & HA$\to$V & \eacc & AV$\to$H & HV$\to$A & HA$\to$V & \eacc & AV$\to$H & HV$\to$A & HA$\to$V & \eacc \\
\midrule
GRAM~\cite{cicchetti2025gram}& 33.0 & 9.0 & 26.7 & 75.5 & 32.0 & 8.0 & 23.2 & 73.8 & 14.0 & 5.6 & 10.1 & 69.3 \\
HyperGRAM~\cite{na2026hypergram}& 30.7 & 9.0 & 25.9 & 75.7 & 25.9 & 8.4 & 20.7 & 73.1 & 12.4 & 5.4 & 7.8 & 65.6 \\
PMRL~\cite{liu2025pmrl}& 27.7 & 10.2 & 24.7 & 77.2 & 21.1 & 7.2 & 17.1 & 71.2 & 10.5 & 4.6 & 7.4 & 64.9 \\
TRIANGLE~\cite{cicchetti2025triangle}& 30.2 & 9.2 & 25.9 & 76.4 & 26.3 & 8.8 & 20.3 & 73.3 & 12.4 & 5.5 & 8.2 & 66.3 \\
\midrule
GRAM\cpup    & \textbf{51.1}\up{18.1} & \underline{29.4}\up{20.4} & \underline{38.8}\up{12.1} & \underline{83.8}\up{8.4} & 91.2\up{59.2} & \underline{54.1}\up{46.1} & \textbf{81.4}\up{58.2} & 95.0\up{21.2} & \textbf{97.7}\up{83.7} & \textbf{73.5}\up{67.9} & \textbf{94.8}\up{84.7} & \textbf{97.3}\up{28.1} \\
HyperGRAM\cpup& \underline{48.9}\up{18.3} & 29.0\up{20.0} & \textbf{39.3}\up{13.4} & \textbf{84.3}\up{8.6} & \underline{91.4}\up{65.5} & 53.7\up{45.3} & \underline{81.1}\up{60.4} & \underline{95.1}\up{22.0} & \textbf{97.7}\up{85.3} & \underline{73.3}\up{67.9} & \underline{94.0}\up{86.2} & \underline{97.2}\up{31.6} \\
PMRL\cpup    & 43.5\up{15.8} & 26.3\up{16.1} & 36.0\up{11.3} & 81.3\up{4.1} & 88.5\up{67.4} & 52.9\up{45.7} & 79.9\up{62.8} & 94.9\up{23.7} & 96.9\up{86.4} & 72.4\up{67.8} & 93.8\up{86.4} & 96.6\up{31.7} \\
TRIANGLE\cpup& 47.6\up{17.4} & \textbf{29.7}\up{20.5} & 38.3\up{12.4} & 83.6\up{7.2} & \textbf{91.7}\up{65.4} & \textbf{54.8}\up{46.0} & \underline{81.1}\up{60.7} & \textbf{95.2}\up{21.9} & \underline{97.2}\up{84.8} & 71.5\up{66.0} & 93.7\up{85.5} & 96.8\up{30.5} \\
\bottomrule
\end{tabular}
\end{adjustbox}
\end{table*}

\begin{figure}[tbp]
\centering
\includegraphics[width=0.575\linewidth]{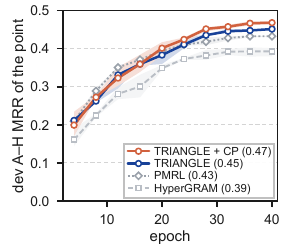}
\caption{\textbf{Point training is preserved under the trajectory loss.} Development A--H MRR of the point during training (fold 1, mean and range over 3 seeds): TRIANGLE in blue, TRIANGLE\,+\,CP in red, PMRL and HyperGRAM in grey for reference. Markers: the evaluations. In brackets: the MRR after 40 epochs.}
\label{fig:curves}
\end{figure}

\paragraph{Higher recall at every cut-off.}
\Cref{tab:recall} extends \cref{tab:main} to R@5 and R@10, showing improvements across all cut-offs and aligners. On Cluster, TRIANGLE R@10 increases from 83.3\% to 91.2\%. On 16\,s HAVdb windows, where point representations rarely retrieve the exact event, R@10 rises from 35.5\% to 90.2\%.

\paragraph{Arbitrary query subsets.}
A single trained CounterPoint model supports any subset of query modalities (\cref{fig:anysubset}) using development-fitted weights in \cref{supp:eq-score}.
Every subset querying audio or haptics from the other improves substantially, whether from a single modality (A$\to$H R@1 rises from 26.9\% to 58.5\%) or two modalities (AV$\to$H rises from 27.3\% to 58.4\%).
Consistent with theory, visual targets and visual-only queries remain near baseline, changing by at most $+0.2$ R@1 points with 50\% event accuracy.

\paragraph{Scaling across HAVdb window lengths.}
\Cref{tab:havdb} lists the HAVdb main-task R@1 (mean of AV$\to$H and HV$\to$A) across window lengths for all aligners (underlying \cref{fig:scaling}), and \cref{tab:windows} breaks it down into each task of \cref{tab:main} together with event accuracy.
For every base aligner, the performance gain from CounterPoint widens monotonically as the temporal window grows.

\begin{figure}[tbp]
\centering
\includegraphics[width=\linewidth]{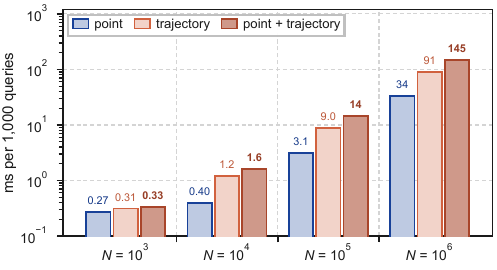}
\caption{\textbf{Scoring cost.} Time to score 1,000 queries against $N$ candidates as one half-precision matrix product on one RTX 4090: point (blue, 512 dimensions), flattened trajectory (light red, $25\times64$) and point\,+\,trajectory (dark red), on a logarithmic axis. Numbers: the times in ms.}
\label{fig:cost}
\end{figure}

\paragraph{Preserving point-level semantics.}
\looseness=-1 CounterPoint adds event discrimination without degrading the semantic information encoded in the point embedding.
Linear probes trained on frozen point representations predict material category, texture, direction, speed, and force with indistinguishable accuracy before and after adding CounterPoint (\cref{tab:downstream}).
Each probe takes standardized, concatenated unit points as input and is trained via Adam (learning rate $10^{-3}$, batch size 200, weight decay $10^{-4}$) with early stopping on a 10\% validation split, averaged over 5 seeds.
Probes are trained on training-split events and evaluated on held-out test textures. For texture classification (where test classes are unseen during encoder training), probes are fitted on one recording from each test twin pair and evaluated on the other.
Across all three probed input combinations (A alone, A+V, and A+V+H, with the latter shown in \cref{tab:downstream}), the smallest Holm-corrected $p$-value comparing TRIANGLE\,+\,CP against TRIANGLE is 0.18.
Point optimization dynamics are likewise unaffected: \cref{fig:curves} tracks the development MRR of the point branch throughout training, showing nearly identical trajectories for TRIANGLE and TRIANGLE\,+\,CP.

\begin{table}[tbp]
\centering
\caption{\textbf{Linear probes on frozen point embeddings.} Cluster S1 with A, V, and H inputs (one of three probed input sets). Metrics are accuracy (\%, $\uparrow$), $R^2$ ($\uparrow$), or MAE ($\downarrow$), with Holm-corrected $p$-values comparing TRIANGLE\,+\,CP against TRIANGLE.}
\label{tab:downstream}
\footnotesize
\setlength{\tabcolsep}{4pt}
\begin{adjustbox}{max width=\linewidth}
\begin{tabular}{lllllll}
\toprule
Task & GRAM & HyperG. & PMRL & TRI. & TRI.\cp & $p$ \\
\midrule
Category (Acc) & 63.8 & 63.8 & 60.3 & 63.3 & 63.5 & 1.00 \\
Texture (Acc) & 98.8 & 99.0 & 99.3 & 99.2 & 99.2 & 1.00 \\
Direction (Acc) & 99.1 & 99.1 & 99.0 & 99.2 & 99.2 & 1.00 \\
Speed (Acc) & 99.5 & 99.4 & 99.2 & 99.6 & 99.4 & 1.00 \\
Speed ($R^2$) & 0.972 & 0.971 & 0.958 & 0.972 & 0.971 & 1.00 \\
Force (Acc) & 98.6 & 98.7 & 98.9 & 98.7 & 98.8 & 1.00 \\
Force (MAE, mN) & 38.0 & 38.9 & 43.3 & 40.0 & 38.9 & 0.34 \\
\bottomrule
\end{tabular}
\end{adjustbox}
\end{table}

\paragraph{Hyperparameter sensitivity.}
\looseness=-1 \Cref{tab:sens} varies the trajectory loss weight $\lambda$ and projection dimension $k$ individually on Cluster S1 fold~1 (evaluated on the development split).
Across all tested configurations, A--H R@1 stays between 58.3\% and 62.1\% and event accuracy between 80.4\% and 83.1\% (within 3 and 2 points of the default $\lambda=0.5, k=64$), indicating low sensitivity to either hyperparameter.
\begin{table}[tbp]
\centering
\caption{\textbf{Hyperparameter sensitivity.} TRIANGLE\,+\,CP trained on Cluster S1 fold~1 and evaluated on the development split (3 seeds). Default settings are $\lambda=0.5$ and $k=64$.}
\label{tab:sens}
\footnotesize
\setlength{\tabcolsep}{4pt}
\begin{adjustbox}{max width=\linewidth}
\begin{tabular}{lll}
\toprule
Setting & A--H R@1 & \eacc \\
\midrule
Default & 59.2 & 81.2 \\
$\lambda=0.1$ & 62.1 & 83.1 \\
$\lambda=0.25$ & 60.1 & 81.4 \\
$\lambda=1.0$ & 59.1 & 80.8 \\
$k=16$ & 59.9 & 81.3 \\
$k=32$ & 59.5 & 81.0 \\
$k=128$ & 58.3 & 80.4 \\
\bottomrule
\end{tabular}
\end{adjustbox}
\end{table}

\paragraph{Computational overhead.}
The trajectory projection heads add 0.05\,M parameters to the 11.11\,M backbone (0.4\%) and 0.08\,GB to the 12.23\,GB peak training memory (0.7\%).
At inference, trajectory consistency reduces to a single inner product of flattened, Frobenius-normalized vectors, allowing standard vector indexing (\cref{fig:cost}). Benchmarked as a half-precision matrix multiplication on a single RTX 4090 GPU, scoring 1,000 queries against $10^6$ candidates requires 34\,ms for points alone, 91\,ms for trajectories alone, and 145\,ms for both combined.

%% .....................................................................
\FloatBarrier
\subsection{Trajectories follow the theory (RQ3)}
\label{sec:more-rq3}

\begin{figure*}[tp]
\centering
\includegraphics[width=\linewidth]{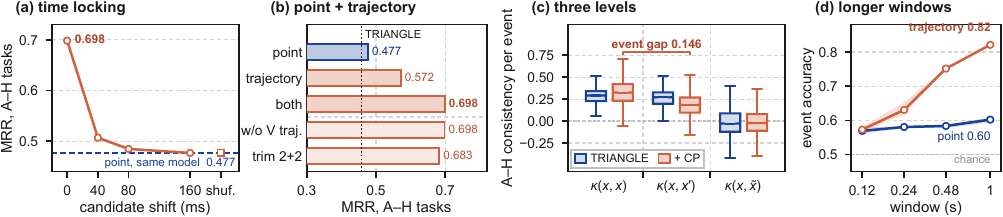}
\caption{\textbf{The gain is time-locked event dynamics, as the theory predicts.}
Cluster S1.
(a) Shifting candidates by one 40\,ms token removes most of the gain, and 160\,ms or shuffling returns to the point score of the same model.
(b) Points and trajectories are complementary. Dropping the visual trajectory or the window ends barely matters (dotted: TRIANGLE).
(c) Per-event audio--haptic consistency with the event, $\kappa(x,x)$, its twin event, $\kappa(x,x')$, and an event $\tilde x$ of another texture, $\kappa(x,\tilde x)$, on the development textures of the 5 folds $\times$ 3 seeds, the data of \cref{fig:mechanism}: TRIANGLE points (blue) barely separate $x$ from $x'$, TRIANGLE\,+\,CP trajectories (red) show the three levels of \cref{supp:thm-levels}. The path-only visual stream has no event gap (0.000 for A--V and V--H).
(d) Event accuracy by window (development split, 3 seeds): points collapse, trajectories scale.
Panels (a) and (b) report MRR, which complements the R@1 of \cref{supp:tab-ablation} and shows the same pattern.}
\label{fig:analysis}
\end{figure*}

\Cref{fig:analysis} summarizes four diagnostic views of the mechanism detailed below and in \cref{sec:more-rq4}: (a)~the time-locking control, (b)~point--trajectory complementarity, (c)~per-event distributions of the three consistency levels on the development split (underlying \cref{fig:mechanism}), and (d)~window scaling of event accuracy.

\begin{figure}[tbp]
\centering
\includegraphics[width=\linewidth]{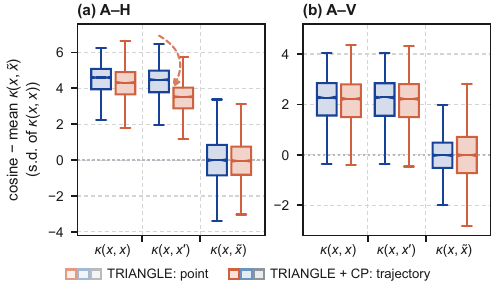}
\caption{\textbf{The three levels, event by event.} Cosine of a query with its own event, $\kappa(x,x)$, with its twin event $x'$, $\kappa(x,x')$, and with an event $\tilde x$ of another texture, $\kappa(x,\tilde x)$ (Cluster S1 fold~1, seed~0, test), centered on the mean of $\kappa(x,\tilde x)$ and scaled by the s.d.\ of $\kappa(x,x)$. Grey edges: the TRIANGLE point, colored edges: the TRIANGLE\,+\,CP trajectory. A--H: the trajectory separates $x$ from $x'$, the point does not. A--V: $x$ and $x'$ coincide for both, since the visual stream is path-only.}
\label{fig:levels}
\end{figure}

\begin{figure}[tbp]
\centering
\includegraphics[width=\linewidth]{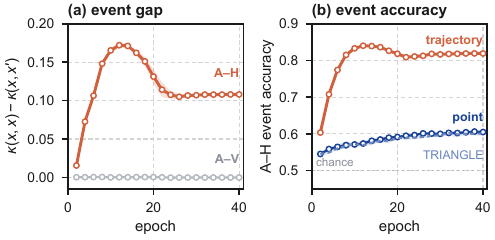}
\caption{\textbf{How the event gap is learned.} TRIANGLE\,+\,CP on Cluster S1 fold~1, development split, evaluated every 2 epochs (mean and $\pm1$ s.d.\ over 3 seeds). (a)~Event gap $\kappa(x,x)-\kappa(x,x')$ of the trajectory consistency: it emerges for A--H and stays at zero for A--V. (b)~A--H event accuracy of the trajectory and of the point of the same model, and of the point of TRIANGLE: the trajectory separates the twin event far above both points.}
\label{fig:epochgap}
\end{figure}

\paragraph{Three correlation levels on held-out test sets.}
The hierarchical ordering predicted by \cref{supp:thm-levels} holds on held-out test textures just as on the development split (\cref{fig:mechanism,fig:analysis}c).
\Cref{fig:levels} plots the per-event distributions on test textures for a single run. Averaged across all 5 test folds $\times$ 3 seeds on A--H, mean trajectory consistency reaches 0.487 for the true event, 0.381 for its twin, and 0.003 for a different texture, separating true events from twins by 0.106. By comparison, TRIANGLE point similarities yield 0.313, 0.302, and 0.024, a separation of only 0.011.
On A--V, true events and twin events yield identical similarities for both trajectories (0.281 vs.\ 0.281) and points (0.199 vs.\ 0.199), confirming the zero event gap predicted for a path-only visual stream.
The absolute magnitudes depend on the textures of each split, and the three-tier ordering holds on development and test textures alike, as predicted by \cref{supp:thm-levels}.

\paragraph{Emergence of the event gap during training.}
\looseness=-1 \Cref{fig:epochgap} tracks consistency and accuracy across training epochs on the development textures of fold~1.
The A--H trajectory event gap emerges within the first 10 epochs, whereas the A--V gap remains at zero throughout training, matching \cref{supp:thm-levels} for a modality lacking an event layer.
Trajectory event accuracy rises in lockstep with the A--H event gap, while the point branch of TRIANGLE\,+\,CP and the standalone TRIANGLE baseline remain flat near 60\%, confirming that within-condition discrimination comes from the trajectory.

\begin{figure}[tbp]
\centering
\includegraphics[width=\linewidth]{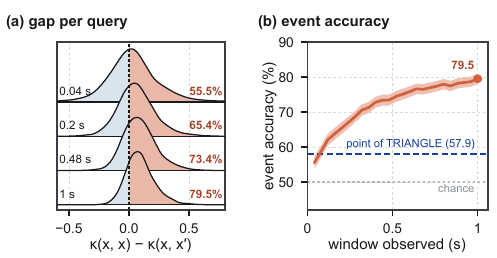}
\caption{\textbf{Evidence accumulates over the window.} One TRIANGLE\,+\,CP model scored on the first instants of its window (Cluster S1 fold~1, seed~0, 3,200 A$\to$H queries). (a)~Per query, the gap $\kappa(x,x)-\kappa(x,x')$ between the A--H trajectory cosine of \cref{supp:eq-kappa} with the true recording and with its twin event, after 0.04, 0.2, 0.48 and 1\,s. Red: queries where $x$ outscores $x'$, with their share. As the window grows, the gap narrows around a positive value. (b)~This share at every instant, the event accuracy ($\pm2$ s.e.): the gain over the point of TRIANGLE grows with every instant observed, the law of \cref{supp:fig-law} inside a single model.}
\label{fig:evidencetime}
\end{figure}

\begin{figure*}[tp]
\centering
\includegraphics[width=\linewidth]{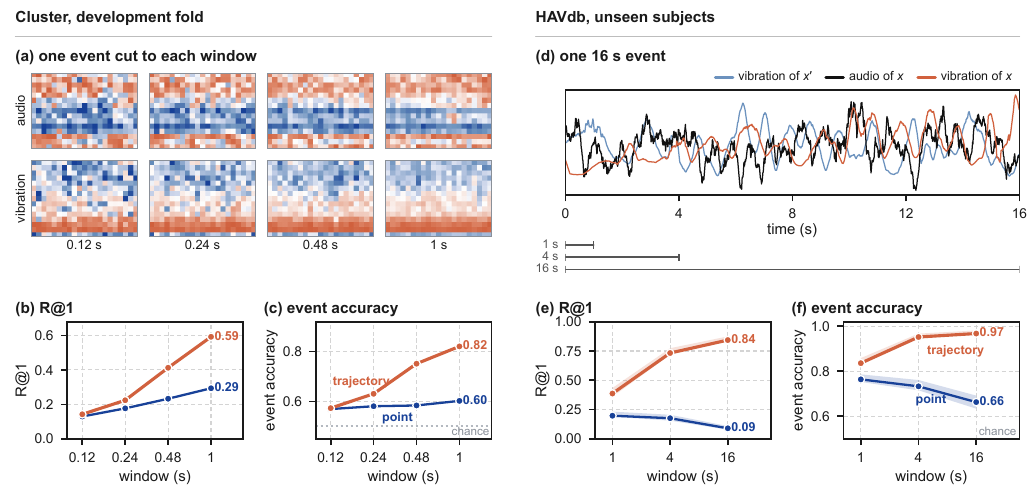}
\caption{\textbf{Points collapse, trajectories scale.} Left, Cluster, development split, 3 seeds: (a)~the audio (top) and vibration (bottom) spectrograms of one event over each window, each on 15 equal frequency bands up to 8 and 2\,kHz and 21 equal time cells of the window, in log power from blue (low) through white (the median of each modality) to red (high), (b)~A--H R@1 of TRIANGLE (blue) and TRIANGLE\,+\,CP (red) and (c)~A--H event accuracy of the point of TRIANGLE (blue) and of the trajectory of TRIANGLE\,+\,CP (red), from 0.12 to 1\,s windows. Right, HAVdb, unseen subjects, with H combining vibration and force as in \cref{tab:havdb,tab:windows}: (d)~the audio (black) and vibration (red) envelopes of one 16\,s event $x$ and the vibration of its twin event $x'$ (blue), with the 1, 4 and 16\,s windows below, (e)~main-task R@1 and (f)~event accuracy of TRIANGLE (blue) and TRIANGLE\,+\,CP (red), mean and interquartile range over the 15 runs. Numbers: values at the longest window.}
\label{fig:windowinsets}
\end{figure*}

\paragraph{Evidence accumulation across window lengths.}
\looseness=-1 Consistent with the bound in \cref{supp:eq-bound}, trajectories accumulate discriminative evidence continuously over time.
Within a single trained model (\cref{fig:evidencetime}), cumulative trajectory consistency for the true event stays above that of its twin from the earliest frames onward, driving event accuracy higher with every observed instant.
Comparing models trained across varying window lengths, \cref{fig:windowinsets} pairs the Cluster window study with the HAVdb scaling curves from \cref{fig:scaling}.
On Cluster, point event accuracy plateaus between 56.9\% and 60.2\% in accordance with \cref{supp:thm-stall}(i), whereas trajectory event accuracy climbs from 57.3\% to 82.1\% (\cref{fig:windowinsets}c).
While points still gain R@1 on Cluster by separating distinct textures (\cref{fig:windowinsets}b), only trajectories resolve twin events within the same texture.
On held-out HAVdb subjects, extending the window from 1\,s to 16\,s causes TRIANGLE mean A--H R@1 to drop from 19.7\% to 8.9\% as points collapse toward the condition mean, whereas TRIANGLE\,+\,CP climbs from 38.6\% to 84.3\% (\cref{tab:havdb}).

% generated by scripts/nov_tables.py; every cell is a percentage
\begin{table*}[tp]
\centering
\caption{\textbf{Single-modality retrieval on Cluster and HAVdb.} R@1 in \%. On HAVdb, H combines vibration and force and V is the tracked fingertip. Superscripts: percentage-point gains over the base aligner. Bold/underline: best/second best in each column.}
\label{tab:single}
\footnotesize
\setlength{\tabcolsep}{2pt}
\begin{adjustbox}{max width=\linewidth}
\begin{tabular}{llllllllllllll}
\toprule
 & \multicolumn{6}{l}{Cluster} & \multicolumn{6}{l}{HAVdb} & \\
\cmidrule(lr){2-7}\cmidrule(lr){8-13}
 & \multicolumn{2}{l}{H--A} & \multicolumn{2}{l}{H--V} & \multicolumn{2}{l}{A--V} & \multicolumn{2}{l}{H--A} & \multicolumn{2}{l}{H--V} & \multicolumn{2}{l}{A--V} & Avg \\
\cmidrule(lr){2-3}\cmidrule(lr){4-5}\cmidrule(lr){6-7}\cmidrule(lr){8-9}\cmidrule(lr){10-11}\cmidrule(lr){12-13}
 & H$\to$A & A$\to$H & H$\to$V & V$\to$H & A$\to$V & V$\to$A & H$\to$A & A$\to$H & H$\to$V & V$\to$H & A$\to$V & V$\to$A & \\
\midrule
GRAM~\cite{cicchetti2025gram}& 25.0 & 25.5 & 2.6 & 2.5 & 2.9 & 2.7 & 5.5 & 4.8 & 9.7 & 8.6 & 1.8 & 2.5 & 7.8 \\
HyperGRAM~\cite{na2026hypergram}& 25.5 & 25.7 & 2.7 & 2.5 & 2.7 & 2.7 & 5.1 & 4.5 & 8.5 & 8.4 & 1.5 & 2.1 & 7.7 \\
PMRL~\cite{liu2025pmrl}& 24.9 & 24.4 & 2.7 & 2.4 & 2.8 & 2.6 & 4.6 & 3.5 & 7.7 & 8.1 & 1.4 & 1.8 & 7.3 \\
TRIANGLE~\cite{cicchetti2025triangle}& 26.5 & 26.9 & \underline{2.9} & \underline{2.7} & 3.0 & \underline{3.0} & 4.8 & 4.9 & 7.4 & 7.9 & 1.8 & 2.0 & 7.8 \\
\midrule
GRAM\cpup    & 56.0\up{30.9} & 56.6\up{31.1} & 2.8\up{0.1} & 2.6\up{0.0} & \underline{3.1}\up{0.2} & 2.9\up{0.2} & \textbf{75.7}\up{70.2} & \textbf{82.2}\up{77.4} & \textbf{95.0}\up{85.4} & \textbf{94.0}\up{85.3} & \textbf{75.7}\up{73.9} & \textbf{68.2}\up{65.7} & \textbf{51.2}\up{43.4} \\
HyperGRAM\cpup& \underline{56.5}\up{31.1} & \underline{57.1}\up{31.4} & \underline{2.9}\up{0.2} & 2.6\up{0.1} & \underline{3.1}\up{0.4} & \underline{3.0}\up{0.2} & \underline{74.7}\up{69.6} & \underline{80.7}\up{76.2} & \underline{93.4}\up{84.9} & \underline{93.1}\up{84.7} & \underline{73.7}\up{72.2} & \underline{67.0}\up{64.9} & \underline{50.7}\up{43.0} \\
PMRL\cpup    & 55.7\up{30.8} & 56.3\up{31.9} & \underline{2.9}\up{0.3} & \underline{2.7}\up{0.3} & 3.0\up{0.2} & 2.8\up{0.2} & 73.4\up{68.8} & 79.8\up{76.2} & 92.1\up{84.4} & 92.7\up{84.5} & 71.6\up{70.2} & 66.2\up{64.4} & 49.9\up{42.7} \\
TRIANGLE\cpup& \textbf{57.7}\up{31.2} & \textbf{58.5}\up{31.6} & \textbf{3.0}\up{0.1} & \textbf{2.8}\up{0.1} & \textbf{3.2}\up{0.1} & \textbf{3.1}\up{0.0} & 73.4\up{68.5} & 79.9\up{74.9} & 93.2\up{85.8} & 93.0\up{85.1} & 72.3\up{70.5} & 65.6\up{63.6} & 50.5\up{42.6} \\
\bottomrule
\end{tabular}
\end{adjustbox}
\end{table*}

% Cluster: results/paper_numbers.json (layers); HAVdb: results/havdb_levels.csv (dev); development data, 5 folds x 3 seeds
\begin{table}[tbp]
\centering
\caption{\textbf{The three levels in the two regimes.} Time-locked consistency of the TRIANGLE\,+\,CP trajectories of an event $x$ with itself, its twin event $x'$ and an event $\tilde x$ of another texture, on the development split (5 folds $\times$ 3 seeds). Twin events share the drive on Cluster (1\,s) and only the condition on HAVdb (16\,s). Event gaps are computed from the unrounded values.}
\label{tab:regimes}
\footnotesize
\setlength{\tabcolsep}{4pt}
\begin{adjustbox}{max width=\linewidth}
\begin{tabular}{lllll}
\toprule
Pair & $\kappa(x,x)$ & $\kappa(x,x')$ & $\kappa(x,\tilde x)$ & Event gap \\
\midrule
\multicolumn{5}{l}{\textit{Cluster, shared drive}} \\
\quad A--H & 0.328 & 0.183 & $-0.013$ & 0.146 \\
\quad A--V & 0.091 & 0.091 & $-0.004$ & 0.000 \\
\quad V--H & 0.088 & 0.087 & $-0.005$ & 0.000 \\
\midrule
\multicolumn{5}{l}{\textit{HAVdb, shared condition}} \\
\quad \Hv--A & 0.347 & 0.002 & $-0.009$ & 0.345 \\
\quad \Hf--A & 0.223 & 0.001 & $-0.007$ & 0.223 \\
\quad A--V & 0.284 & $-0.001$ & $-0.008$ & 0.284 \\
\quad \Hf--V & 0.518 & 0.006 & $-0.012$ & 0.512 \\
\quad \Hv--\Hf & 0.273 & 0.004 & $-0.007$ & 0.268 \\
\quad \Hv--V & 0.347 & 0.003 & $-0.008$ & 0.344 \\
\bottomrule
\end{tabular}
\end{adjustbox}
\end{table}

\begin{figure}[tbp]
\centering
\includegraphics[width=\linewidth]{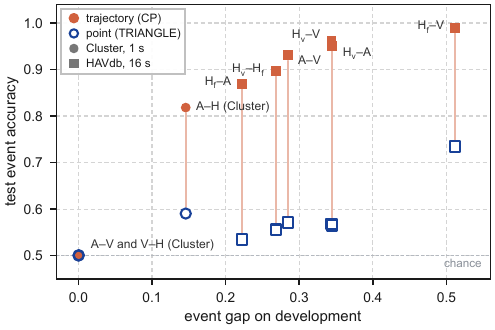}
\caption{\textbf{Event accuracy follows the event gap.} Each sensor pair of both datasets, placed by its event gap on the development split (\cref{tab:regimes}) and its event accuracy on test data, mean over 5 folds $\times$ 3 seeds: the trajectory of TRIANGLE\,+\,CP (filled) and the point of TRIANGLE (open), joined by a line. Circles: Cluster, 1\,s. Squares: HAVdb, 16\,s. Dotted line: chance.}
\label{fig:gapacc}
\end{figure}

\begin{figure}[tbp]
\centering
\includegraphics[width=\linewidth]{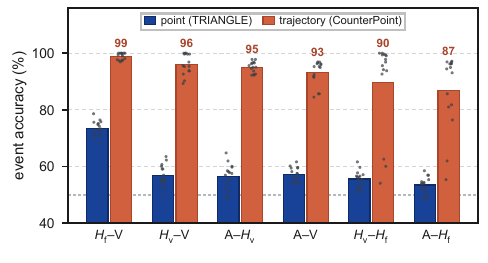}
\caption{\textbf{Every HAVdb sensor pair couples.} Unseen subjects, 16\,s windows, 15 runs: event accuracy of the point (TRIANGLE) and of the trajectory (CounterPoint) for every pair of sensors. H$_\mathrm{v}$ vibration, H$_\mathrm{f}$ force, A audio, V the camera-tracked fingertip. Bars: mean, dots: runs. The tracked fingertip exhibits cross-modal dynamics, while the path-only Cluster visual stream serves as a null control (\cref{fig:mechanism} of the main paper).}
\label{fig:havdbpairs}
\end{figure}

\paragraph{Retrieval gains track the event gap.}
\Cref{tab:single} details every single-modality retrieval direction behind the event-gap analysis of \cref{sec:exp_levels}. On Cluster, gains concentrate on the audio--haptic pair where an event gap exists. On HAVdb, because fingertip tracking captures freehand motion dynamics, every modality pair exhibits an event gap and improves substantially.

\paragraph{Contrasting the two dataset regimes.}
\Cref{tab:regimes} reports the empirical three-level correlations on the development splits of both datasets following \cref{sec:impl}.
In Cluster's shared-drive regime, twin A--H events maintain an intermediate consistency of 0.183 between true events (0.328) and unrelated textures ($-0.013$) due to their shared trajectory drive.
In HAVdb's shared-condition regime, freehand twin events drop to near zero across all sensor pairs (at most 0.006, within 0.02 of unrelated textures), matching the independent-drive prediction of \cref{supp:thm-levels}. Because the twin baseline vanishes, the event gap spans nearly the entire same-event consistency, accounting for the larger retrieval gains on HAVdb in \cref{tab:main}.

\paragraph{Event gap predicts test event accuracy.}
\looseness=-1 \Cref{fig:gapacc} plots all nine sensor pairs across both datasets by their development-set event gap against their test-set event accuracy.
On 1\,s Cluster windows, the two path-only visual pairs have zero event gap and remain near chance, whereas audio--haptics has an event gap of 0.146 and reaches 81.8\% accuracy.
On 16\,s HAVdb windows, every one of the six sensor pairs has a positive event gap and a trajectory event accuracy of at least 87\%, and the pair with the widest development event gap, force--position (0.512), reaches the highest accuracy (98.9\%), while the pair with the narrowest gap, force--audio (0.223), reaches the lowest (87.0\%).
By contrast, TRIANGLE point accuracy remains between 50.0\% and 59.0\% on every pair except force--position (73.5\%).
Within each dataset, a wider event gap reduces the constant $K\propto(\bar\rho^{e}_{qc})^{-2}$ in \cref{supp:thm-stall}(ii), allowing development-set event gaps to anticipate test-set event accuracy.

\paragraph{All HAVdb sensor pairs couple strongly.}
\Cref{fig:havdbpairs} displays 16\,s event accuracy across all six HAVdb sensor pairs, and \cref{tab:havdbpairs} reports pairwise MRR and event accuracy across all three window lengths.
Trajectory MRR and event accuracy improve with window length for every sensor pair, whereas TRIANGLE point MRR declines from 1\,s to 16\,s across every pair as point embeddings converge toward the condition limit (\cref{supp:thm-collapse}).

\begin{table*}[tp]
\centering
\caption{\textbf{Pairwise retrieval on HAVdb across window lengths.} Evaluated on unseen subjects (5 folds $\times$ 3 seeds, all values in \%). Point: TRIANGLE pairwise MRR. Traj.: CounterPoint trajectory pairwise MRR. \eacc: trajectory event accuracy. Bold indicates the higher MRR.}
\label{tab:havdbpairs}
\footnotesize
\setlength{\tabcolsep}{4.5pt}
\begin{adjustbox}{max width=\linewidth}
\begin{tabular}{llllllllll}
\toprule
 & \multicolumn{3}{l}{1\,s} & \multicolumn{3}{l}{4\,s} & \multicolumn{3}{l}{16\,s} \\
\cmidrule(lr){2-4}\cmidrule(lr){5-7}\cmidrule(lr){8-10}
Pair & Point & Traj. & \eacc & Point & Traj. & \eacc & Point & Traj. & \eacc \\
\midrule
\Hf--A & 11.7 & \textbf{16.3} & 68.0 & 10.5 & \textbf{42.9} & 86.6 & 8.3 & \textbf{56.7} & 87.0 \\
\Hv--A & 12.5 & \textbf{33.6} & 77.7 & 12.5 & \textbf{62.3} & 91.1 & 9.9 & \textbf{78.9} & 95.0 \\
A--V & 10.3 & \textbf{21.0} & 71.7 & 8.2 & \textbf{54.6} & 88.9 & 6.2 & \textbf{73.7} & 93.1 \\
\Hf--V & 37.6 & \textbf{45.5} & 87.2 & 31.4 & \textbf{85.3} & 98.1 & 15.6 & \textbf{91.4} & 98.9 \\
\Hv--\Hf & 16.5 & \textbf{23.0} & 76.1 & 13.4 & \textbf{53.5} & 88.7 & 9.9 & \textbf{71.3} & 89.7 \\
\Hv--V & 12.6 & \textbf{28.5} & 81.5 & 8.7 & \textbf{67.9} & 91.8 & 6.3 & \textbf{88.3} & 96.1 \\
\bottomrule
\end{tabular}
\end{adjustbox}
\end{table*}

\begin{table}[tbp]
\centering
\caption{\textbf{Ablation of CounterPoint components} (reproduced from \cref{tab:ablation} of the main paper). TRIANGLE\,+\,CP evaluated on Cluster S1 (\%). Time-locking rows shift or permute candidate trajectories at test time, while the remaining rows remove twin negatives during training or omit the point score at inference.}
\label{supp:tab-ablation}
\footnotesize
\setlength{\tabcolsep}{3.5pt}
\begin{adjustbox}{max width=\linewidth}
\begin{tabular}{llll}
\toprule
Variant & A--H & Avg & \eacc \\
\midrule
TRIANGLE & 27.4 & 19.4 & 58.5 \\
TRIANGLE\,+\,CP & 58.2 & 40.0 & 81.1 \\
\midrule
\quad candidates shifted by 40\,ms (one token) & 32.3 & 22.7 & 60.7 \\
\quad candidates shifted by 160\,ms & 29.0 & 20.4 & 59.1 \\
\quad candidate tokens permuted & 29.0 & 20.5 & 59.2 \\
\quad no twin event in any training batch & 55.7 & 38.3 & 78.8 \\
\quad demeaned, trajectory-only & 45.8 & 31.2 & 81.7 \\
\bottomrule
\end{tabular}
\end{adjustbox}
\end{table}

%% .....................................................................
\FloatBarrier
\subsection{Component ablations (RQ4)}
\label{sec:more-rq4}
\paragraph{Temporal synchronization drives the trajectory gain.}
\Cref{supp:tab-ablation} reproduces the main ablation table, and \cref{fig:analysis}a,\,b visualize its primary findings.
Shifting candidate trajectories by just 40\,ms (one token) eliminates most of the retrieval gain, while a 160\,ms shift or a random token permutation reduces performance to 29.0\% A--H R@1 and 20.4--20.5\% Avg R@1.
This matches the point-only score of the same model ($\beta_q=0$ in \cref{supp:eq-score}, which achieves the same 47.7\% MRR in \cref{fig:analysis}a), confirming that the trajectory branch contributes through frame-accurate temporal alignment.
Trimming two tokens from each boundary retains 38.5\% Avg R@1 (versus 40.0\% on the full window), indicating that discriminative cues are distributed throughout the recording.
Dropping visual trajectories changes Avg R@1 by only 0.1 points, consistent with their zero event gap.

% generated by scripts/nov_tables.py; every cell is a percentage
\begin{table*}[tp]
\centering
\caption{\textbf{Temporal alignment methods per fold.} Cluster S1 test, mean over 3 seeds per fold (\%). Each score replaces the trajectory score of TRIANGLE\,+\,CP in training and test, with the same encoders and loss, to which FATE adds its offset loss. Rows below the rule score demeaned trajectories. Joint: with the point, as in \cref{supp:eq-score}. Trajectory only: the trajectory score alone. CounterPoint leads every temporal alignment method on every fold.}
\label{tab:temporalfold}
\footnotesize
\setlength{\tabcolsep}{4pt}
\begin{adjustbox}{max width=\linewidth}
\begin{tabular}{lllllllllll}
\toprule
 & & \multicolumn{5}{l}{Joint A--H R@1 per fold} & \multicolumn{2}{l}{Joint, mean} & \multicolumn{2}{l}{Trajectory only} \\
\cmidrule(lr){3-7}\cmidrule(lr){8-9}\cmidrule(lr){10-11}
Score & Instants matched & 1 & 2 & 3 & 4 & 5 & A--H R@1 & \eacc & A--H R@1 & \eacc \\
\midrule
TRIANGLE (point) & pooled & 25.7 & 29.4 & 26.1 & 29.3 & 26.3 & 27.4 & 58.5 & -- & -- \\
CAV-MAE Sync~\cite{araujo2025cavmae} & same & 26.7 & 30.7 & 27.5 & 30.0 & 29.9 & 29.0 & 59.6 & 29.2 & 60.4 \\
FATE~\cite{guan2026fate} & same & 25.7 & 30.4 & 26.3 & 29.1 & 27.4 & 27.8 & 58.4 & 27.0 & 58.5 \\
\midrule
MaxSim~\cite{reddy2025videocolbert} & any & 25.3 & 29.7 & 25.5 & 27.8 & 26.6 & 27.0 & 57.7 & 19.4 & 55.0 \\
Soft-DTW~\cite{yang2023tempclr} & monotone warp & 34.2 & 38.6 & 35.7 & 37.1 & 40.8 & 37.3 & 65.1 & 25.5 & 64.4 \\
Best shift~\cite{chung2016outoftime} & shift $\le$\,160\,ms & 43.3 & 50.6 & 46.8 & 49.3 & 47.8 & 47.6 & 74.1 & 40.9 & 75.1 \\
\rowcolor{bpOursFill}
CounterPoint & same & \textbf{53.6} & \textbf{60.2} & \textbf{58.6} & \textbf{62.2} & \textbf{56.6} & \textbf{58.2} & \textbf{81.1} & \textbf{45.8} & \textbf{81.7} \\
\bottomrule
\end{tabular}
\end{adjustbox}
\end{table*}

\paragraph{Demeaning and zero-lag scoring outperform alternative alignment objectives.}
\Cref{tab:temporalfold} provides per-fold results for the temporal alignment baselines in \cref{tab:temporal}, along with standalone trajectory scores (without the point term).
CounterPoint outperforms every alternative alignment method on every fold, leading the second-best method (Best shift) by at least 8.8 points in A--H R@1 (fold~5).
Under the paired test of \cref{sec:impl}, all gains in A--H R@1, Avg R@1, and event accuracy are statistically significant (Holm-corrected $p<0.001$), with a 95\% confidence interval of $[8.6, 12.8]$ points for the A--H R@1 margin over Best shift.
Evaluating standalone trajectory scores preserves the same ranking, descending from CounterPoint (45.8\%) to Best shift (40.9\%), Soft-DTW (25.5\%), and MaxSim (19.4\%).
By contrast, CAV-MAE Sync (which omits temporal demeaning) achieves 29.2\% standalone and 29.0\% combined with the point score, confirming that without the $I-P$ projection the trajectory branch retains the condition layer and behaves like a point representation.
Similarly, FATE's offset classification loss achieves 27.0\% standalone, remaining at the point baseline.
Finally, Best shift trails CounterPoint on every fold, corroborating \cref{supp:cor-instants}: when twin events share a common drive, allowing temporal shifts can raise only the twin similarity and never widens the event gap.

\begin{figure}[tbp]
\centering
\includegraphics[width=\linewidth]{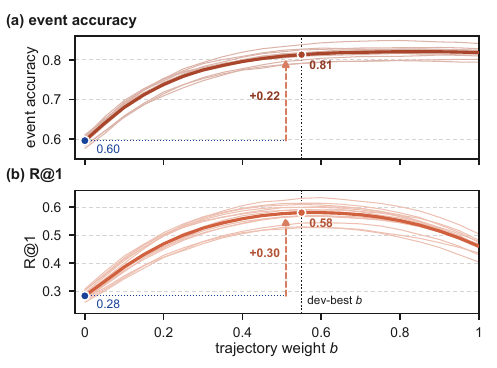}
\caption{\textbf{Weighting point and trajectory.} R@1 and event accuracy for the score $(1-b)\,\tilde z+b\,\tilde c$ with standardized point and trajectory cosines, A$\to$H and H$\to$A (Cluster S1 test). (a)~Event accuracy and (b)~R@1. Thin lines: the 15 runs (5 folds $\times$ 3 seeds). Bold line: their mean. Blue dots: the point alone ($b=0$). Dotted: $b=0.55$, the grid value with the best development MRR in 12 of the 15 runs, the others selecting 0.5 or 0.6. Arrows: the gain over the point alone, from 0.60 to 0.81 in event accuracy and from 0.28 to 0.58 in R@1, where R@1 peaks.}
\label{fig:betasweep}
\end{figure}

\paragraph{Complementarity of points and trajectories.}
\Cref{fig:betasweep} sweeps the trajectory mixing weight $b\in[0,1]$ in the standardized score $(1-b)\,\tilde z+b\,\tilde c$, where $\tilde z$ and $\tilde c$ denote the standardized point and trajectory similarities (\cref{sec:impl}). Because standardization applies a query-independent affine shift, this formulation induces the exact same candidate ranking as \cref{supp:eq-score} with $\alpha_q=(1-b)/\mathrm{sd}_z$ and $\beta_q=b/\mathrm{sd}_c$.
Evaluated across 21 grid values on single-modality queries (A$\to$H and H$\to$A), the point-only score ($b=0$) yields 28.3\% R@1 and 59.6\% event accuracy, while the trajectory-only score ($b=1$) achieves 46.1\% R@1 and 81.8\% event accuracy.
Combining both branches peaks at 58.1\% R@1 for $b=0.55$ (the median development-optimal weight across the 15 runs, ranging from 0.50 to 0.60), closely matching the 58.2\% A--H R@1 obtained by L-BFGS weight calibration on multi-modality queries (\cref{supp:tab-ablation}).
R@1 remains within 2 points of its peak across $b\in[0.45, 0.75]$ while event accuracy stays flat for $b\ge 0.5$, confirming that the two branches provide complementary signals over a broad operating range.

%% =====================================================================

\FloatBarrier
\section{Extended related work}
\label{sec:related-ext}

\paragraph{Geometric multimodal alignment.}
\looseness=-1 Contrastive alignment maps each modality to a single embedding vector and scores pairs via cosine similarity~\cite{radford2021clip}.
Recent extensions generalize this pairwise formulation to constrain all modalities simultaneously: GRAM~\cite{cicchetti2025gram} minimizes the volume of the parallelotope spanned by the embeddings, TRIANGLE~\cite{cicchetti2025triangle} minimizes the area of their triangle, PMRL~\cite{liu2025pmrl} regularizes the singular values of the stacked embeddings, HyperGRAM~\cite{na2026hypergram} computes Gramian volumes in hyperbolic space, and BaryBind~\cite{tang2026barybind} aligns representations via a Wasserstein barycenter.
Although these objectives differ in their underlying geometry, each vertex remains a time-pooled vector subject to the condition collapse of \cref{supp:thm-collapse}.
Multi-vector retrieval retains multiple embeddings per item, either to represent distinct semantic senses~\cite{song2019pvse} or to match tokens via late interaction~\cite{reddy2025videocolbert}. However, late interaction pairs each token with its highest-scoring counterpart regardless of timestamp, rather than enforcing alignment at corresponding instants on a shared clock.
In multimodal sensing, FOCAL~\cite{liu2023focal} factorizes time-series embeddings into shared and modality-private subspaces, recent work advocates shared physical representations across sensors~\cite{tan2026beyond}, and diffusion models synthesize audio and haptic signals from a joint latent space~\cite{kotani2026audiohapticdiffusion}.

\paragraph{Temporal correspondence across modalities.}
Temporal alignment has been studied extensively in audio--visual learning, where lip-synchronization networks minimize distance at the true temporal offset~\cite{chung2016outoftime,chung2019perfectmatch} and self-supervised models contrast synchronized streams against temporally shifted negatives~\cite{owens2018multisensory,korbar2018avts}.
More recently, FATE~\cite{guan2026fate} and CAV-MAE Sync~\cite{araujo2025cavmae} align audio and video frame by frame, sequence-level methods align video steps with text~\cite{yang2023tempclr,lin2024norton}, local music alignment is learned from global supervision~\cite{bukey2026music}, and MoBind~\cite{nguyen2026mobind} binds IMU signals to video poses at fine temporal resolution.
Because these approaches either search over unknown offsets or construct negatives from different clips and shifted windows, they primarily test whether content matches across time.
In our setting, the sensor clock is shared and the offset is known, so the retrieval task is to distinguish a specific recording from twin events captured under identical conditions.
While hard negatives from the same semantic class are known to sharpen contrastive representations~\cite{khosla2020supcon,robinson2021hard}, twin events provide stricter supervision by sharing the stationary condition (and often the external drive) while differing solely in event-specific dynamics.

\paragraph{Ergodic theory, Koopman operators, and summary statistics.}
The Koopman operator~\cite{koopman1931hamiltonian} represents nonlinear dynamics as a linear operator on observables, enabling dynamical systems to be compared through their limiting ergodic averages~\cite{mezic2004comparison}.
Prior machine learning studies use Koopman operators to disentangle sequential factors~\cite{berman2023koopman} and link joint-embedding predictive architectures to Koopman invariants that govern the clustering of single-modality time series~\cite{ruizmorales2025koopman}. Those works focus on the invariant subspace of the dynamics, corresponding to the condition layer in our decomposition.
To characterize the remaining layers, our analysis combines Birkhoff's ergodic theorem~\cite{birkhoff1931proof} with weak mixing and relative products~\cite{furstenberg1977ergodic}, while the saturation of pooled points parallels fundamental limits on hypothesis testing from compressed statistics~\cite{ahlswede1986hypothesis}.
Finally, our twin-event estimator in \cref{eq:twin-estimate} echoes the shift predictor in neuroscience, which subtracts trial-shuffled correlations to isolate single-trial neural coupling~\cite{aertsen1989dynamics,cohen2011correlations}, while auditory texture perception relies on time-averaged summary statistics~\cite{mcdermott2013summary} analogous to pooled point representations.
Together, these tools formalize what time-pooled multimodal embeddings preserve and what they discard.
\section{Scope, outlook, and broader impact}
\label{sec:impact}

\paragraph{Scope and outlook.}
\looseness=-1 CounterPoint exploits the shared clock of synchronized multisensor systems at 40\,ms token resolution (\cref{sec:more-rq4}).
Retrieval gains track cross-modal coupling at the event layer. On Cluster, audio and haptics share an event layer and improve sharply, whereas the path-only visual stream lacks one and remains unchanged. On HAVdb, force and fingertip position both capture freehand dynamics and achieve the highest trajectory MRR and event accuracy at every window length (\cref{tab:havdbpairs}). The empirical event gap measured on twin trials therefore indicates how much a new sensor pair gains from trajectory alignment, and for audio and haptics on Cluster the raw envelopes already show this gap before any learning (\cref{fig:layers}b).

\looseness=-1 Our evaluation spans two complementary regimes, pairing motorized 1\,s strokes on Cluster with unconstrained freehand trajectories up to 16\,s on HAVdb. Extending controlled, shared-drive benchmarks to longer horizons is a natural next test. Theoretically, our guarantees assume stationarity, ergodicity, and relative weak mixing for a fixed encoder (\cref{supp:thm-stall}), leaving joint encoder optimization dynamics during training as an open analytical direction.

\paragraph{Broader impact.}
\looseness=-1 Resolving the event layer links a sensor stream to its exact recording instance, supporting fine-grained quality control, process monitoring, and multisensory data provenance. Because this property also makes individual trials identifiable, deployments with human subjects should treat event-preserving trajectories as potentially identifying data. All experiments use public datasets of surface textures, physical sensor measurements, and tracked fingertip motion.

\end{document}